\documentclass[twocolumn]{svjour3}

\usepackage[latin1]{inputenc}
\usepackage{times}

\usepackage{psfrag, amsmath, amssymb, amscd, amsfonts, latexsym, epsf, graphicx}
\usepackage{natbib}
\usepackage[switch]{lineno}

\def\bbbr{{\mathbb R}} 

\def\bbbs{{\mathbb S}}

\journalname{arXiv preprint}

\begin{document}

\title{\bf Hybrid Lie semi-group and cascade
  structures for the generalized Gaussian
  derivative model for visual receptive fields%
\thanks{The support from the Swedish Research Council 
              (contract 2022-02969) is gratefully acknowledged. }}

\titlerunning{Hybrid Lie semi-group and cascade structures for the generalized Gaussian
  derivative model for visual receptive fields}

\author{Tony Lindeberg}

\institute{Tony Lindeberg \at
                Computational Brain Science Lab,
                Department of Computational Science and Technology,
                KTH Royal Institute of Technology,
                SE-100 44 Stockholm, Sweden.
                \email{tony@kth.se}}

\date{Received: date / Accepted: date}

\maketitle

\begin{abstract}
  Because of the variabilities of real-world image structures under
  the natural image transformations that arise when observing similar
  objects or spatio-temporal events under different viewing
  conditions, the receptive field responses
  computed in the earliest layers of the visual hierarchy may be
  strongly influenced by such geometric image transformations.
  One way of handling this variability is by basing the vision system
  on covariant receptive field families, which expand the receptive
  field shapes over the degrees of freedom in the image
  transformations.

  This paper addresses the problem of deriving relationships between
  spatial and spatio-temporal receptive field responses obtained
  for different values of the shape
  parameters in the resulting multi-parameter families of receptive
  fields. For this purpose, we derive both (i)~infinitesimal relationships,
  roughly corresponding to a combination of notions from semi-groups
  and Lie groups, as well as (ii)~macroscopic cascade smoothing properties,
  which describe how receptive field responses at coarser spatial and
  temporal scales can be computed by applying smaller support
  incremental filters to the output from corresponding receptive
  fields at finer spatial and temporal scales,
  structurally related to the notion 
  of Lie algebras, although with directional preferences. 

  For the receptive field models based on spatio-temporal smoothing
  using a sole combination of affine Gaussian smoothing kernels over
  image space with non-causal temporal Gaussian kernels over the
  temporal domain, we derive reasonably complete results in this
  respect, by exploiting the specific structure of the joint 2+1-D
  Gaussian spatio-temporal model. This permits characterizations in terms of
  higher-dimensional generalizations of the notion of Hermite polynomials, as well
  as characterizations in terms of joint spatio-temporal covariance
  matrices.

  For the time-causal spatio-temporal receptive field model,
  where the temporal smoothing is instead performed using the
  time-causal limit kernel, we derive less complete results, however,
  still highly useful for the special case when the velocity
  parameters assume equal values in both the incremental convolution
  kernel as well as in the corresponding layers of the spatio-temporal
  scale-space representation.

  The presented results provide (i)~a deeper understanding of the
  relationships between spatial and spatio-temporal receptive field
  responses for different values of the filter parameters, which can
  be used for both (ii)~designing more efficient schemes for computing
  receptive field responses over populations of multi-parameter
  families of receptive fields, as well as (iii)~formulating idealized
  theoretical models of the computations of simple cells in biological vision.

\end{abstract}

\section{Introduction}
\label{sec-intro}

When a visual observer views objects or spatial-temporal events in the
environment, the image data reaching the visual sensor may be subject
to substantial variabilities caused by variabilities in the viewing
conditions, as resulting from varying the distance, the viewing direction
and the relative motion between the objects in the world and the
observer. The influence of the resulting geometric image
transformations will, in turn, strongly affect the responses of the
receptive fields in the early layers the visual hierarchy.

To handle such variabilities in image structures, caused by variations
in the viewing conditions, the notion of multi-parameter scale spaces
has been proposed regarding the families of receptive fields
(Lindeberg \citeyear{Lin10-JMIV,Lin21-Heliyon}).
Specifically, from the desirable property of provable
covariance properties of the families of spatial and/or
spatio-temporal receptive fields under the appropriate families of
geometric image transformations,
the shapes of the spatial or spatio-temporal receptive fields ought to be
expanded over the degrees of freedom of the corresponding 
image transformations (Lindeberg \citeyear{Lin25-JMIV}).
Notably, such a notion has been proposed as a tentative model
for variabilities of simple cells in the primary cortex of higher mammals
(Lindeberg \citeyear{Lin25-arXiv-cov-props-review}).

If one in an idealized vision system is to take the view that a rich
set of receptive field responses is to be computed for a rich variety of
shape parameters of the receptive fields, one can raise  the question
if this could be done in a more efficient manner than computing
each receptive field response separately. Similarly, one may ask
how receptive field responses computed for different values of the
filter parameters could be related. For example, for the
pure spatial scale-space representation, defined from convolution with either
isotropic or anisotropic affine Gaussian kernels, the underlying spatial smoothing filters
obey cascade smoothing properties over spatial scales, implying that any
representation at a coarser spatial scale can be computed by applying
a filter or a set of filters to the representations at any finer
scale.

For purposes of computer implementation, making use of that
cascade smoothing property can substantially reduce the computational
work, by making it possible to build the implementation based on
a set of filtering operations of smaller spatial support compared to
a naive implementation of each receptive field response,
which thereby reduces the amount of computations.
Similarly, by the special property of the spatial smoothing operation,
the cascade smoothing property resulting from the semi-group
property of the underlying either isotropic or anisotropic affine
Gaussian smoothing kernels guarantees a simplifying property from
finer to coarser level of scales, crucial for the formal definition of
a simplifying property from finer to coarser scales in
the corresponding spatial scale-space representation
(Iijima \citeyear{Iij62}, Koenderink \citeyear{Koe84},
Koenderink and van Doorn \citeyear{KoeDoo92-PAMI},
Lindeberg \citeyear{Lin96-ScSp}, \citeyear{Lin10-JMIV}, 
Weickert {\em et al.\/}\ \citeyear{WeiIshImi99-JMIV}).

The subject of this paper, is to generalize the above mentioned ideas for the
specific generalized Gaussian derivative model for visual receptive
fields, initially derived in Lindeberg (\citeyear{Lin10-JMIV,Lin13-BICY}) and
then refined in Lindeberg (\citeyear{Lin16-JMIV,Lin21-Heliyon,Lin23-FrontCompNeuroSci,Lin25-JMIV, Lin25-arXiv-cov-props-review}).
Using the algebraic properties of the idealized models
for visual receptive fields according to this framework,
we will derive closed-form expressions
for derivatives with respect to the filter parameters in this model
as well as explicit macroscopic cascade relations between
receptive field responses for different parameter settings,
roughly corresponding to a combination of Lie group
structures with multi-parameter semi-group structures,
comprising 3 effective parameters in the purely spatial case and
4 or 6 effective parameters in the joint spatio-temporal case,
although for special reasons of separating the degree of freedom
corresponding to purely spatial scaling transformations,
overparameterized over 4 or 7 parameters when using affine Gaussian
kernels for the spatial smoothing operations.

This will result in (i)~sets of infinitesimal generators and
(ii)~sets of macroscopic cascade smoothing properties over subsets
of the parameter spaces, for which the evolution properties can only
performed in the positive parameter directions. For other subsets in
the parameter space, the corresponding relationships between receptive
field responses for different parameter settings will, however, be
bidirectional.

The intention behind these theoretical results to be presented is that
they could be used (i)~in network structures that combine responses computed for different
values of the filter parameters in an actual either computer or
biophysical implementation of receptive field responses, and for
(ii)~understanding the theoretical relationships between receptive
field responses, that are to be computed from different sets of filter parameters for
receptive field models based on multi-parameter scale spaces.

Specifically, in relation to the recently proposed tentative model
for variabilities of simple cells in the primary cortex of higher
mammals in Lindeberg (\citeyear{Lin25-arXiv-cov-props-review}),
the presented results provide a detailed theoretical foundation
for how to realize the alternative conceptual model in that paper,
where the receptive field responses corresponding to simple cells are not
computed for all spatial and/or temporal scales, but instead because
of the semi-group properties with the spatial and the temporal scale
parameters only computed at the finest spatial and/or temporal scales
at each image position and temporal moment. Thereby, as considered as
a possible design option for an idealized vision system, the image representations at
higher layers in the visual hierarchy would not make use of explicit
receptive field responses computed for simple cells
for each spatial and/or temporal scale, but
could instead compute equivalent receptive field responses at higher
levels in the visual hierarchy,
from the responses of the explicitly computed responses of simple
cells as only implemented at the finest spatial and/or temporal
scales.

Additionally, if an idealized vision system would be based on computing
receptive field responses as corresponding to the output from simple
cells, then the relationships between receptive field responses for
different values of the filter parameters constitute a theoretical
foundation for how such responses could be computed in a hierarchical
network structure, where receptive field responses are computed in a
way that constitutes an extension to the cascade smoothing properties
over the spatial and/or the temporal scale parameters, to enable a
computationally more efficient implementation.

Although we will in the following theoretical analysis not make
use of any explicit Lie group or Lie algebra formalisms
in the derivations of the relationships to be
presented between
receptive field responses for different parameter settings,
the derived relations will reveal relationships very closely related
to both the notions of differential Lie group
structure and macroscopic Lie algebra structures
of the corresponding representations for receptive field responses
computed for different values of the filter parameters.
An important distinction, however, is that the evolution properties over some of
the parameters will correspond to unidirectional semi-groups instead of
bidirectional groups,
because the evolution properties of those parameters can only be
performed in one direction.

\subsection{Structure of this article}

The presentation is organized as follows:
After describing relations to previous work in
Section~\ref{sec-rel-work},
Section~\ref{sec-gen-gauss-der-model} starts by reviewing the
generalized Gaussian derivative model for visual receptive fields,
including its covariance properties under important classes of
geometric image transformations in terms of
(i)~spatial scaling transformations,
(ii)~spatial affine transformations,
(iii)~Galilean transformations and
(iv)~temporal scaling transformations.
From the closedness properties of the receptive field responses under
these classes of geometric image transformation, we specifically
explain why the aim of being able to match the receptive field responses
computed under different viewing conditions leads to the notion of
multi-parameter filter banks of receptive fields, which motivates this
study of theoretically showing how receptive field responses computed
for different values of the filter parameters can be related.
This section also defines the notation and the terminology in the way
as it will be used in the forthcoming more technical sections.

Section~\ref{sec-inf-rels-non-caus-rfs} then begins the new
theoretical work by deriving infinitesimal relationships between
receptive field responses for the subclass of receptive field models, that
are based on spatial or spatio-temporal smoothing solely based on
Gaussian kernels over image space or joint space-time.
Section~\ref{sec-macroscop-rels} then continues by deriving
macroscopic  relations between filter responses computed for different
sets of filter parameters, based on either purely spatial or joint
spatio-temporal cascade smoothing properties of the corresponding
receptive field representations, defined from spatial or
spatio-temporal smoothing based on solely using
Gaussian smoothing kernels over image space or joint space-time.

Section~\ref{sec-inf-macro-rels-time-caus-spat-temp} then complements
with a set of less complete results for the time-causal theory of
visual receptive fields, where the previous use of non-causal
smoothing over the temporal domain is replaced by smoothing with the
time-causal limit kernel aimed at real-world image data, for which the
future cannot be accessed.
Finally, Section~\ref{sec-summ-disc} concludes with a summary and discussion.

\section{Relations to previous work}
\label{sec-rel-work}

Concerning theoretical modelling of visual receptive fields, the regular
Gaussian derivative model was initially proposed by
Koenderink and van Doorn (\citeyear{Koe84,KoeDoo87-BC,KoeDoo92-PAMI})
and used for modelling biological receptive fields by
Young (\citeyear{You87-SV}) and his co-workers in
Young {\em et al.\/} (\citeyear{YouLesMey01-SV,YouLes01-SV}).
This regular Gaussian derivative model
has also been used as a component in more developed models of biological vision by
Lowe (\citeyear{Low00-BIO}),
May and Georgeson (\citeyear{MayGeo05-VisRes}),
Hesse and Georgeson (\citeyear{HesGeo05-VisRes}),
Georgeson  {\em et al.\/}\ (\citeyear{GeoMayFreHes07-JVis}),
Wallis and Georgeson (\citeyear{WalGeo09-VisRes}),
Hansen and Neumann (\citeyear{HanNeu09-JVis}),
Wang and Spratling (\citeyear{WanSpra16-CognComp}),
Pei {\em et al.\/}\ (\citeyear{PeiGaoHaoQiaAi16-NeurRegen}),
Ghodrati {\em et al.\/}\ (\citeyear{GhoKhaLeh17-ProNeurobiol}),
Kristensen and Sandberg (\citeyear{KriSan21-SciRep}),
Abballe and Asari (\citeyear{AbbAsa22-PONE}),
Ruslim {\em et al.\/}\ (\citeyear{RusBurLia23-bioRxiv}) and
Wendt and Faul (\citeyear{WenFay24-JVis}).

The generalization into receptive field families based on
multi-parameter scale spaces goes back to the formulation
of affine Gaussian scale space representation
in Lindeberg and G{\aa}rding (\citeyear{LG96-IVC})
and early formulations of discrete
scale-space representations over either purely spatial or joint
spatio-temporal domains in Lindeberg
(\citeyear{Lin97-ICSSTCV,CVAP257}).
These notions were then extended into the generalized Gaussian derivative model for
either purely spatial or joint spatio-temporal domains in 
Lindeberg (\citeyear{Lin10-JMIV,Lin13-BICY,Lin21-Heliyon}) and
with additional extensions concerning provable covariance
properties of receptive field families in
Lindeberg (\citeyear{Lin23-FrontCompNeuroSci,Lin25-JMIV, Lin25-arXiv-cov-props-review}).

The variabilities in receptive field shapes genererated by the
affine Gaussian scale space have been used for
computing more accurate estimates of local surface orientation from
monocular and binocular cues by
Lindeberg and G{\aa}rding (\citeyear{LG96-IVC}) and
Rodr{\'\i}guez {\em et al.\/} (\citeyear{RodDelMor18-SIAM}),
for computing affine invariant image features for
image matching under wide baselines
by Baumberg (\citeyear{Bau00-CVPR}),
Mikolajczyk and Schmid (\citeyear{MikSch04-IJCV}),
Mikolajczyk {\em et al.\/} (\citeyear{MikTuySchZisMatSchKadGoo05-IJCV}),
Tuytelaars and van Gool (\citeyear{TuyGoo04-IJCV}),
Lazebnik {\em et al.\/}\ (\citeyear{LazSchPon05-PAMI}),
Rothganger {\em et al.\/}\ (\citeyear{RotLazSchPon06-IJCV,RotLazSchPon07-PAMI}),
Lia {\em et al.\/}\ (\citeyear{LiaLiuHui13-PRL}),
Eichhardt and Chetverikov (\citeyear{EicChe18-ECCV}) and
Dai {\em et al.\/} (\citeyear{DaiJinZhaNgu20-IP}),
for performing affine invariant segmentation 
by Ball\-ester and Gonz{\'a}lez (\citeyear{BalGon98-JMIV}),
for constructing affine covariant SIFT descriptors 
by Morel and Yu (\citeyear{MorGuo09-SIAM-JIS}),
Yu and Morel (\citeyear{YuMor09-ASSP}) and
Sadek {\em et al.\/}\ (\citeyear{SadConMeiBalCas12-SIM-JIS}),
for modelling receptive fields in biological vision
by Lindeberg (\citeyear{Lin13-BICY,Lin21-Heliyon}),
for affine invariant tracking
by Giannarou {\em et al.\/}\ (\citeyear{GiaVisYan13-PAMI}),
and for formulating affine covariant metrics
by Fedorov {\em et al.\/} (\citeyear{FedAriSadFacBal15-SIAM}).
This rich variety of application domains clearly demonstrates how the use of
variabilities in the shapes of the receptive fields, as implied by
covariant receptive field representations (here with regard to spatial
affine transformations), can be used for improving the performance of
visual operations.

For modelling spatial receptive fields in vision, also Gabor models
have been used by Marcelja (\citeyear{Mar80-JOSA}),
Jones and Palmer (\citeyear{JonPal87a,JonPal87b}),
Ringach (\citeyear{Rin01-JNeuroPhys,Rin04-JPhys}),
Serre {\em et al.\/} (\citeyear{SerWolBilRiePog07-PAMI}),
Baspinar {\em et al.\/} (\citeyear{BarCitSanSar12-JPhysParis,BasCitSar18-JMIV,BasSarCit20-MathNeuroSci}),
De and Horwitz (\citeyear{DeHor21-JNPhys}) and others.
The use of Lie symmetries for deriving receptive field shapes based on Gabor filters under
the SE(2) group has been studied by
Citti and Sarti {\em et al.\/} (\citeyear{CitSar06-JMIV})
and under the 2-D affine group by
Sarti {\em et al.\/} (\citeyear{SarCitPet08-BICY}).
Modelling of spatio-temporal receptive fields using Gabor
functions has been proposed by
Cocci {\em et al.\/} (\citeyear{CocBarSar11-JOSA}) and
Barbieri {\em et al.\/} (\citeyear{BarCitCocSar14-JMIV}).

Both the complex-valued Gabor functions used in the Gabor model
and the real-valued Gaussian functions underlying the Gaussian
derivative model minimize the uncertainty relation
(Barbieri {\em et al.\/} \citeyear{BarCitSanSar12-JPhysParis},
Lindeberg \citeyear{Lin13-ImPhys}).

For handling the influence of geometric image transformations in terms
of spatial scaling transformations in deep networks, provably scale
covariant (also referred to as scale equivariant) network architectures have been developed by
 Worrall and Welling (\citeyear{WorWel19-NeuroIPS}),
Bekkers (\citeyear{Bek20-ICLR}),
Sosnovik {\em et al.\/}
(\citeyear{SosSzmSme20-ICLR,SosMosSme21-BMVC}),
Lindeberg (\citeyear{Lin20-JMIV,Lin22-JMIV}),
Jansson and Lindeberg (\citeyear{JanLin21-ICPR,JanLin22-JMIV}),
Zhu {\em et al.\/} (\citeyear{ZhuQiuCalSapChe22-JMLR}),
Penaud {\em et al.\/} (\citeyear{PenVelAng22-ICIP}),
Sangalli {\em et al.\/} (\citeyear{SanBluVelAng22-BMVC}),
Zhan {\em et al.\/} (\citeyear{ZhaSunLi22-ICCRE}),
Yang  {\em et al.\/} (\citeyear{YanDasMah23-arXiv}),
Wimmer {\em et al.\/}  (\citeyear{WimGolDanMaiCre23-arXiv}),
Barisin  {\em et al.\/}
(\citeyear{BarSchRed24-JMIV,BarAngSchRed24-SIIMS})
and Perzanowski and Lindeberg (\citeyear{PerLin25-JMIV}).
These classes of deep networks have specifically been demonstrated to
handle image data subject to scaling variations in a much more robust
way compared to non-covariant deep networks.

With the combined treatment of the 4 main classes of geometric image
transformations considered in this paper,
involving (i)~spatial scaling transformations,
(ii)~spatial affine transformations, (iii)~Galilean transformations
and (iv)~temporal scaling transformations, the underlying aims of this
work bear close similarities to approaches in the area of geometric
deep learning (Bronstein {\em et al.\/} \citeyear{BroBruCohVel21-arXiv},
Gerken {\em et al.\/} \citeyear{GerAroCarLinOhlPetPer23-AIRev}), which
also consider visual processing operations that are covariant under
wider classes of geometric image transformations.

For an introductory overview of Lie groups and Lie algebras,
to which the presented relations between receptive field responses for
different values of the filter parameters will be closely related to,
see {\em e.g.\/}\ Hall (\citeyear{Hal15-book}).
For an extensive treatment of one-parameter semi-groups and their
infinitesimal generators, see Hille and Phillips (\citeyear{HilPhi57}).
For further relations between semi-groups and partial differential equations
as well as extensions of the theory to two-parameter semi-groups,
to which the presented methodology will also bear very close relations,
see Pazy (\citeyear{Paz83-Book}), Goldstein (\citeyear{Gol85-book}) and
Al-Sharif and Khalil (\citeyear{ShaKha04-ApplMathComp}).

\section{The generalized Gaussian derivative model for visual
  receptive fields}
\label{sec-gen-gauss-der-model}

\subsection{Spatial and spatio-temporal smoothing kernels}

The generalized Gaussian derivative model for visual receptive fields
(Lindeberg \citeyear{Lin21-Heliyon})
is based on:
\begin{itemize}
\item
  smoothing any purely spatial image
  $f \colon \bbbr^2 \rightarrow \bbbr$ with 
  affine Gaussian kernels%
  \footnote{In this treatment, $\bbbs_+^2$ denotes the set
    of symmetric positive definite $2 \times 2$ matrices.}
  $T \colon \bbbr^2 \times \bbbr_+ \times \bbbs_+^2 \rightarrow \bbbr$
  of the form
  \begin{equation}
    \label{eq-gauss-fcn-2D}
    T(x;\; s, \Sigma) = g(x;\; s, \Sigma)
    = \frac{1}{2 \pi \, s \sqrt{\det \Sigma}} \, e^{-x^T  \Sigma^{-1} x/2 s},
  \end{equation}
  where
  \begin{itemize}
  \item
    $x = (x_1, x_2) \in \bbbr^2$ denotes the image coordinates,
  \item
    $s \in \bbbr_+$ is the spatial scale parameter, and
  \item
    $\Sigma \in \bbbs_+^2$ is a spatial covariance matrix that specifies
    the shape of the affine Gaussian kernel, or
  \end{itemize}
\item
  smoothing any video sequence or video stream
  $f \colon \bbbr^2 \times \bbbr \rightarrow \bbbr$ 
  with joint spatio-temporal smoothing kernels
  $T \colon \bbbr^2 \times \bbbr \times \bbbr_+ \times \bbbs_+^2 \times \bbbr_+
  \times \bbbr^2 \rightarrow \bbbr$
  of the form
  \begin{equation}
    \label{eq-form-spat-temp-kernel}
    T(x, t;\; s, \Sigma, \tau, v) = g(x - v \, t;\; s, \Sigma) \, h(t;\; \tau),
  \end{equation}
  where
  \begin{itemize}
  \item
    $t \in \bbbr$ is the time variable,
  \item
    $v = (v_1, v_2)^T \in \bbbr^2$ is an image velocity,
  \item
    $\tau \in \bbbr_+$ is the temporal scale parameter, and
  \item
    $h \colon \bbbr \times \bbbr_+ \rightarrow$ is a temporal smoothing
    kernel.
  \end{itemize}
\end{itemize}
In the case of a non-causal temporal domain, {\em i.e.,\/} for
pre-recorded video, for which the future can be accessed,
the temporal smoothing kernel $h(t;\; \tau)$ in the spatio-temporal smoothing
kernel $T(x, t;\; s, \Sigma, \tau, v)$ can be chosen as a Gaussian kernel
\begin{equation}
  \label{eq-non-caus-temp-gauss}
  h(t;\; \tau) = g(t;\; \tau) = \frac{1}{\sqrt{2 \pi} \sqrt{\tau}} \, e^{-t^2/2\tau}.
\end{equation}
In the case of a time-causal domain, {\em i.e.\/}, for real-time
situations when the future cannot be accessed, the temporal smoothing
kernel $h(t;\; \tau)$ in the spatio-temporal smoothing
kernel $T(x, t;\; s, \Sigma, \tau, v)$ can be chosen as the time-causal limit kernel
(Lindeberg \citeyear{Lin16-JMIV} Section~5;
 Lindeberg \citeyear{Lin23-BICY} Section~3)
\begin{equation}
  \label{eq-time-caus-lim-kern}
  h(t;\; \tau) = \psi(t;\; \tau, c),
\end{equation}
characterized by having a Fourier transform of the form
\begin{equation}
  \label{eq-FT-comp-kern-log-distr-limit}
     \hat{\Psi}(\omega;\; \tau, c) 
     = \prod_{k=1}^{\infty} \frac{1}{1 + i \, c^{-k} \sqrt{c^2-1} \, \sqrt{\tau} \, \omega},
\end{equation}
where $c > 1$ is a distribution parameter describing the ratio
between adjacent discrete temporal scale levels
\begin{equation}
  \label{eq-disc-temp-sc-limit-kern}
  \tau_k = \tau_0 \, c^{2k}
\end{equation}
for some initial temporal scale level $\tau_0 \in \bbbr_+$.
For common purposes of implementation,  the
distribution parameter $c$ is often
chosen as $c = \sqrt{2}$ or $c = 2$.

In fact, it is shown in an axiomatic way in Lindeberg (\citeyear{Lin10-JMIV}), that the
forms of the spatial smoothing kernel (\ref{eq-gauss-fcn-2D}) and
the spatio-temporal smoothing kernel (\ref{eq-form-spat-temp-kernel}) are
uniquely determined, given natural symmetry requirements on a visual front end, with
conceptual extensions of those ideas to a time-causal temporal domain
in Lindeberg (\citeyear{Lin16-JMIV,Lin21-Heliyon}).
The underlying symmetry requirements used for the axiomatic
derivations are based on symmetry properties of the
environment, in combination with internal consistency requirements, to
guarantee theoretically well-founded treatment of image structures
over multiple spatial and temporal scales,
see Lindeberg (\citeyear{Lin10-JMIV,Lin21-Heliyon}) for further details.

\begin{figure*}[hbtp]
  \begin{center}
    \begin{tabular}{ccccccc}
      & $\varphi = 0$ & $\varphi = \pi/6$ & $\varphi = \pi/3$
      & $\varphi = \pi/2$ & $\varphi = 2\pi/3$ & $\varphi = 5\pi/6$
      \\
      $\scriptsize{\sigma_1 = 4, \sigma_2 = 4}$
      & \includegraphics[width=0.12\textwidth]{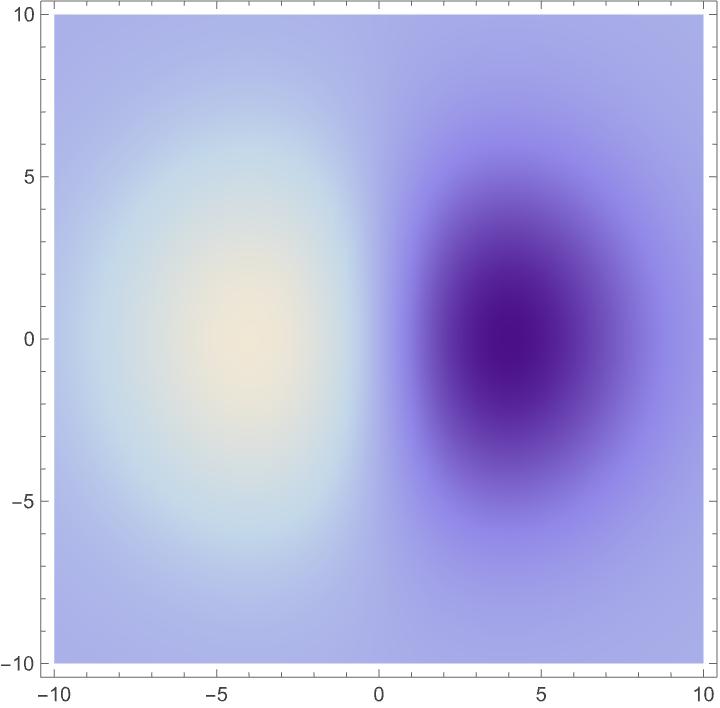}
      & \includegraphics[width=0.12\textwidth]{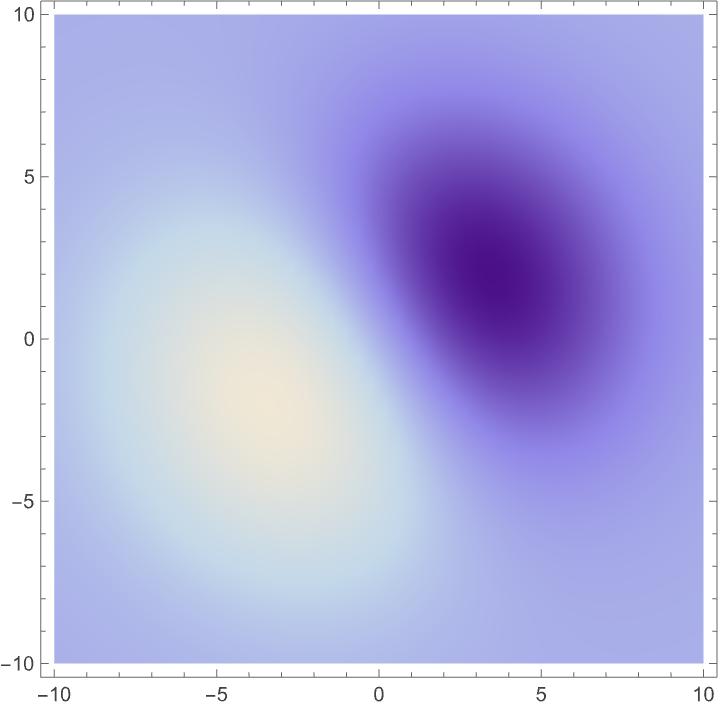}
      & \includegraphics[width=0.12\textwidth]{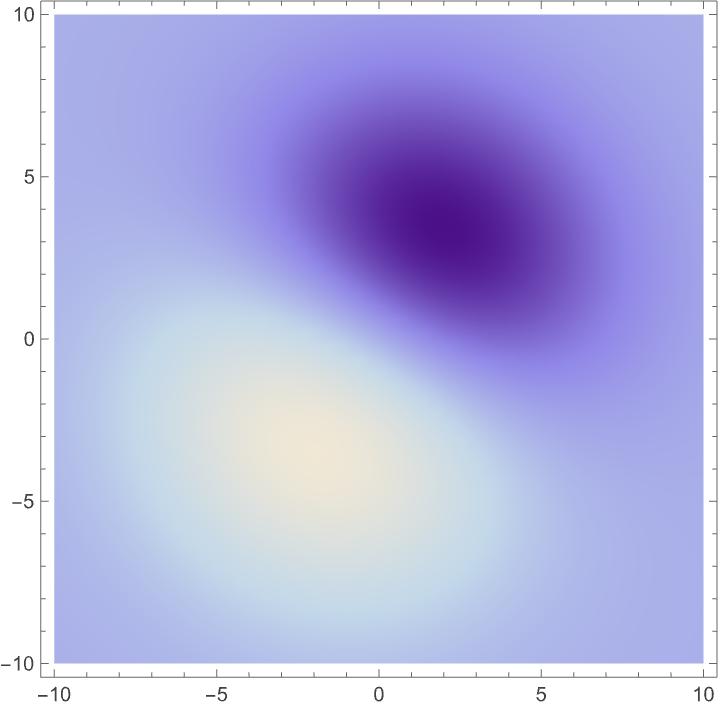}
      & \includegraphics[width=0.12\textwidth]{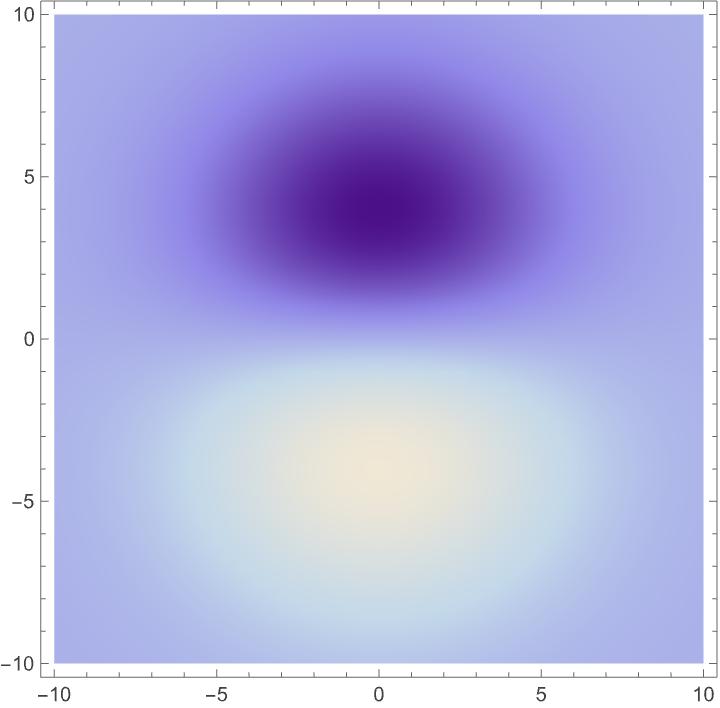}
      &   \includegraphics[width=0.12\textwidth]{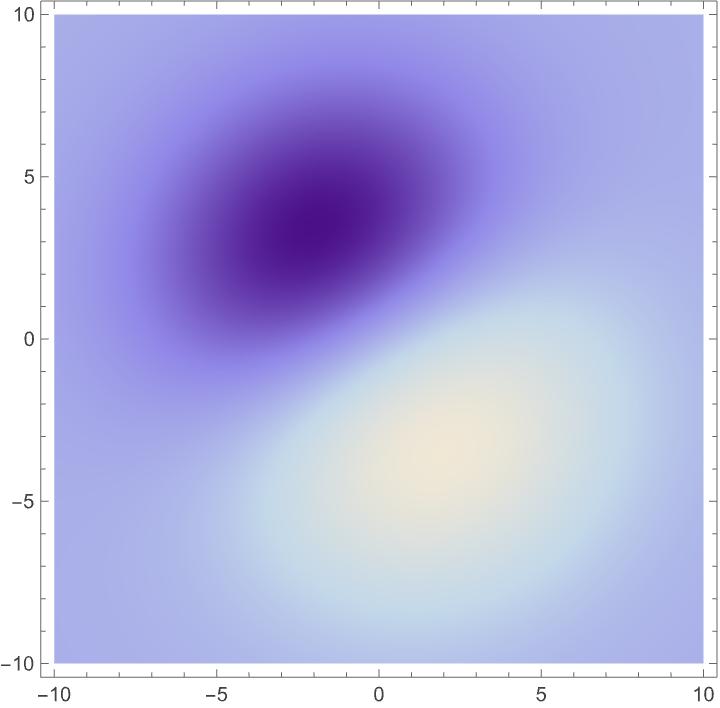}
      & \includegraphics[width=0.12\textwidth]{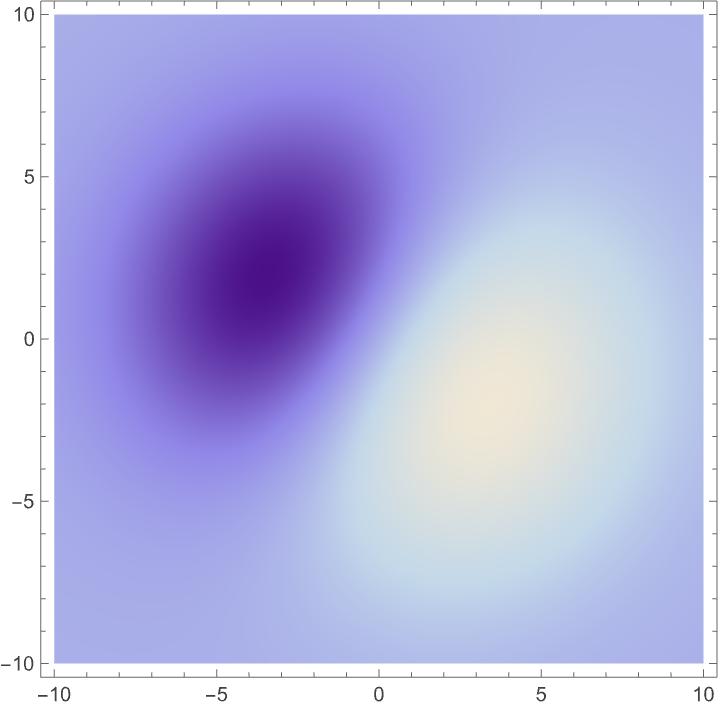}
      \\
       $\scriptsize{\sigma_1 = 2, \sigma_2 = 4}$
      & \includegraphics[width=0.12\textwidth]{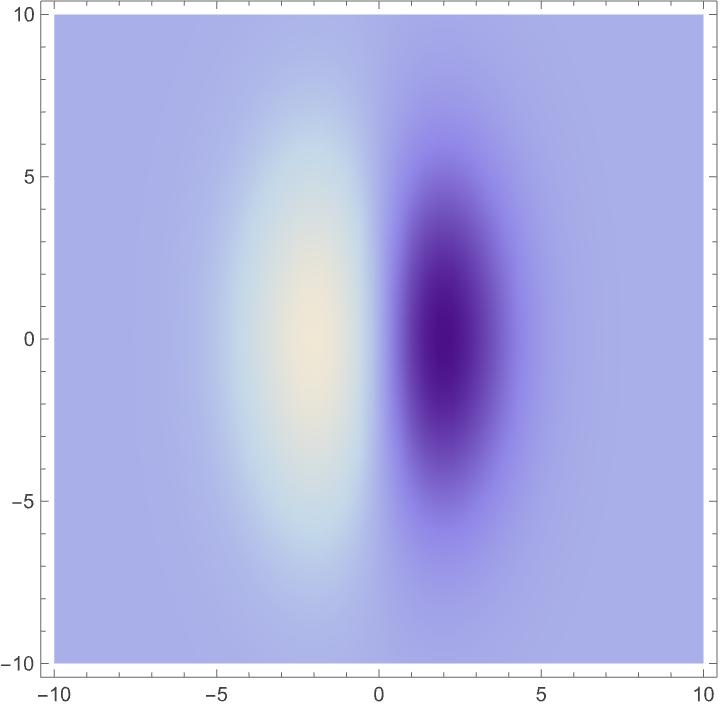}
      & \includegraphics[width=0.12\textwidth]{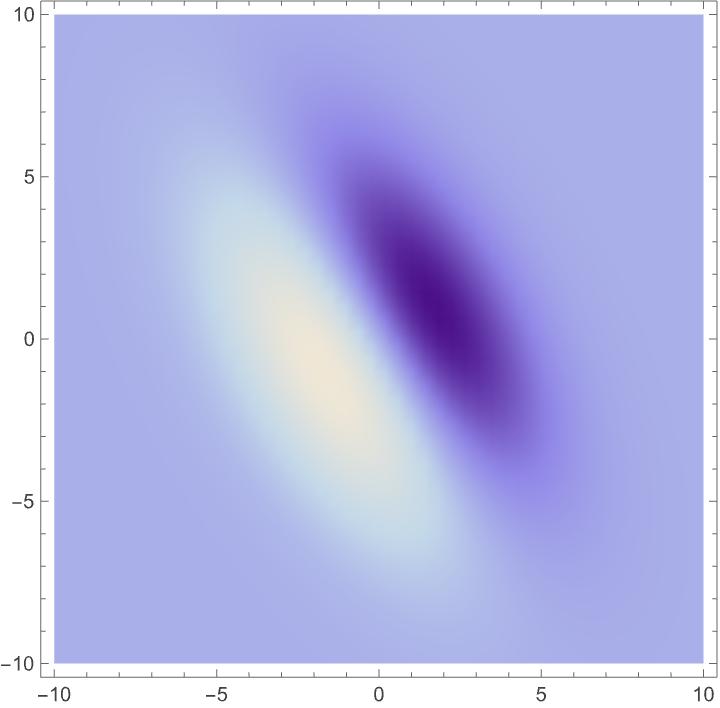}
      & \includegraphics[width=0.12\textwidth]{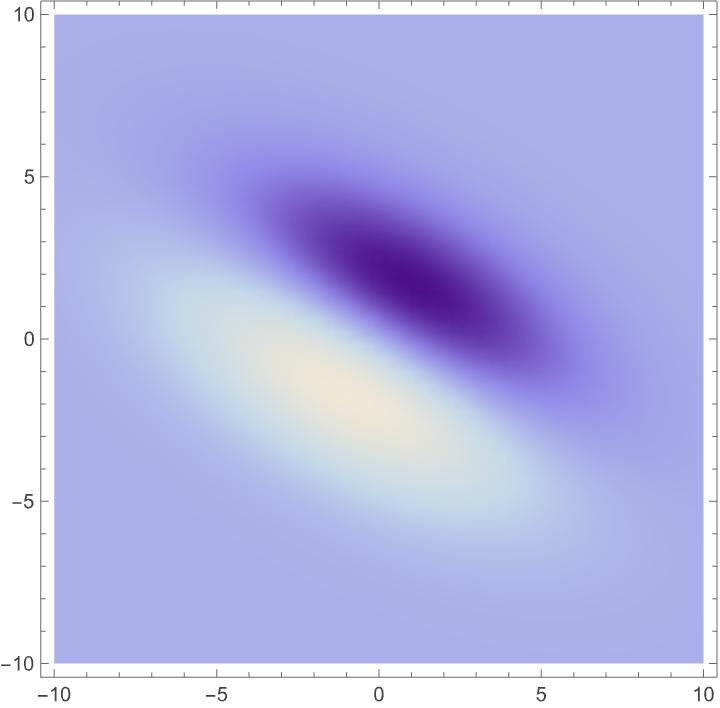}
      & \includegraphics[width=0.12\textwidth]{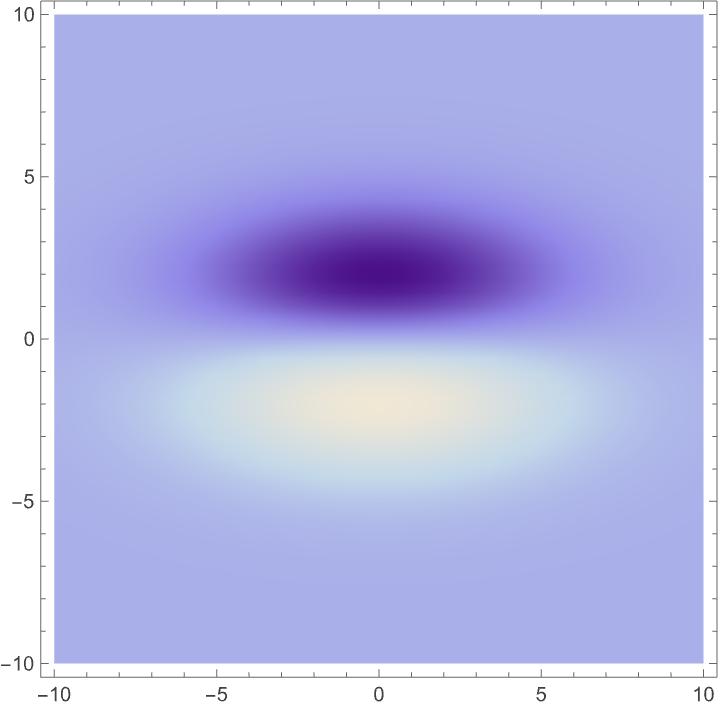}
      &   \includegraphics[width=0.12\textwidth]{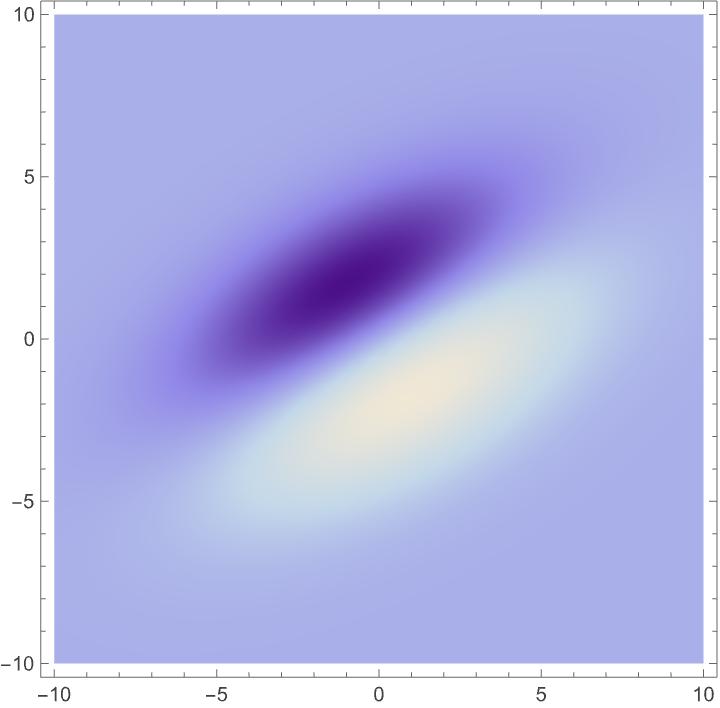}
      & \includegraphics[width=0.12\textwidth]{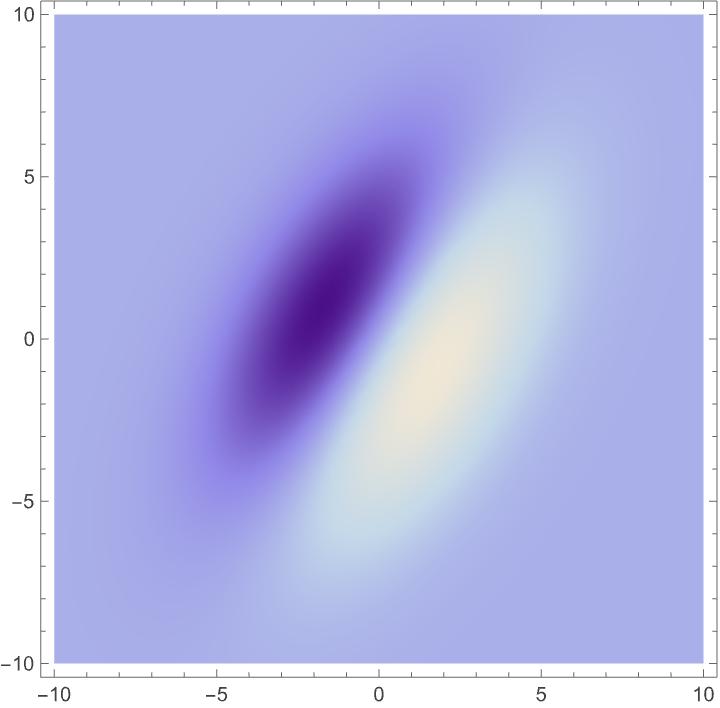}
      \\
      $\scriptsize{\sigma_1 = 2, \sigma_2 = 2}$
      & \includegraphics[width=0.12\textwidth]{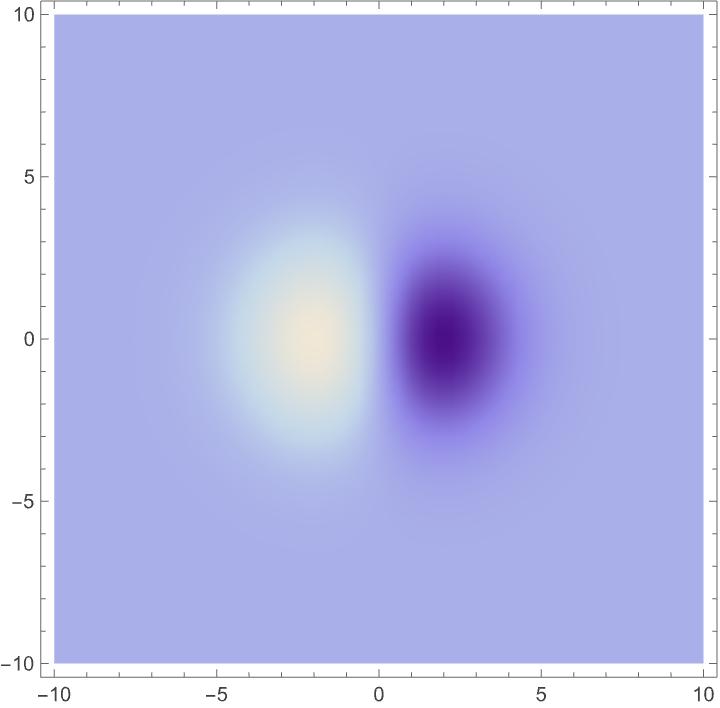}
      & \includegraphics[width=0.12\textwidth]{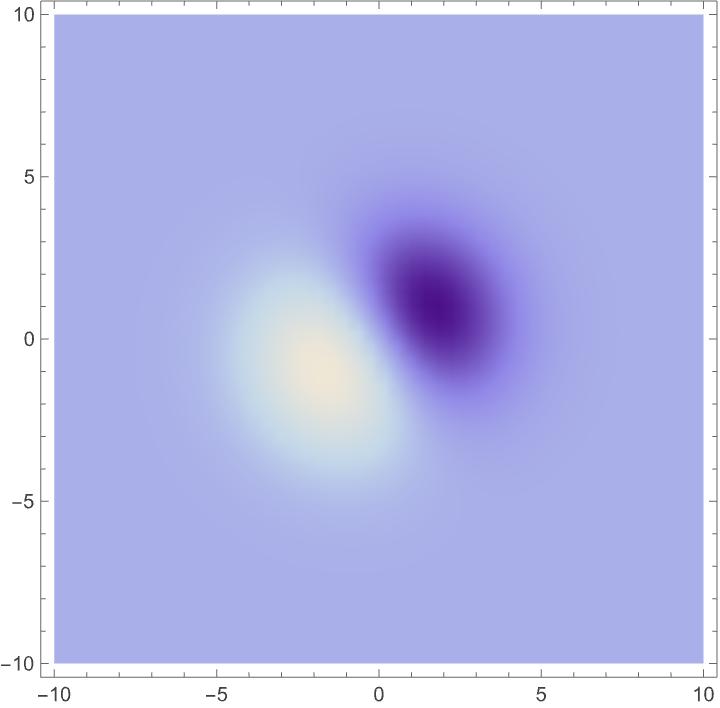}
      & \includegraphics[width=0.12\textwidth]{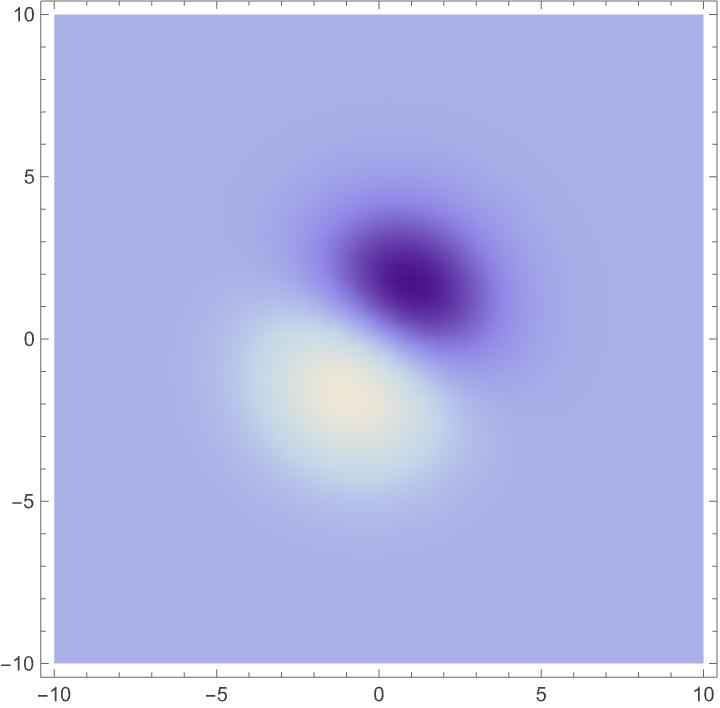}
      & \includegraphics[width=0.12\textwidth]{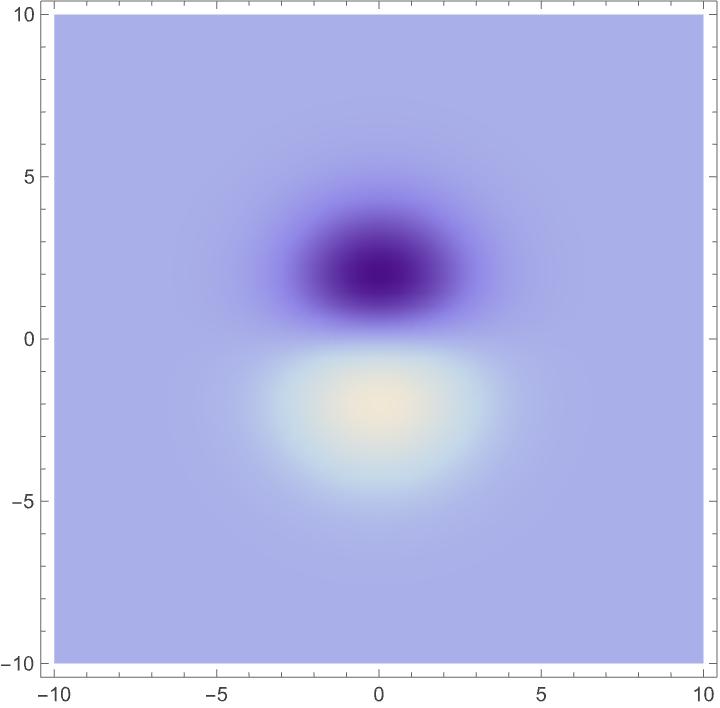}
      &   \includegraphics[width=0.12\textwidth]{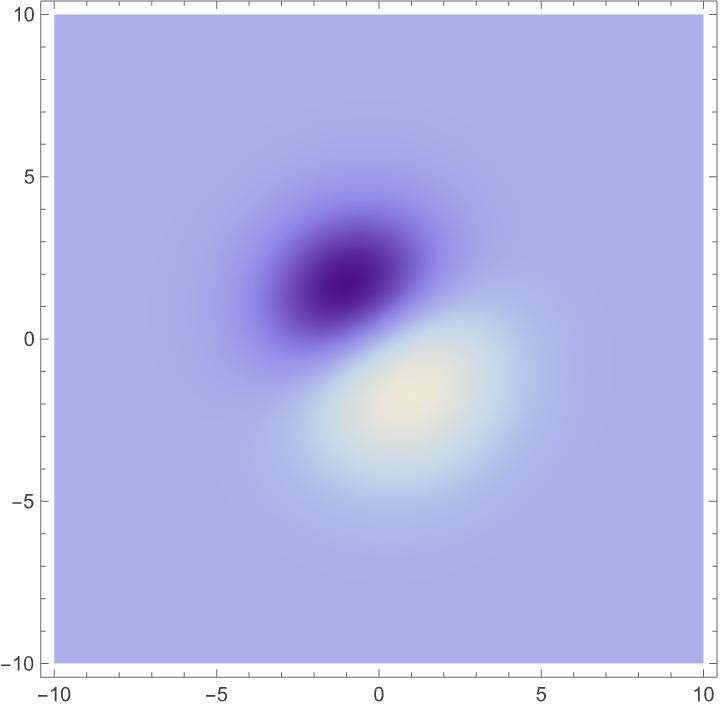}
      & \includegraphics[width=0.12\textwidth]{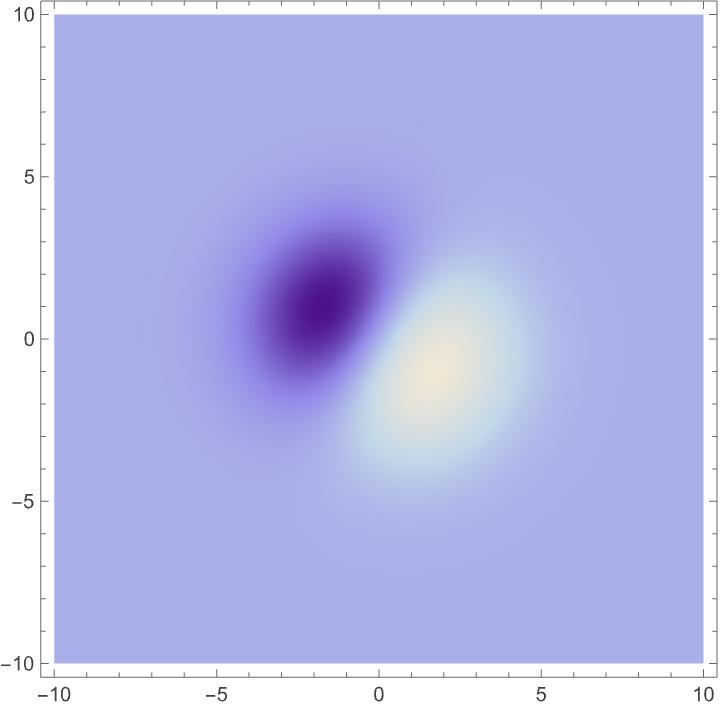}
      \\
      $\scriptsize{\sigma_1 = 1, \sigma_2 = 2}$
      & \includegraphics[width=0.12\textwidth]{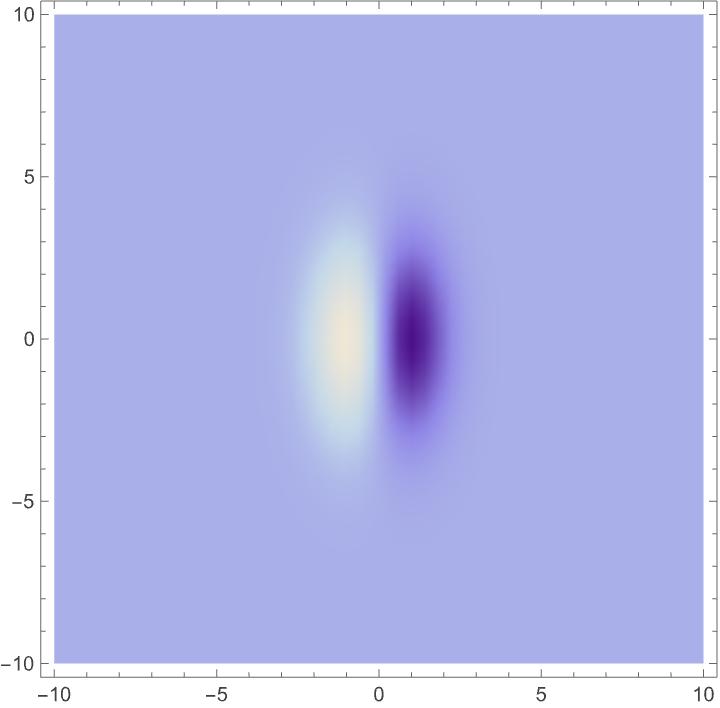}
      & \includegraphics[width=0.12\textwidth]{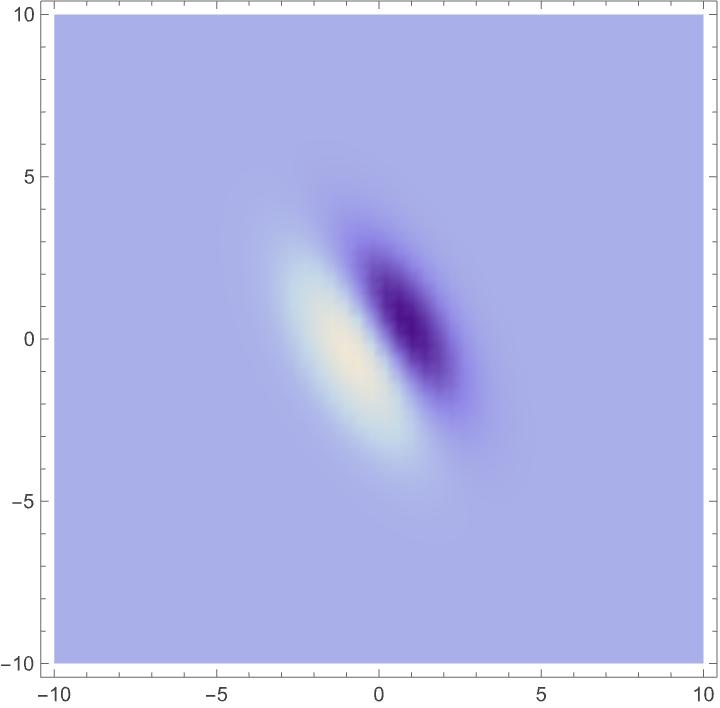}
      & \includegraphics[width=0.12\textwidth]{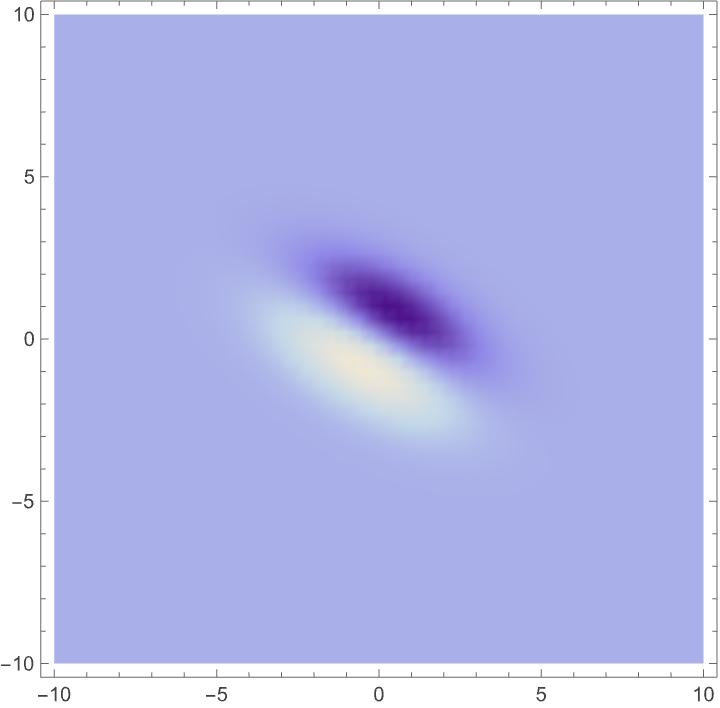}
      & \includegraphics[width=0.12\textwidth]{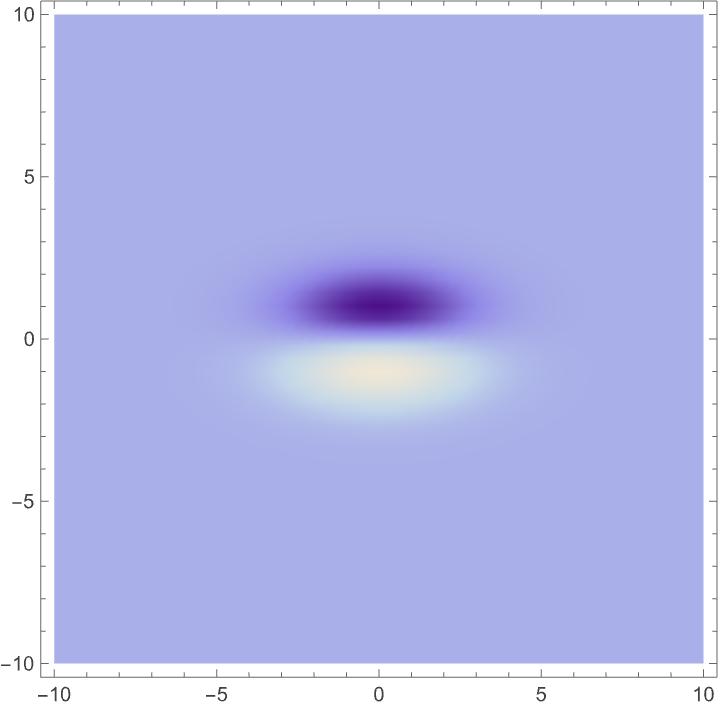}
      &   \includegraphics[width=0.12\textwidth]{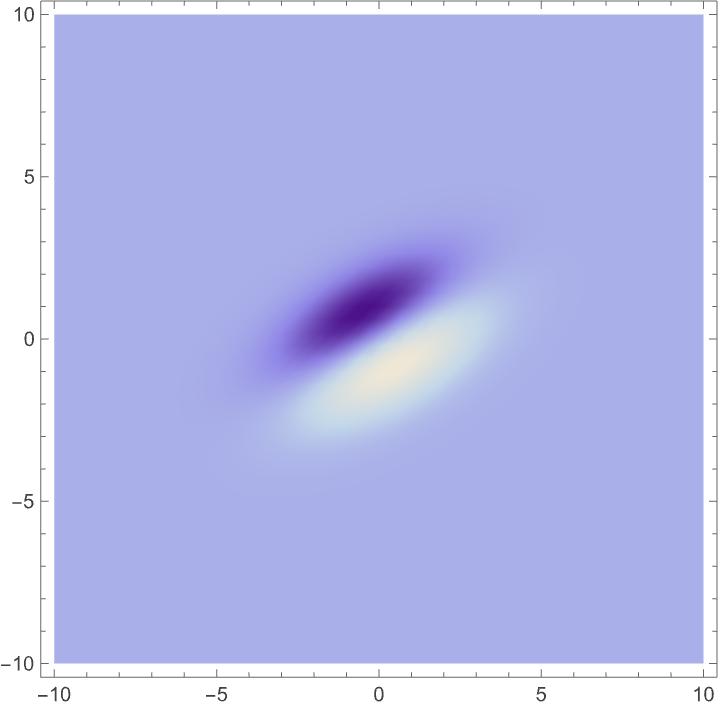}
      & \includegraphics[width=0.12\textwidth]{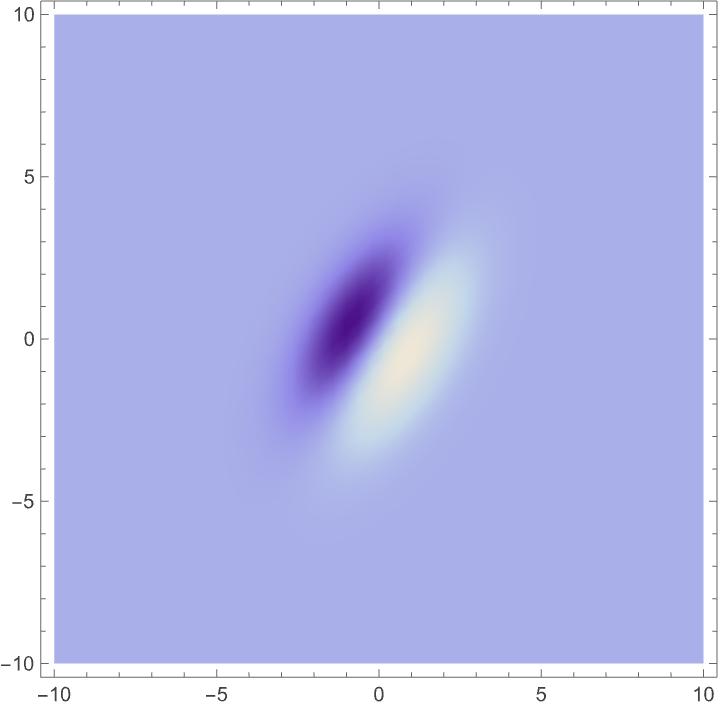}
      \\
       $\scriptsize{\sigma_1 = 1, \sigma_2 = 1}$
      & \includegraphics[width=0.12\textwidth]{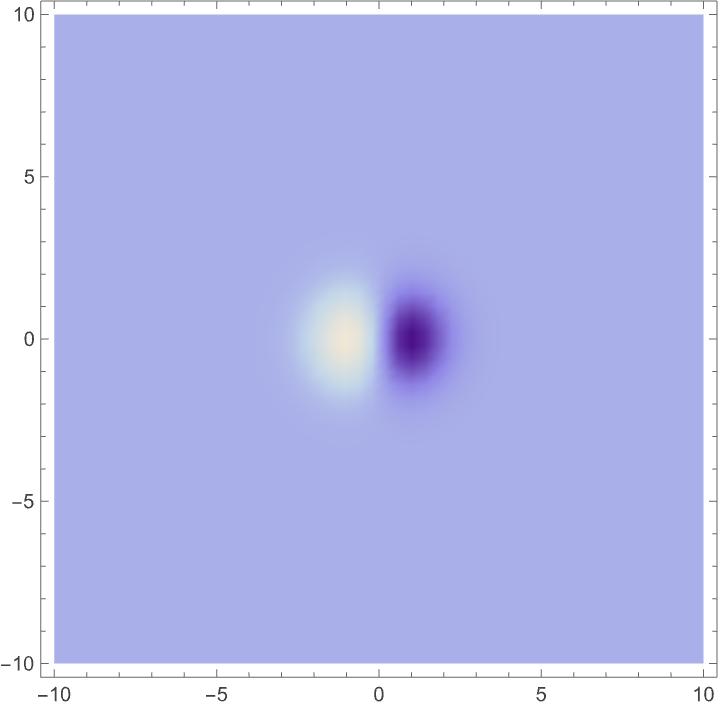}
      & \includegraphics[width=0.12\textwidth]{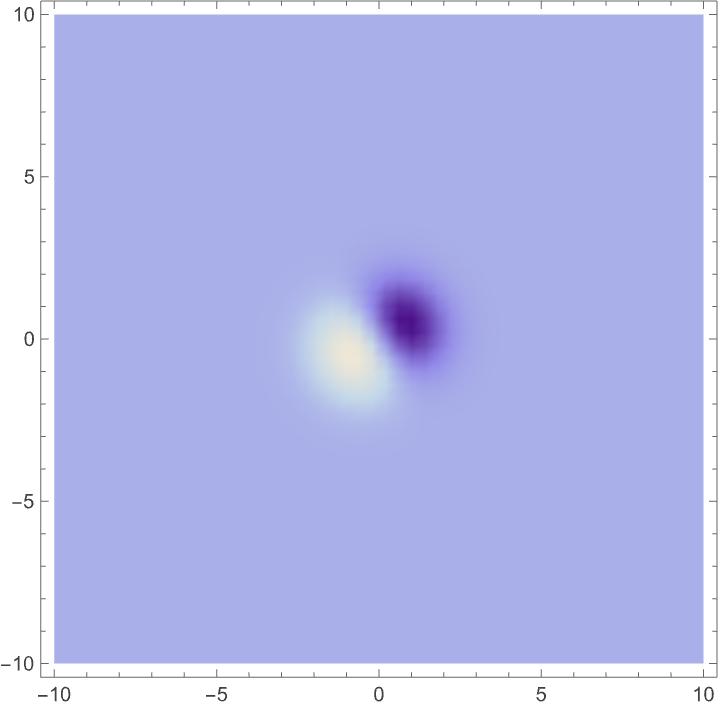}
      & \includegraphics[width=0.12\textwidth]{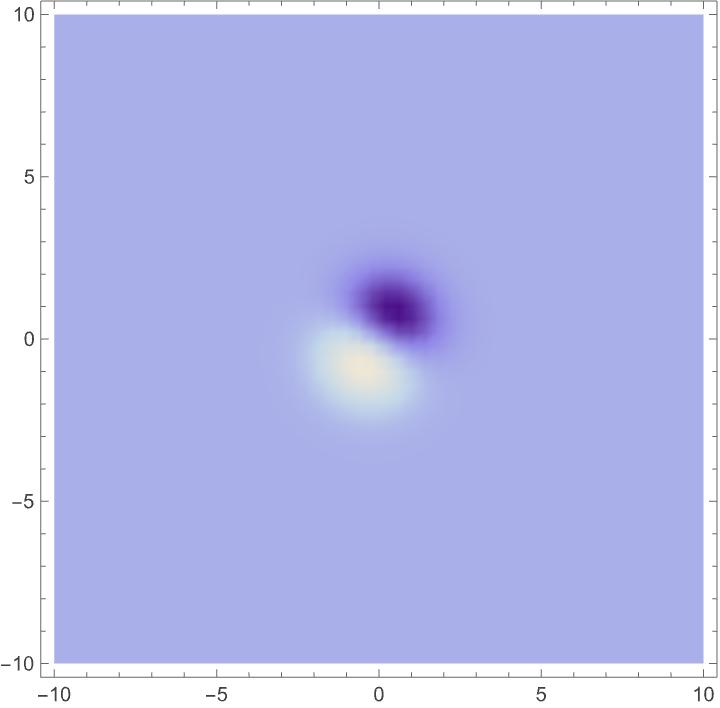}
      & \includegraphics[width=0.12\textwidth]{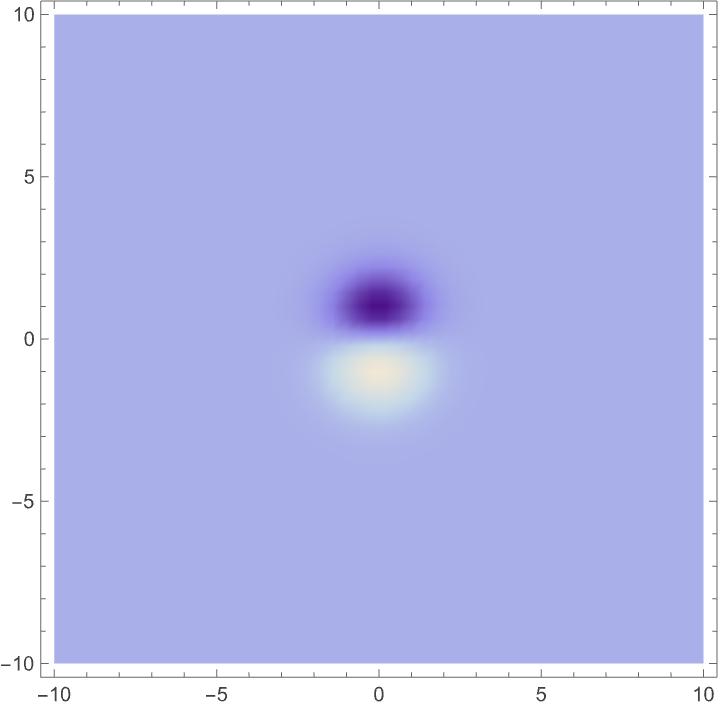}
      &   \includegraphics[width=0.12\textwidth]{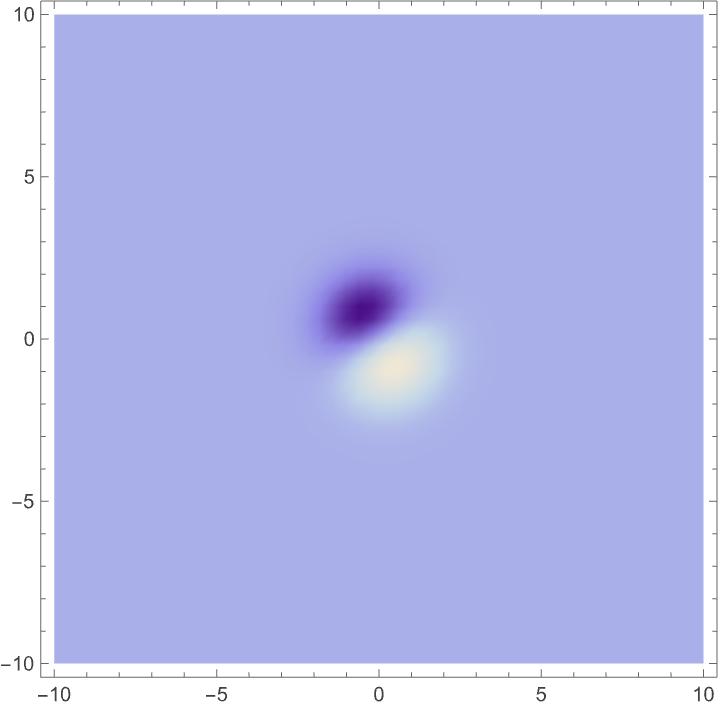}
      & \includegraphics[width=0.12\textwidth]{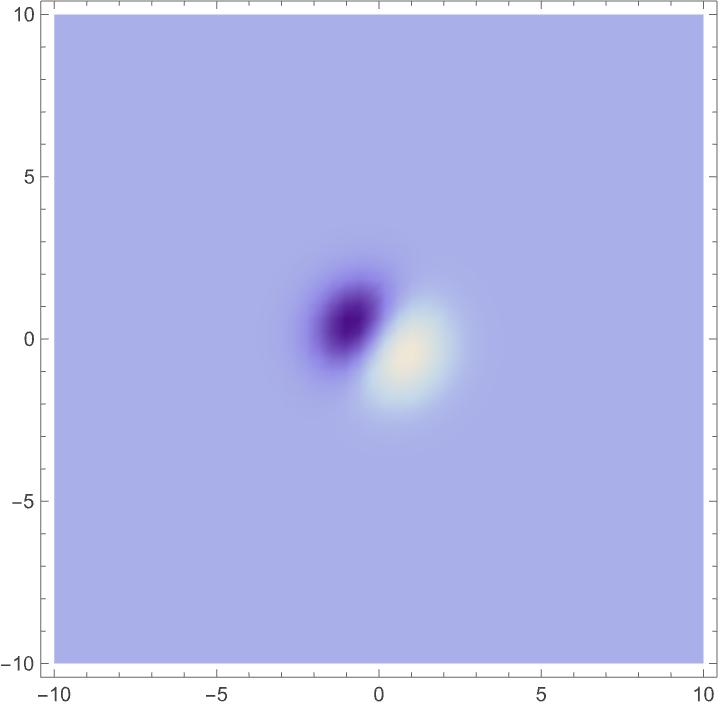}
      \\
      $\scriptsize{\sigma_1 = 4, \sigma_2 = 4}$
      & \includegraphics[width=0.12\textwidth]{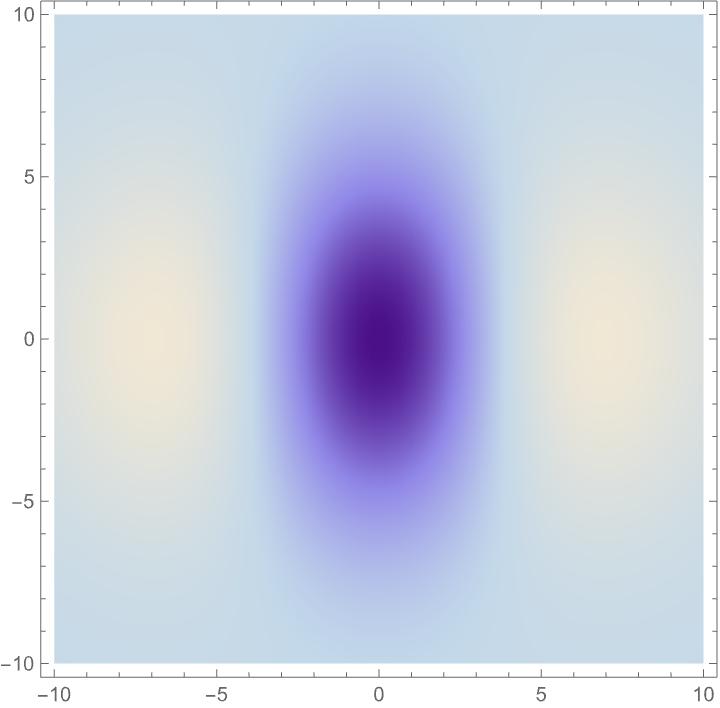}
      & \includegraphics[width=0.12\textwidth]{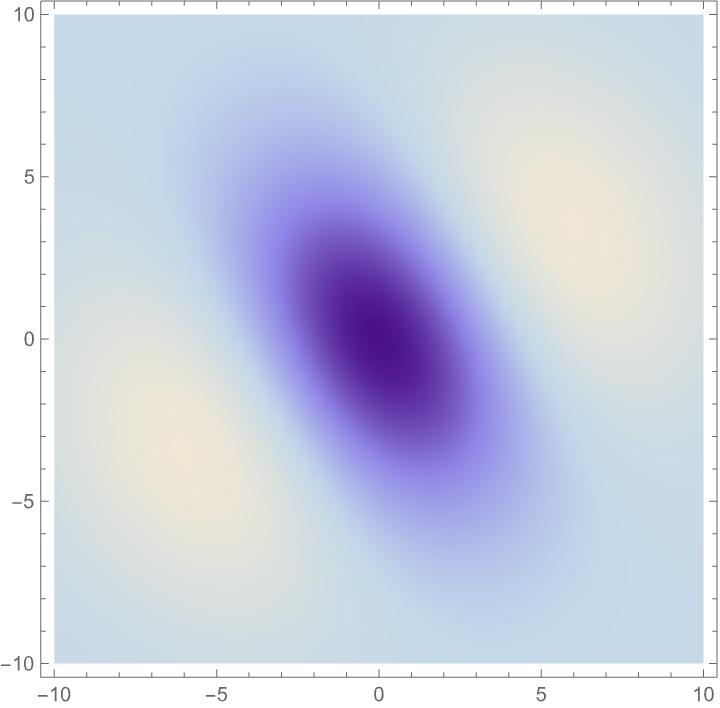}
      & \includegraphics[width=0.12\textwidth]{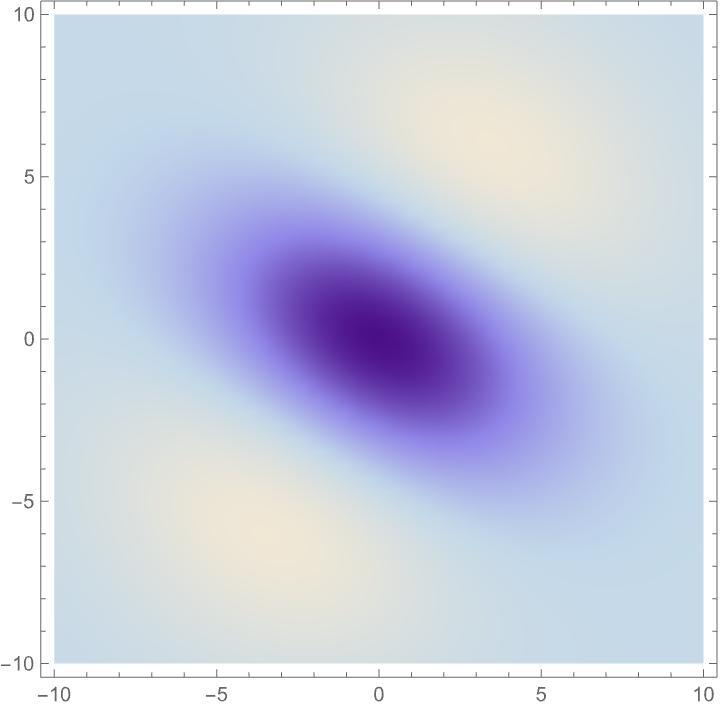}
      & \includegraphics[width=0.12\textwidth]{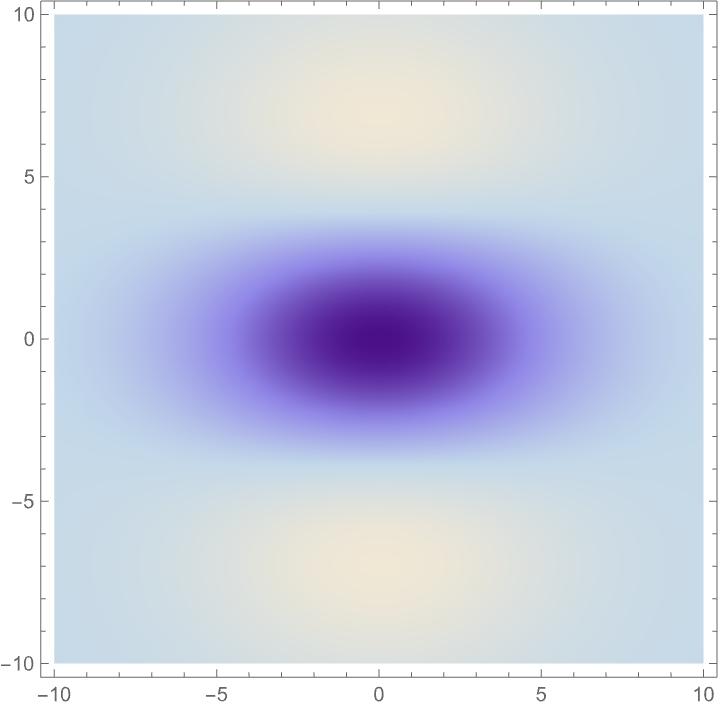}
      &   \includegraphics[width=0.12\textwidth]{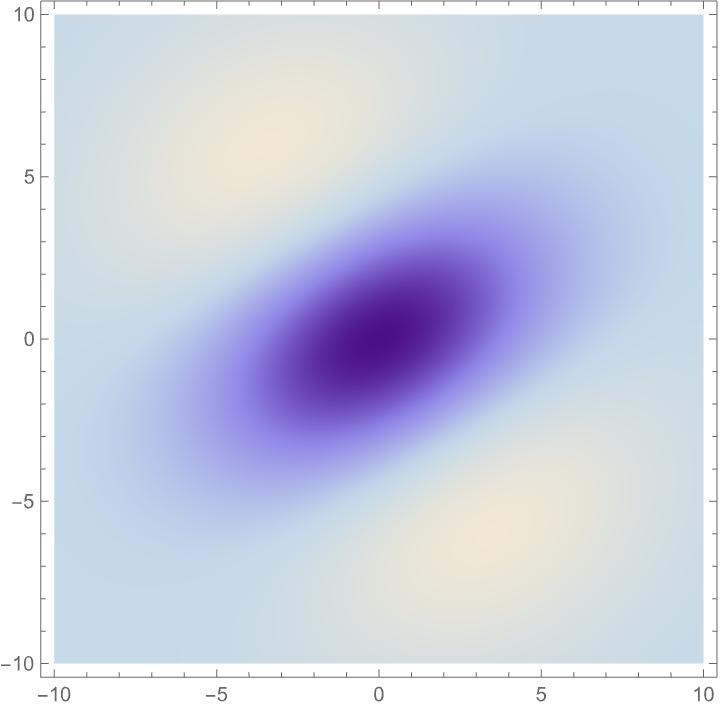}
      & \includegraphics[width=0.12\textwidth]{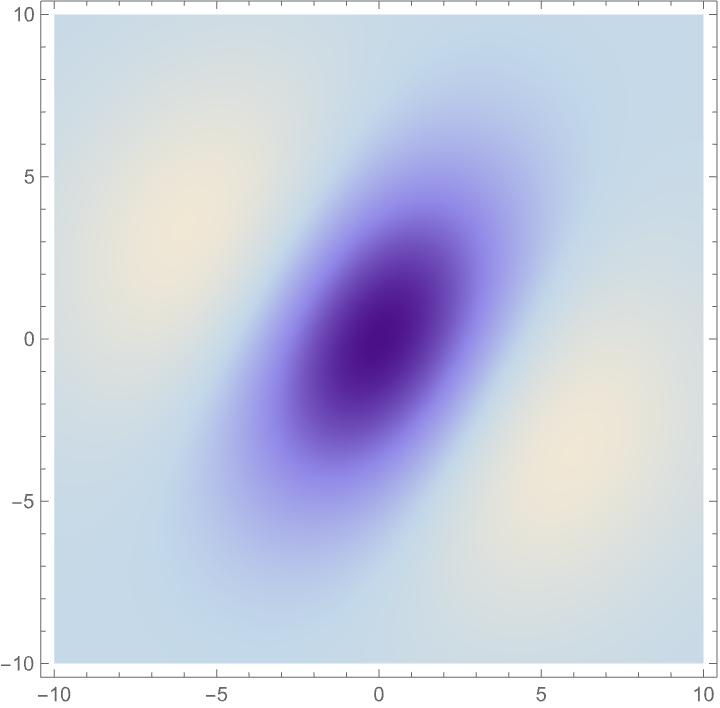}
      \\
       $\scriptsize{\sigma_1 = 2, \sigma_2 = 4}$
      & \includegraphics[width=0.12\textwidth]{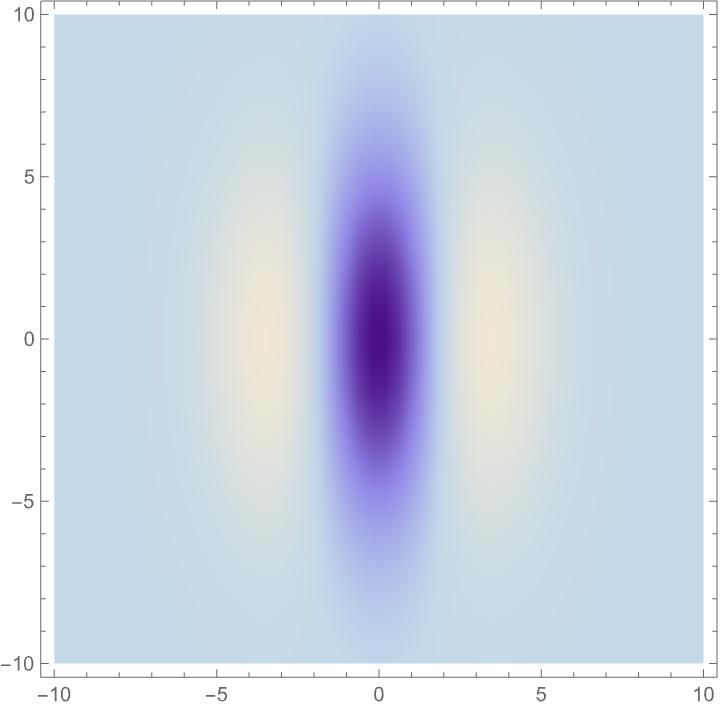}
      & \includegraphics[width=0.12\textwidth]{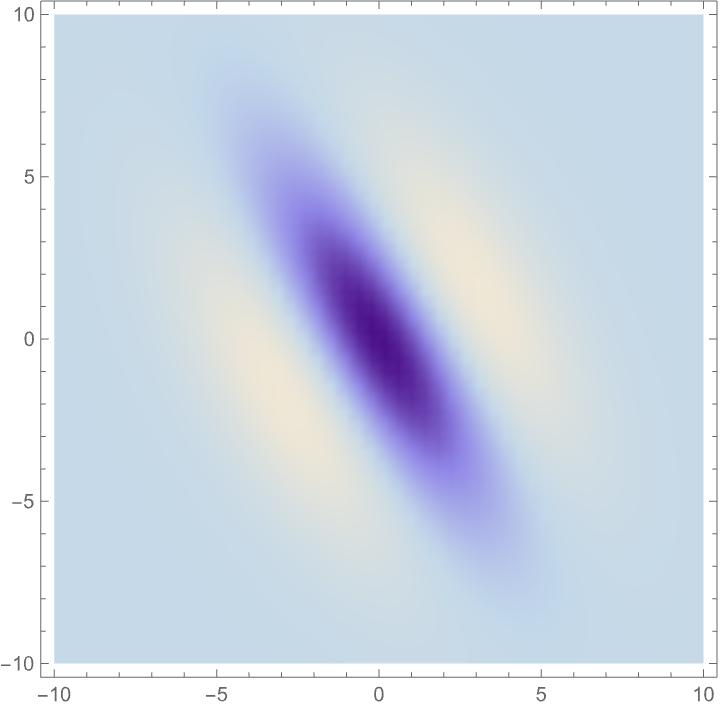}
      & \includegraphics[width=0.12\textwidth]{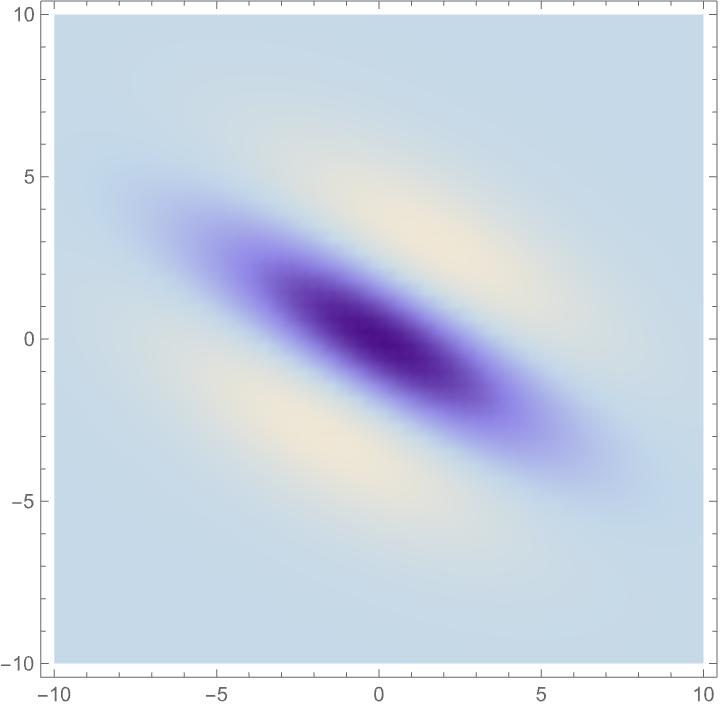}
      & \includegraphics[width=0.12\textwidth]{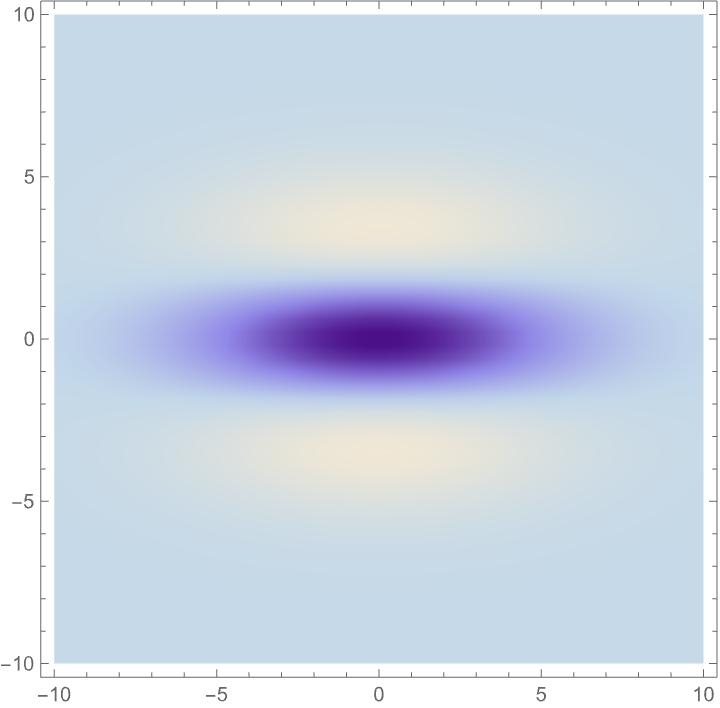}
      &   \includegraphics[width=0.12\textwidth]{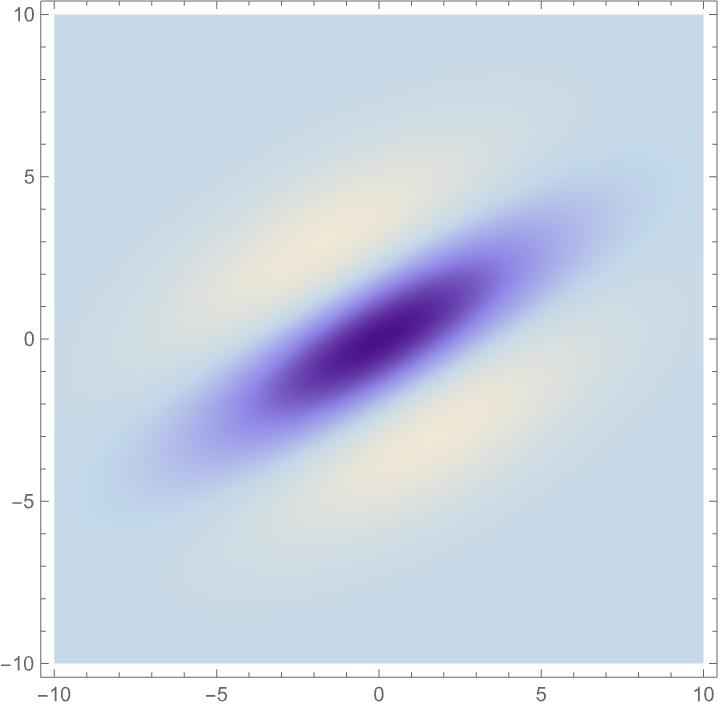}
      & \includegraphics[width=0.12\textwidth]{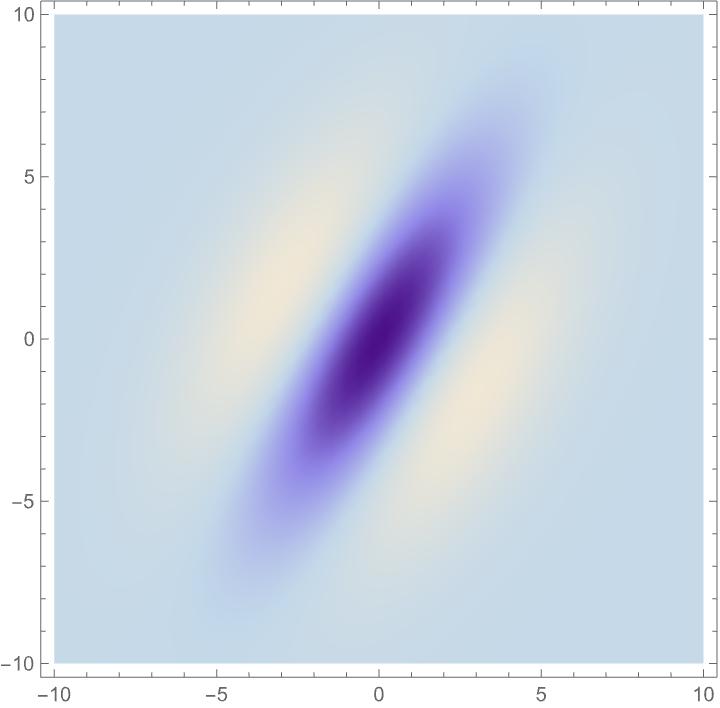}
      \\
      $\scriptsize{\sigma_1 = 2, \sigma_2 = 2}$
      & \includegraphics[width=0.12\textwidth]{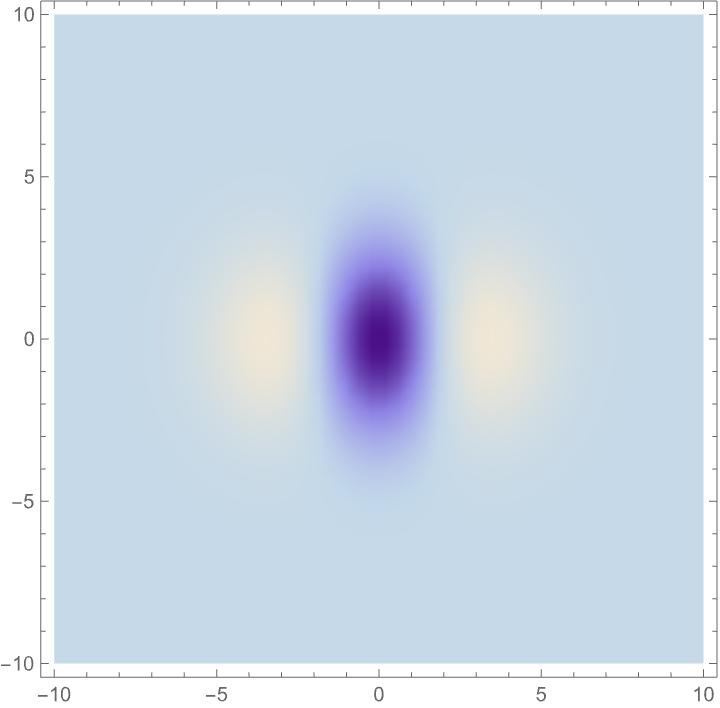}
      & \includegraphics[width=0.12\textwidth]{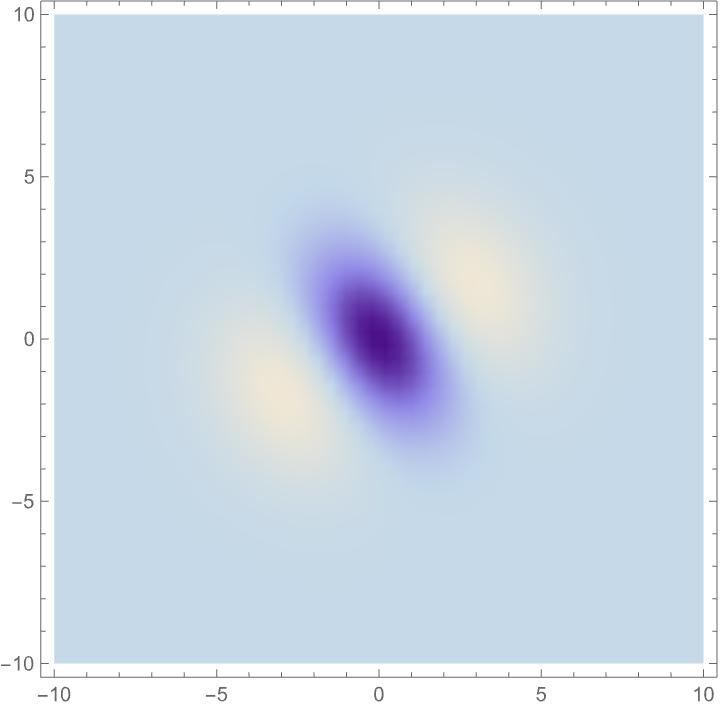}
      & \includegraphics[width=0.12\textwidth]{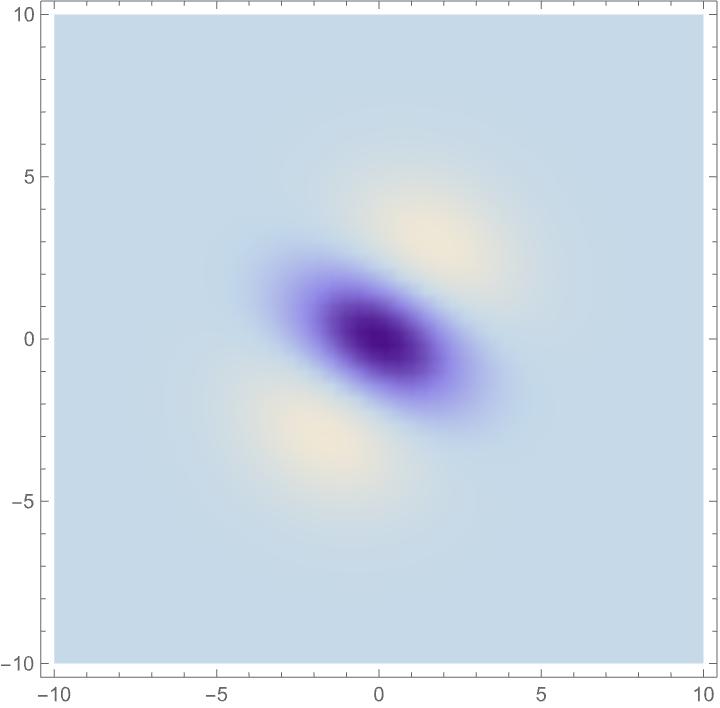}
      & \includegraphics[width=0.12\textwidth]{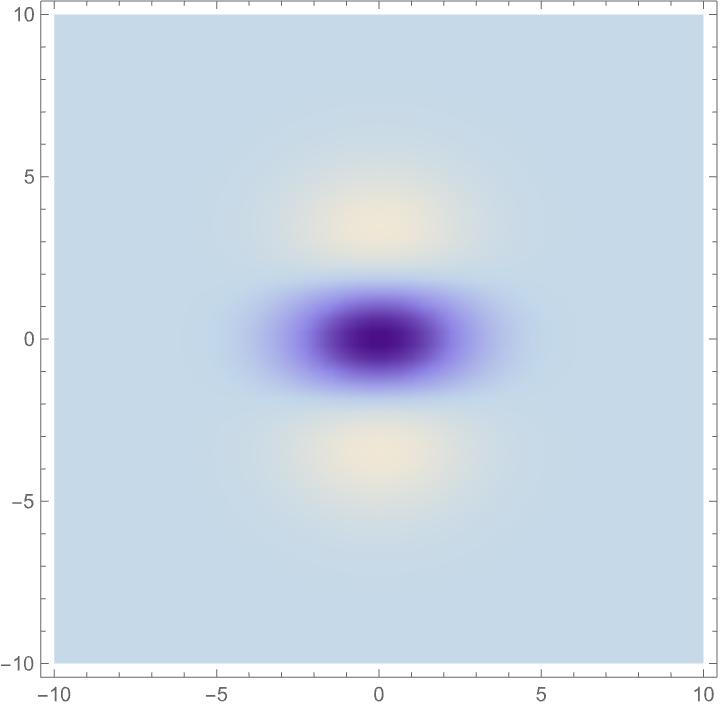}
      &   \includegraphics[width=0.12\textwidth]{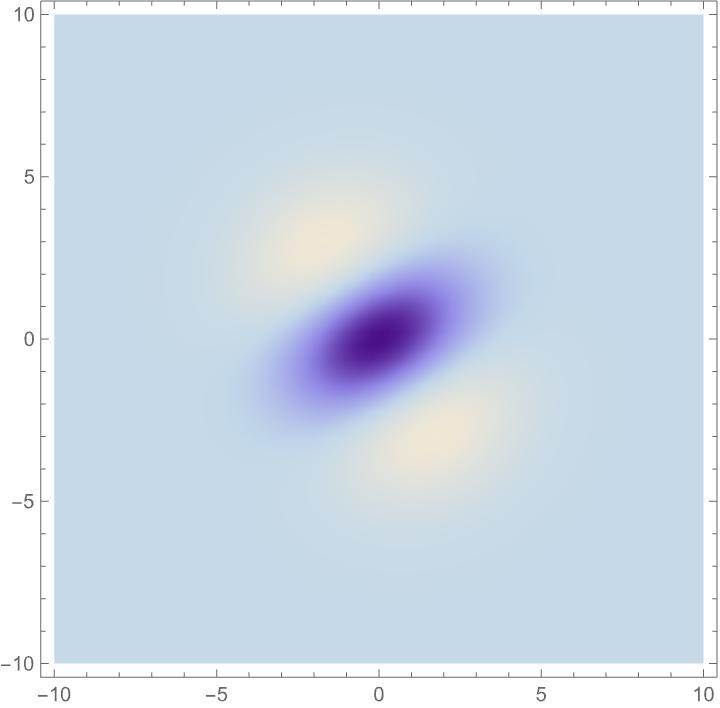}
      & \includegraphics[width=0.12\textwidth]{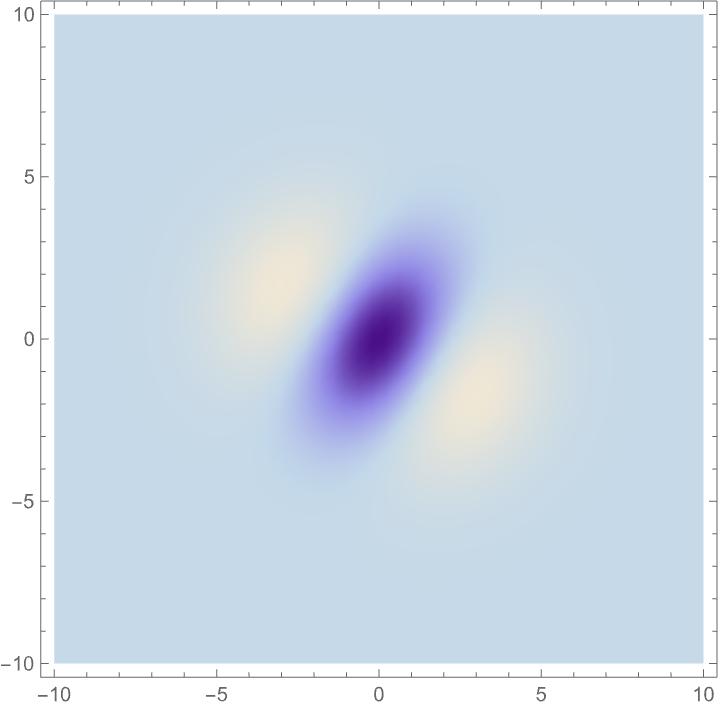}
      \\
      $\scriptsize{\sigma_1 = 1, \sigma_2 = 2}$
      & \includegraphics[width=0.12\textwidth]{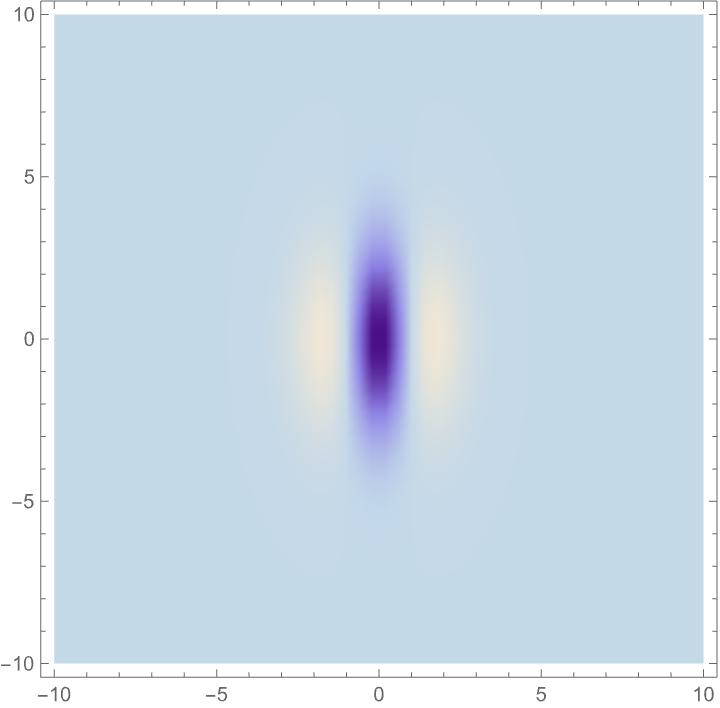}
      & \includegraphics[width=0.12\textwidth]{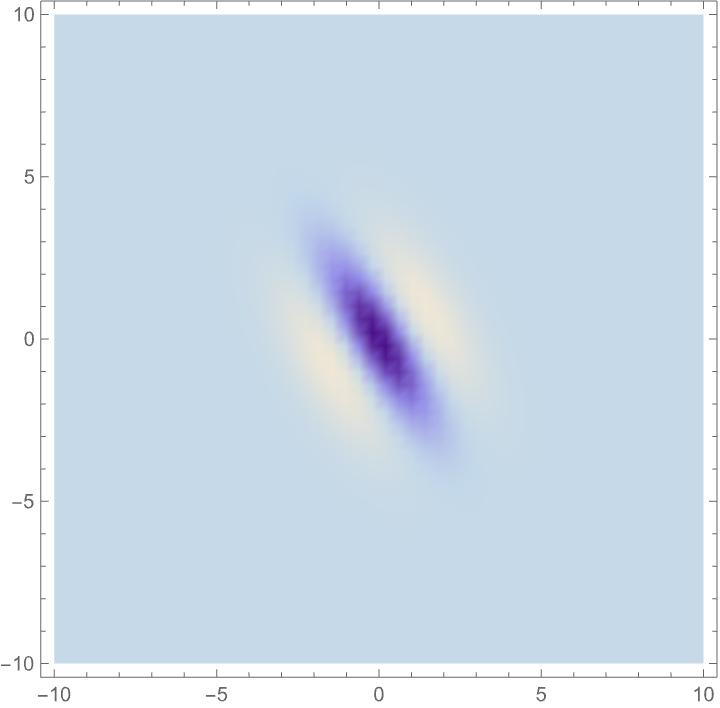}
      & \includegraphics[width=0.12\textwidth]{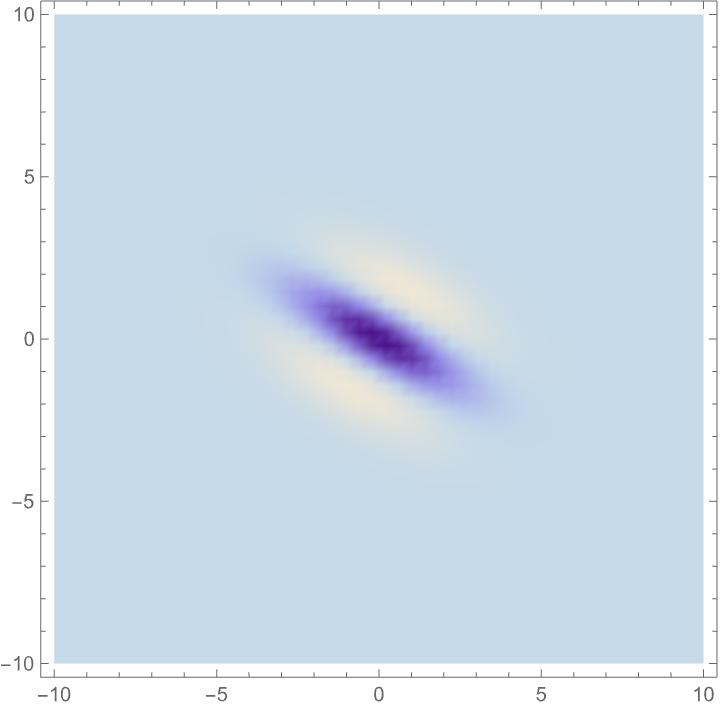}
      & \includegraphics[width=0.12\textwidth]{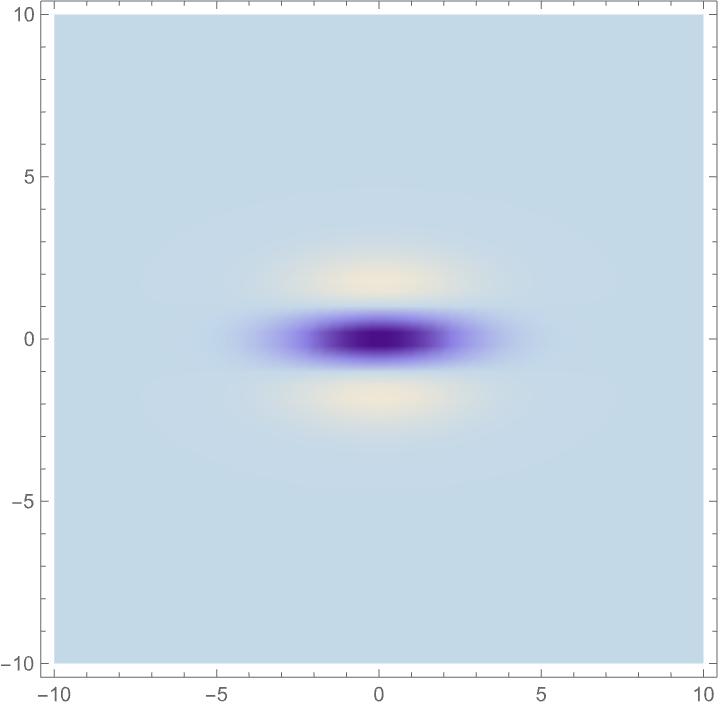}
      &   \includegraphics[width=0.12\textwidth]{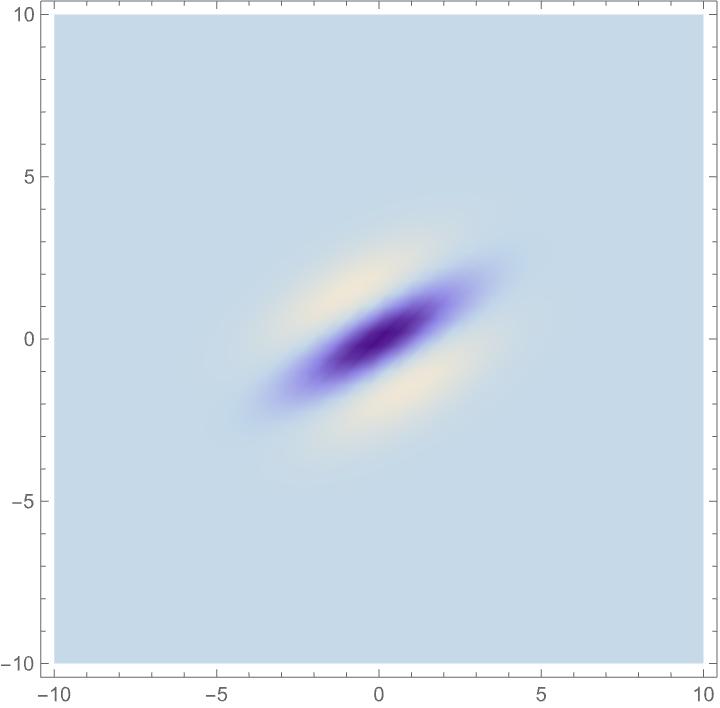}
      & \includegraphics[width=0.12\textwidth]{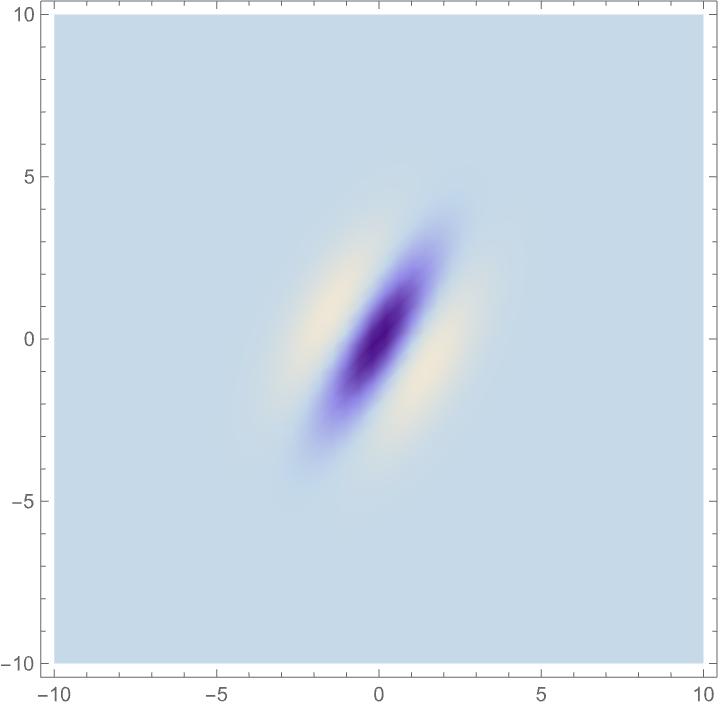}
      \\
      $\scriptsize{\sigma_1 = 1, \sigma_2 = 1}$
      & \includegraphics[width=0.12\textwidth]{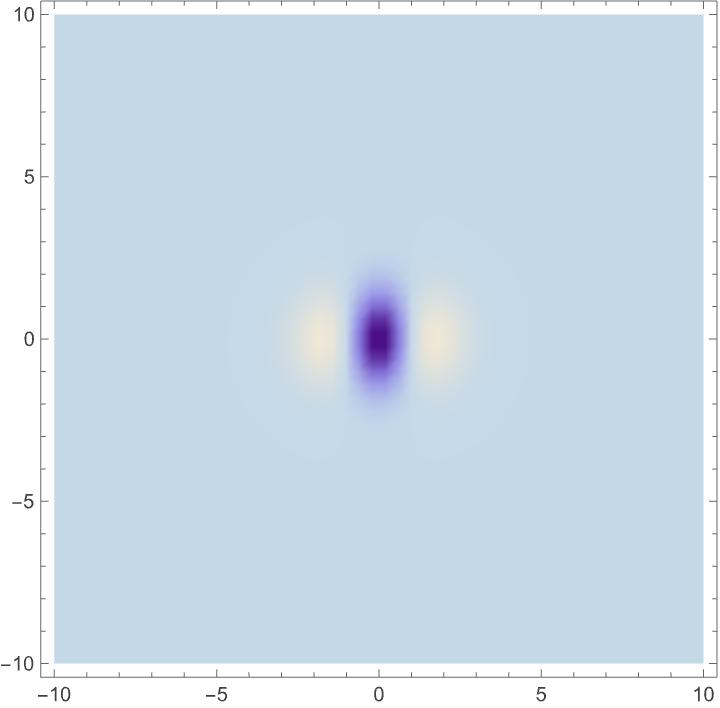}
      & \includegraphics[width=0.12\textwidth]{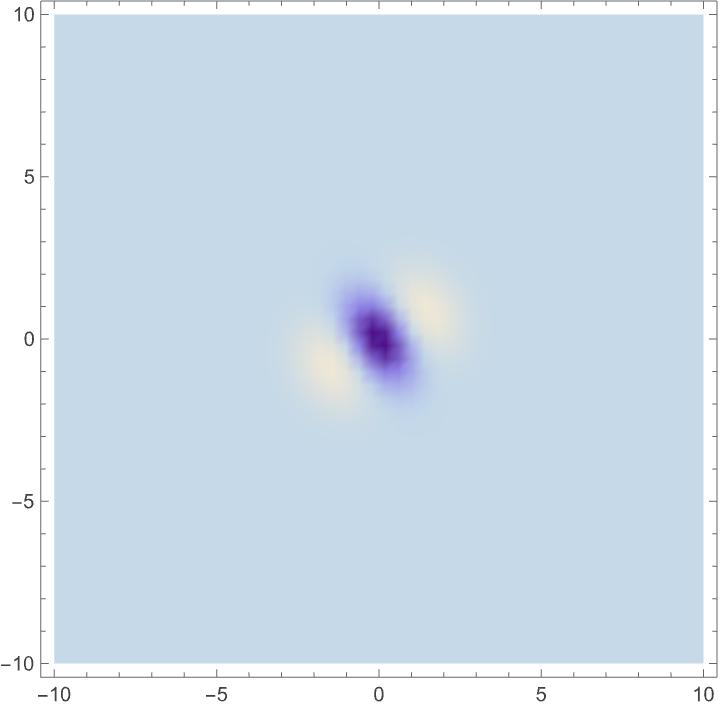}
      & \includegraphics[width=0.12\textwidth]{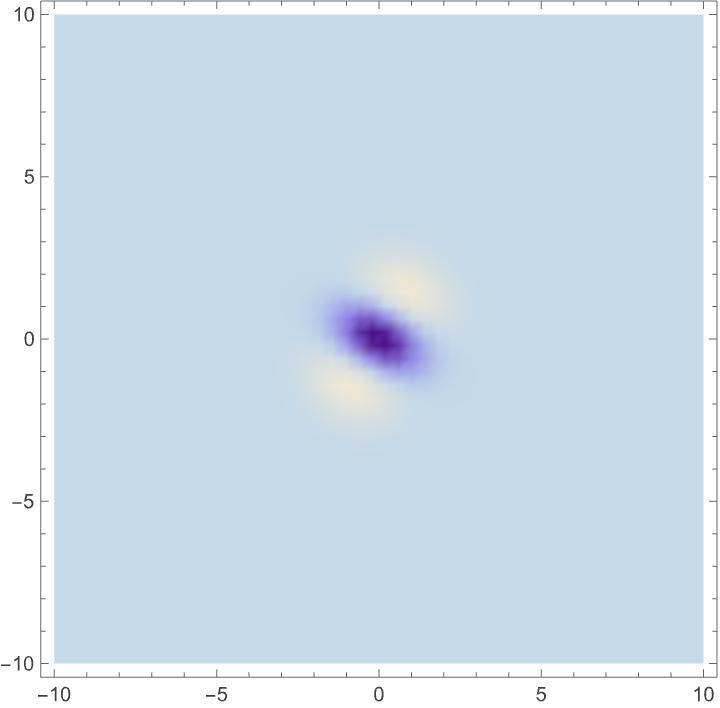}
      & \includegraphics[width=0.12\textwidth]{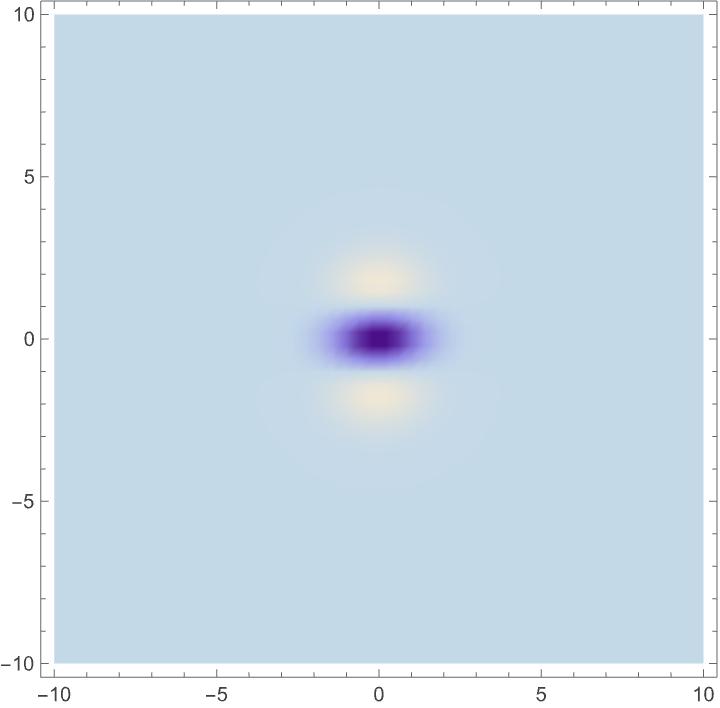}
      & \includegraphics[width=0.12\textwidth]{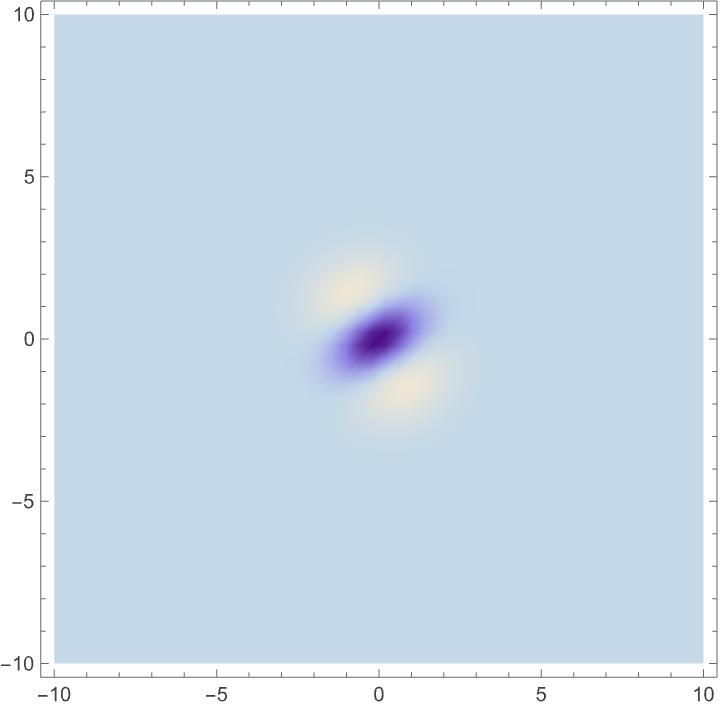}
      & \includegraphics[width=0.12\textwidth]{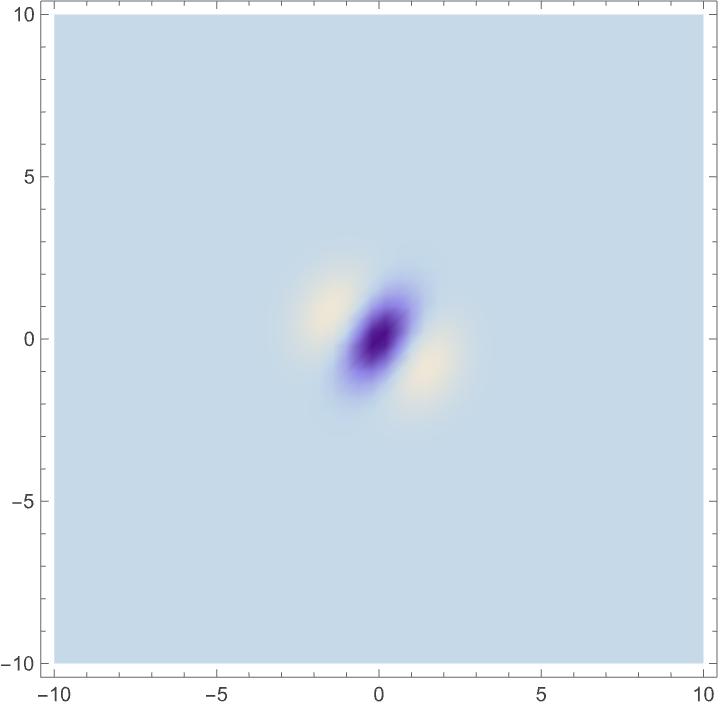}
      \\
    \end{tabular}
  \end{center}
  \caption{Purely spatial receptive fields in terms of
    directional derivatives $\partial_{\varphi}^m$ of affine Gaussian
    kernels $g(x;\; s, \Sigma)$ of the
    form (\ref{eq-gauss-fcn-2D}) for orders $m = 1$ and $m = 2$, shown for different
    combinations of the spatial scale parameters $\sigma_1$ and
    $\sigma_2$, corresponding to two different eccentricities
  $\epsilon = \sigma_2/\sigma_1 \in \{1, 2 \}$ of the receptive
  fields, according to the explicit parameterization of the affine
  Gaussian kernels according to (\ref{eq-expl-par-Cxx})--(\ref{eq-expl-par-Cyy})
  and
  (\ref{eq-def-aff-gauss-cont})--(\ref{eq-def-aff-gauss-cont-arg}).
  (Horizontal axes: Horizontal image coordinate $x_1 \in [-10, 10]$.
  Vertical axes: Vertical image coordinate $x_2 \in [-10, 10]$.)}
  \label{fig-affgaussders}
\end{figure*}

\begin{figure*}[hbtp]
  \begin{center}
    \begin{tabular}{cccccc}
      & $v = -1$ & $v = -1/2$ & $v = 0$ & $v = 1/2$ & $v = 1$
      \\
      $\scriptsize{\sigma_x = 2, \sigma_t = 2}$
      & \includegraphics[width=0.12\textwidth]{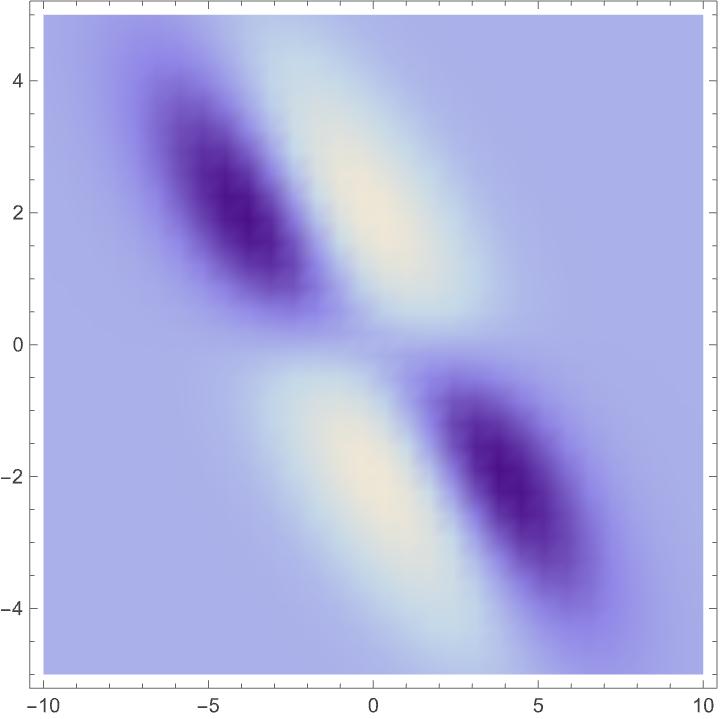}
      & \includegraphics[width=0.12\textwidth]{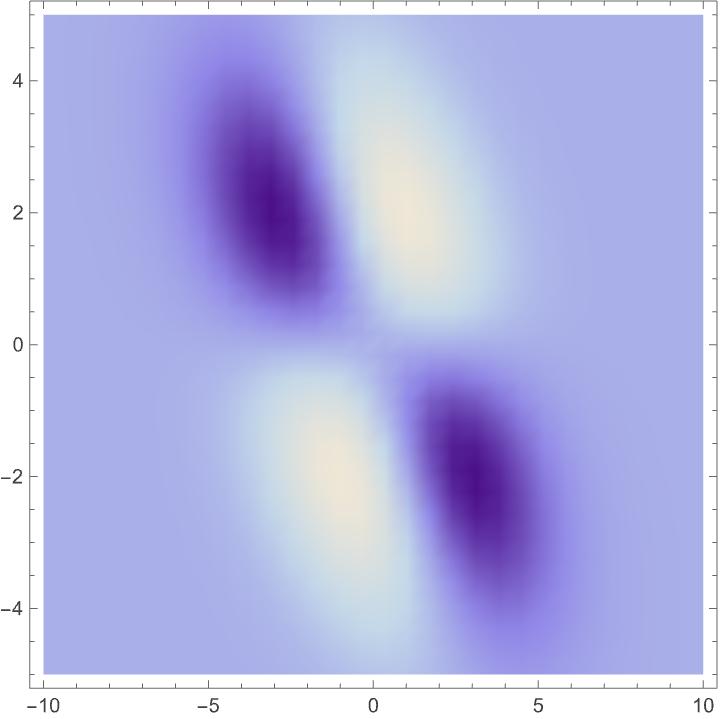}
      & \includegraphics[width=0.12\textwidth]{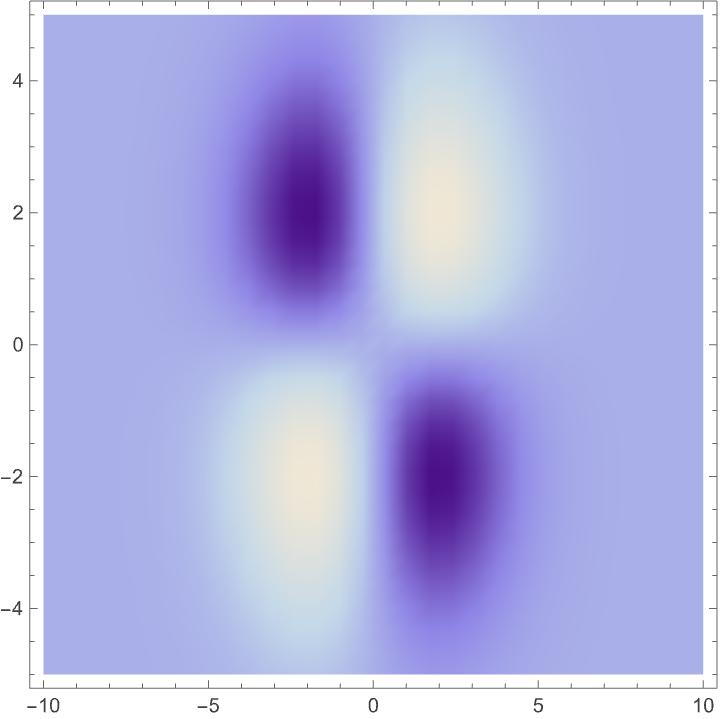}
      & \includegraphics[width=0.12\textwidth]{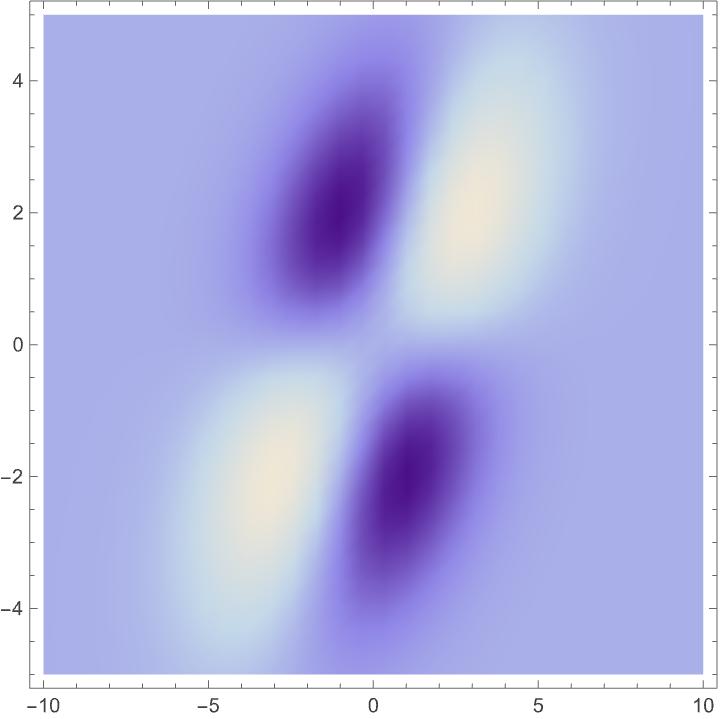}
      & \includegraphics[width=0.12\textwidth]{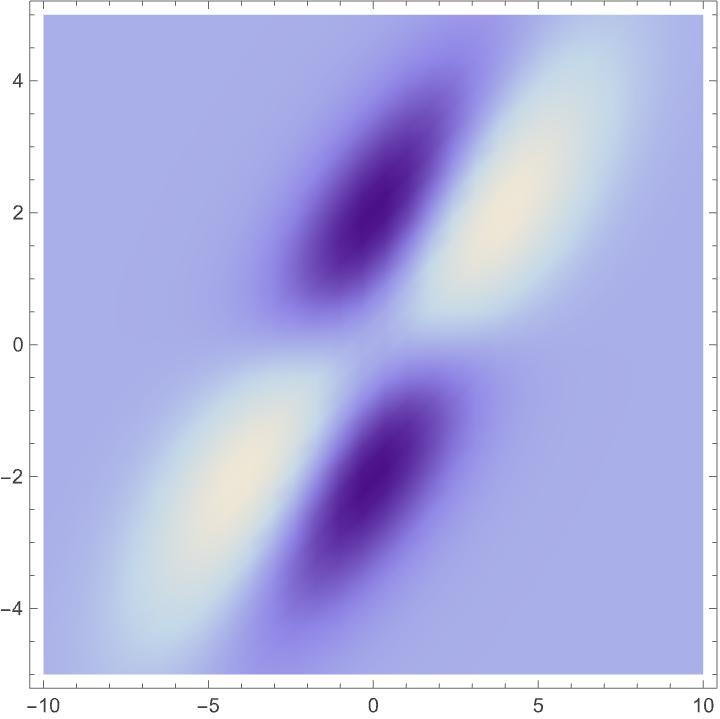}                                          
      \\
      $\scriptsize{\sigma_x = 1, \sigma_t = 2}$
      & \includegraphics[width=0.12\textwidth]{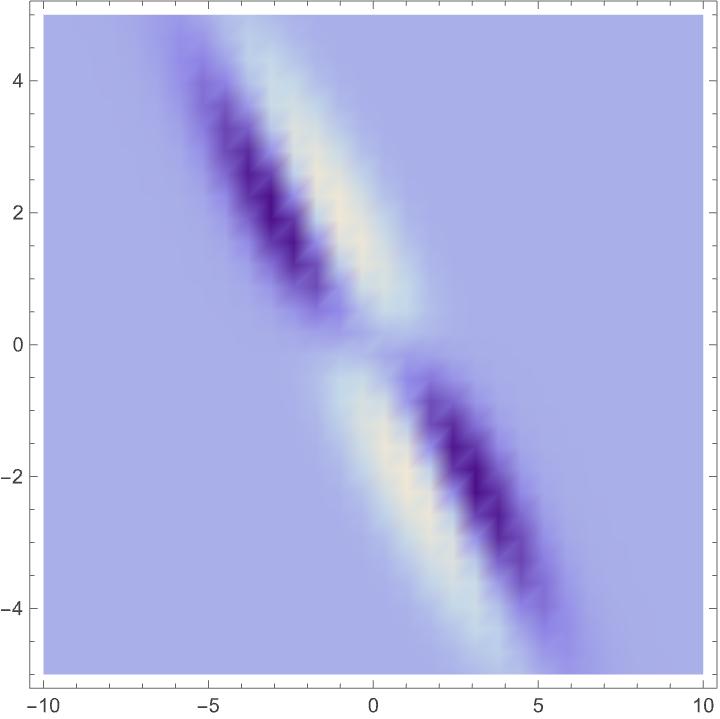}
      & \includegraphics[width=0.12\textwidth]{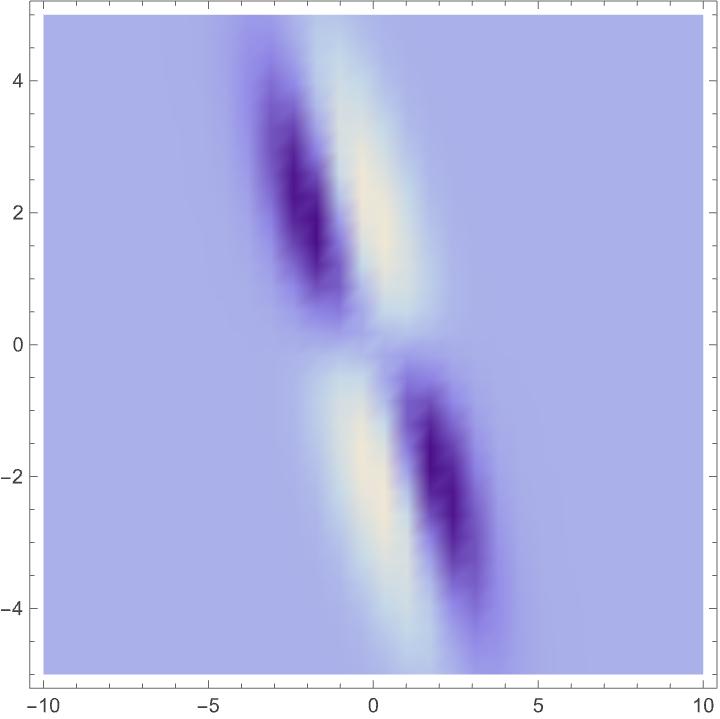}
      & \includegraphics[width=0.12\textwidth]{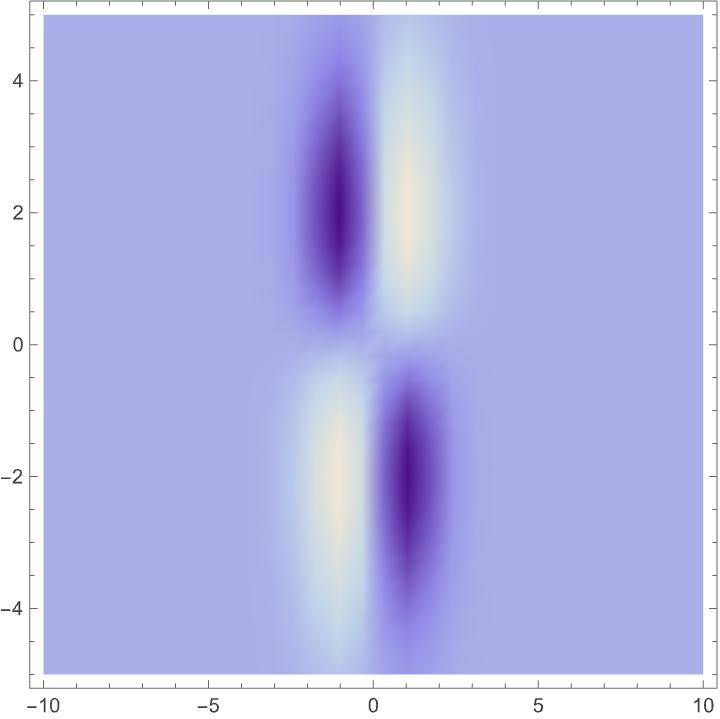}
      & \includegraphics[width=0.12\textwidth]{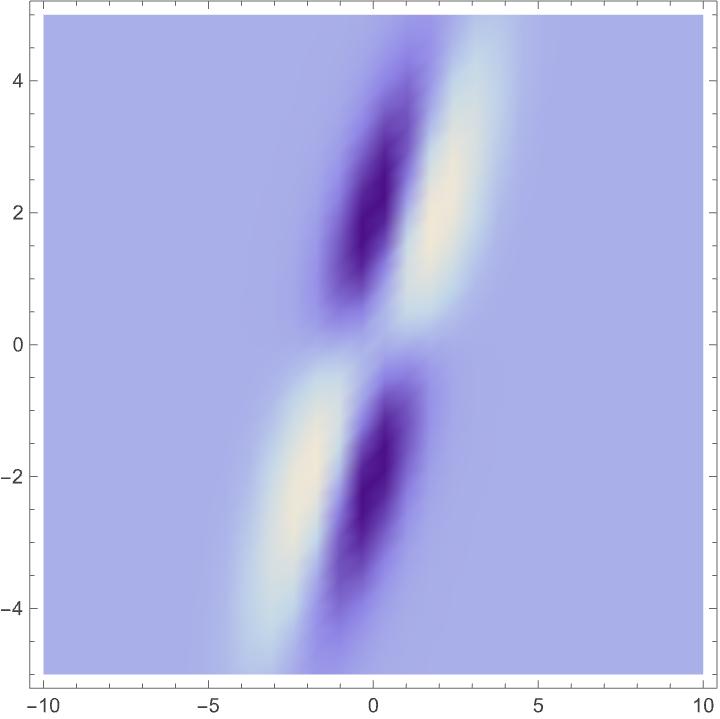}
      & \includegraphics[width=0.12\textwidth]{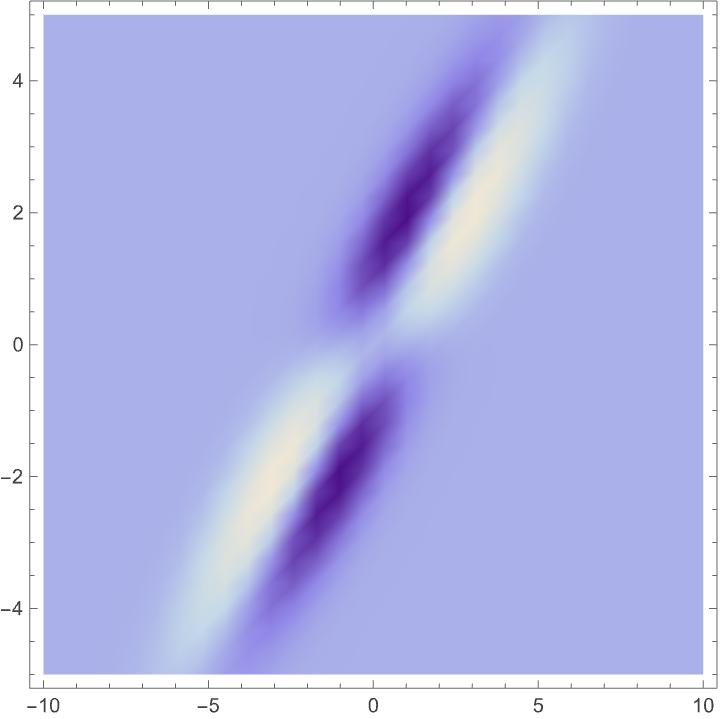}                               
      \\
      $\scriptsize{\sigma_x = 2, \sigma_t = 1}$
      & \includegraphics[width=0.12\textwidth]{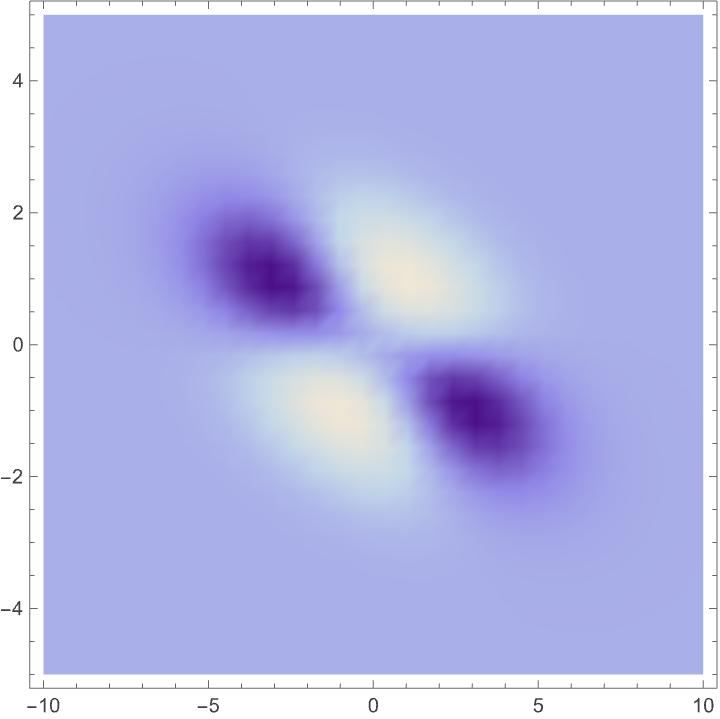}
      & \includegraphics[width=0.12\textwidth]{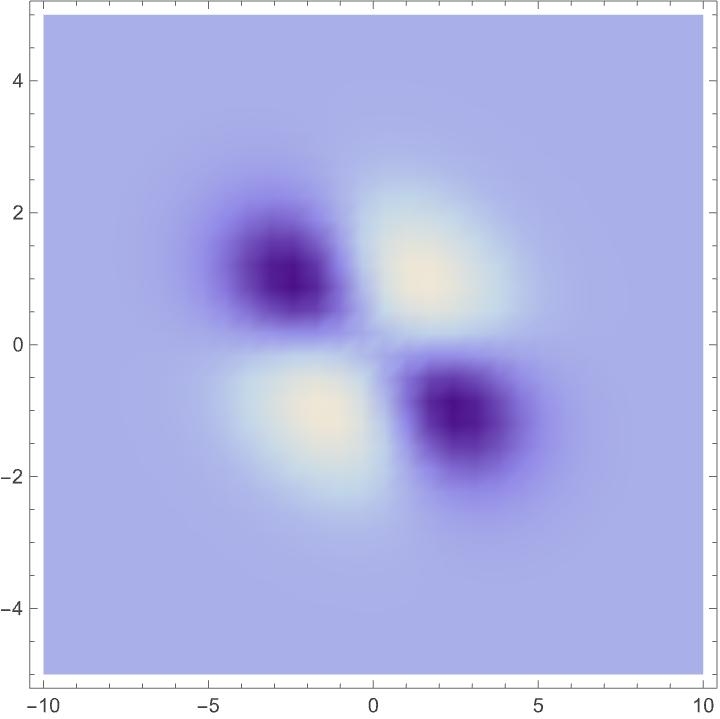}
      & \includegraphics[width=0.12\textwidth]{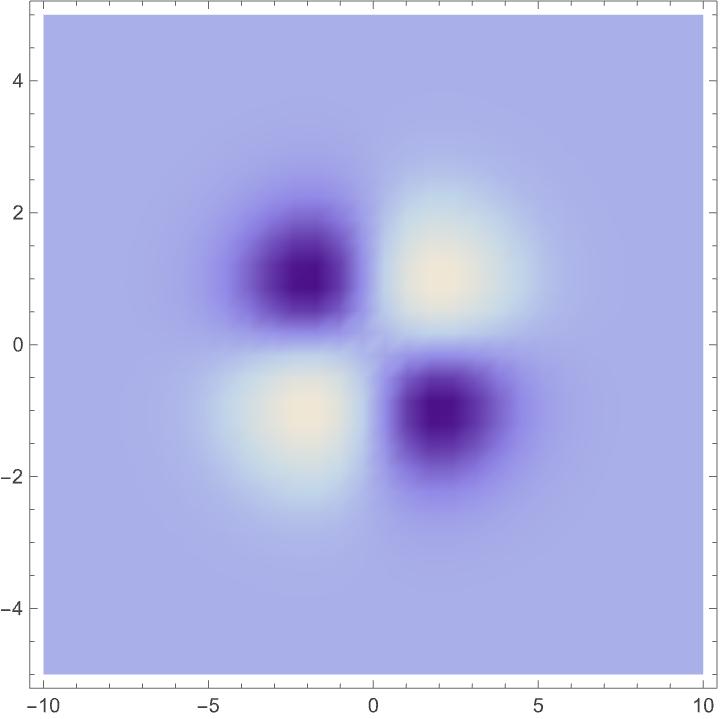}
      & \includegraphics[width=0.12\textwidth]{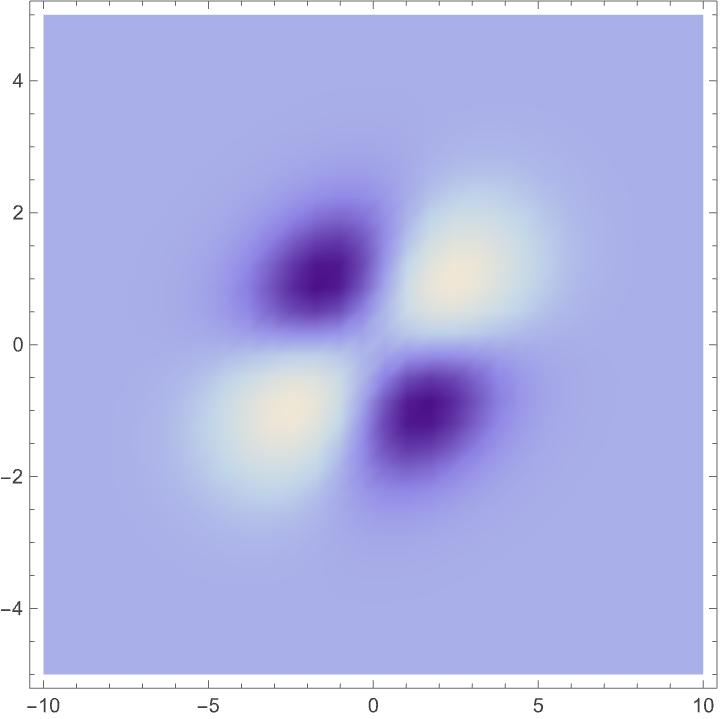}
      & \includegraphics[width=0.12\textwidth]{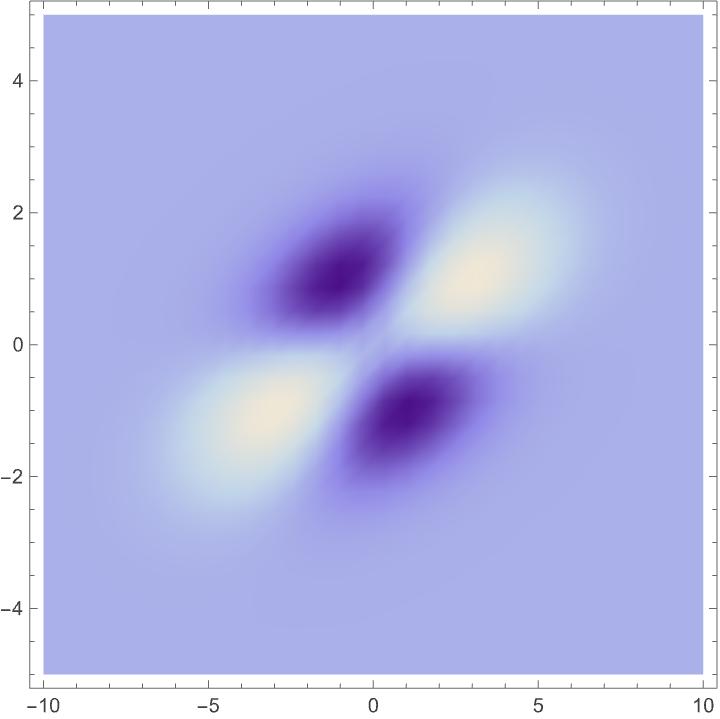}                                          
      \\
      $\scriptsize{\sigma_x = 1, \sigma_t = 1}$
      & \includegraphics[width=0.12\textwidth]{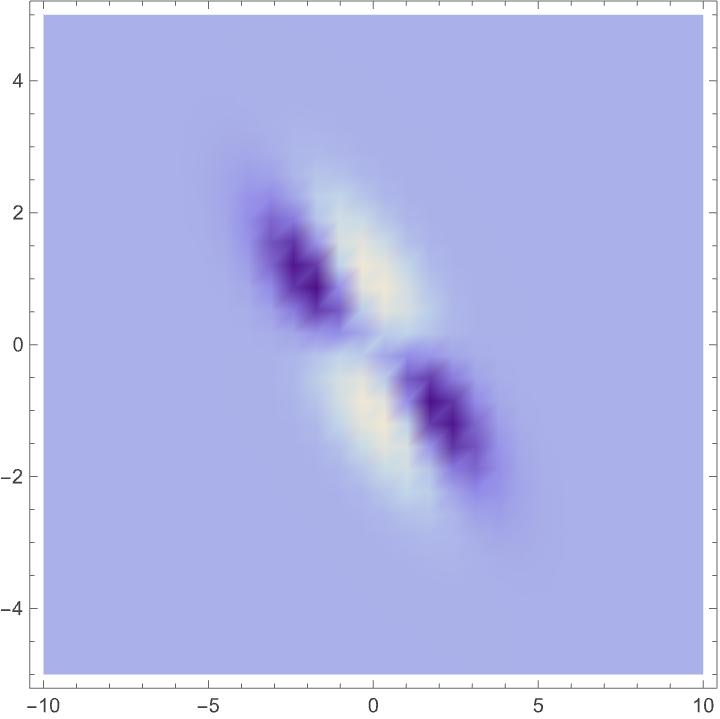}
      & \includegraphics[width=0.12\textwidth]{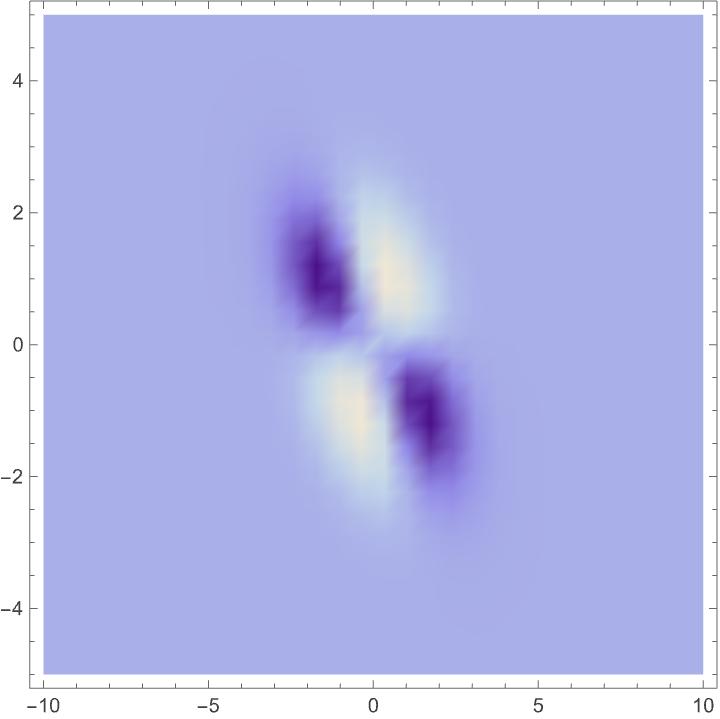}
      & \includegraphics[width=0.12\textwidth]{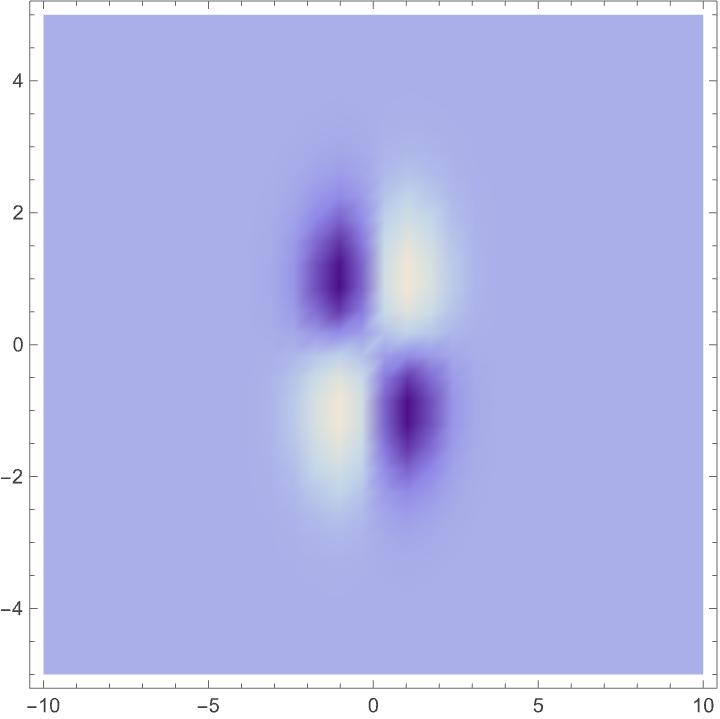}
      & \includegraphics[width=0.12\textwidth]{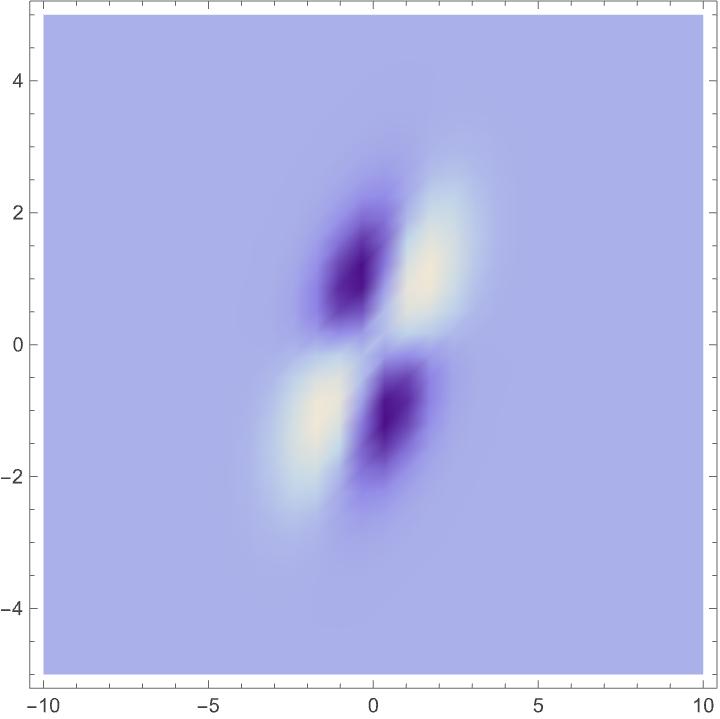}
      & \includegraphics[width=0.12\textwidth]{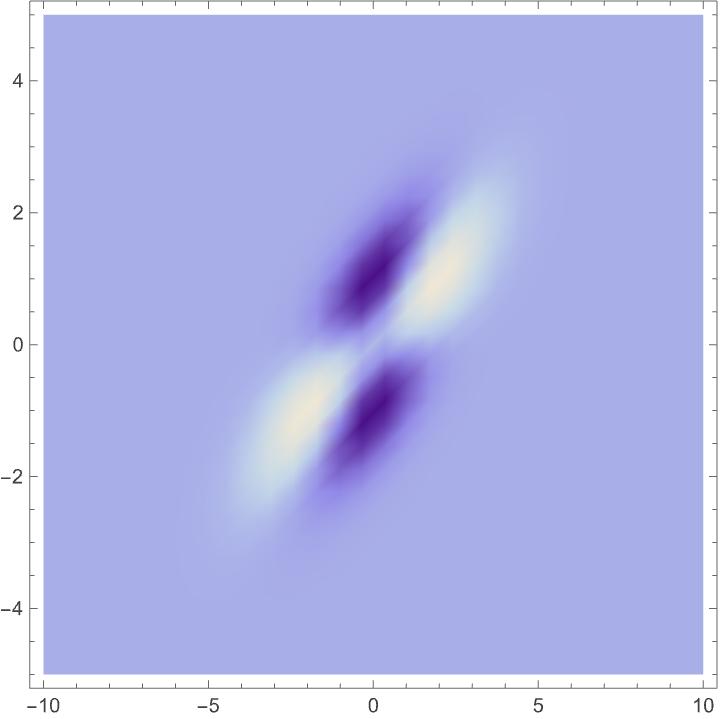}                                          
      \\
    \end{tabular}
  \end{center}
  \caption{Non-causal joint spatio-temporal receptive fields over
    a 1+1D spatio-temporal domain in terms of the mixed first-order
    spatial derivative and the first-order velocity-adapted temporal
    derivative of the form $T_{x\bar{t}}(x, t;\; s, \tau, v)$
    according to (\ref{eq-mixed-strf-1-1}) as the product of a
    velocity-adapted 1-D Gaussian kernel $g_{1D}(x - v \, t;\; s)$
    over the spatial domain and the non-causal Gaussian kernel
    $h(t;\; \tau) = g(t;\; \tau)$ over the
    temporal domain according to (\ref{eq-non-caus-temp-gauss}).
    The spatio-temporal receptive fields are shown for different
    values of the spatial scale parameter $\sigma_x = \sqrt{s}$ and
    the temporal scale parameter $\sigma_t = \sqrt{\tau}$ in
    dimensions of $[\mbox{length}]$ and $[\mbox{time}]$.
    (Horizontal axes: Spatial image coordinate $x \in [-10, 10]$.
  Vertical axes: Temporal variable $t \in [-4, 4]$.)}
  \label{fig-noncaus-strfs}

  \bigskip
  
  \begin{center}
    \begin{tabular}{cccccc}
      & $v = -1$ & $v = -1/2$ & $v = 0$ & $v = 1/2$ & $v = 1$
      \\
      $\scriptsize{\sigma_x = 2, \sigma_t = 2}$
      & \includegraphics[width=0.12\textwidth]{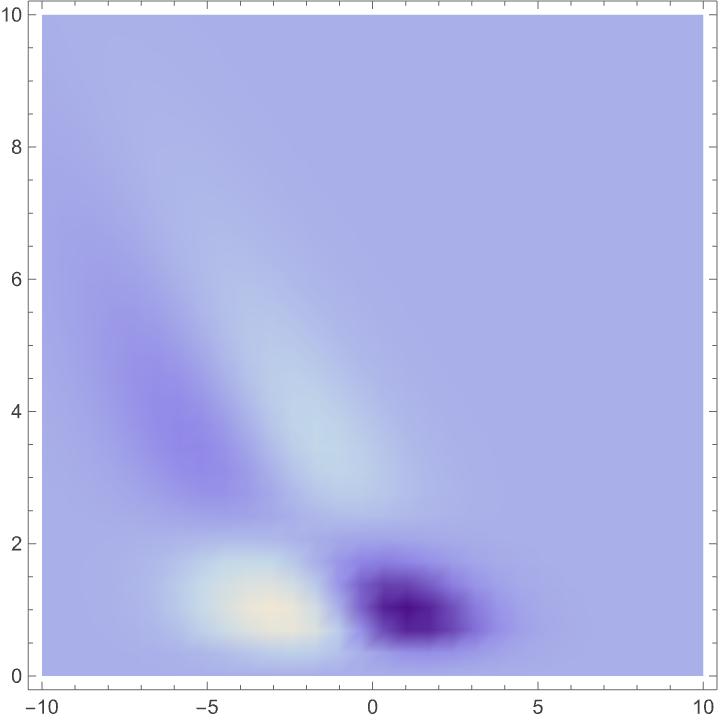}
      & \includegraphics[width=0.12\textwidth]{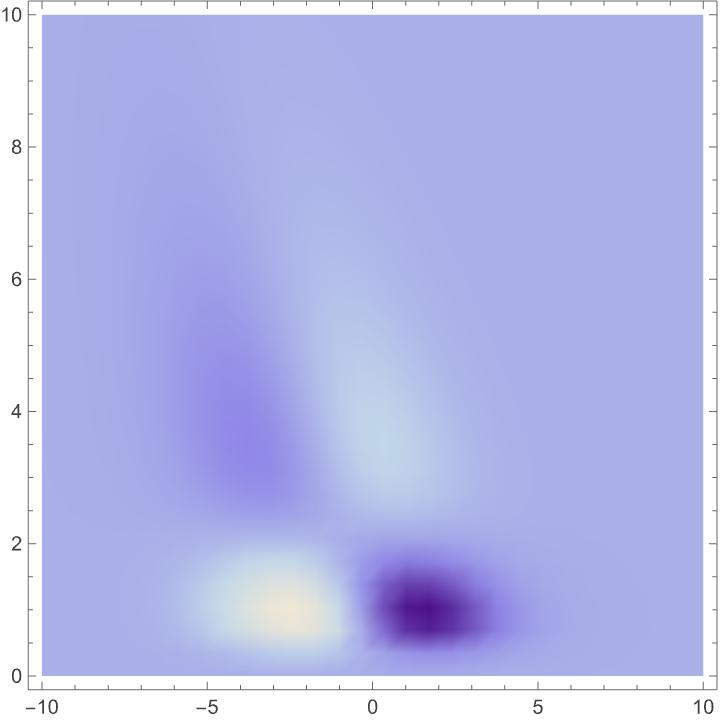}
      & \includegraphics[width=0.12\textwidth]{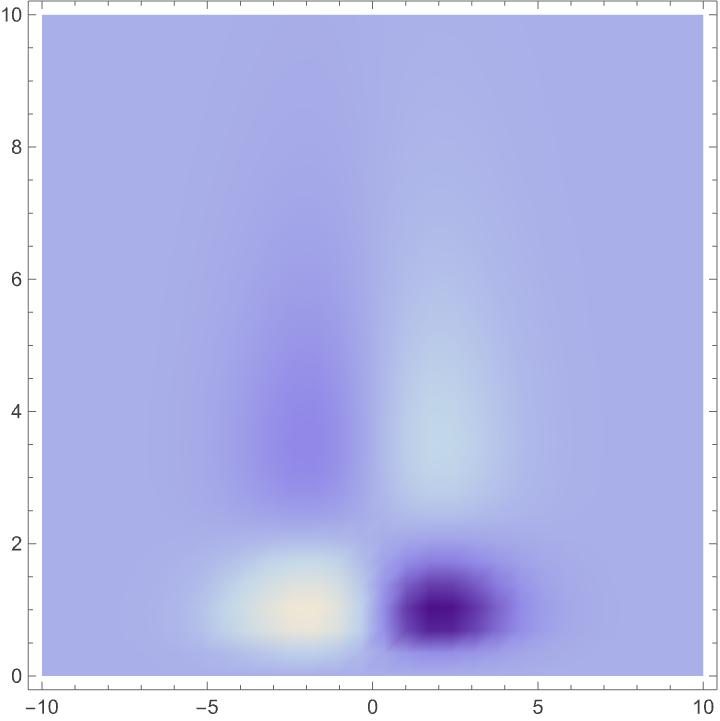}
      & \includegraphics[width=0.12\textwidth]{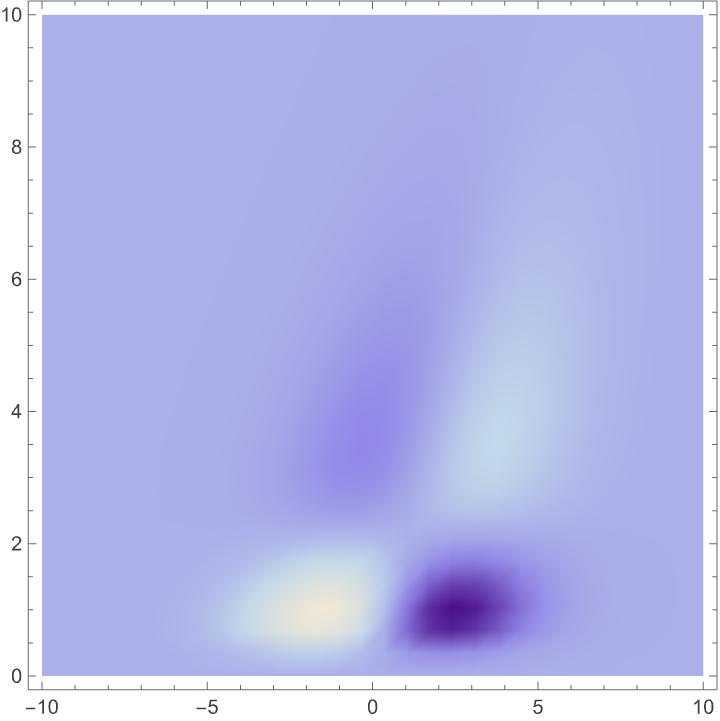}
      & \includegraphics[width=0.12\textwidth]{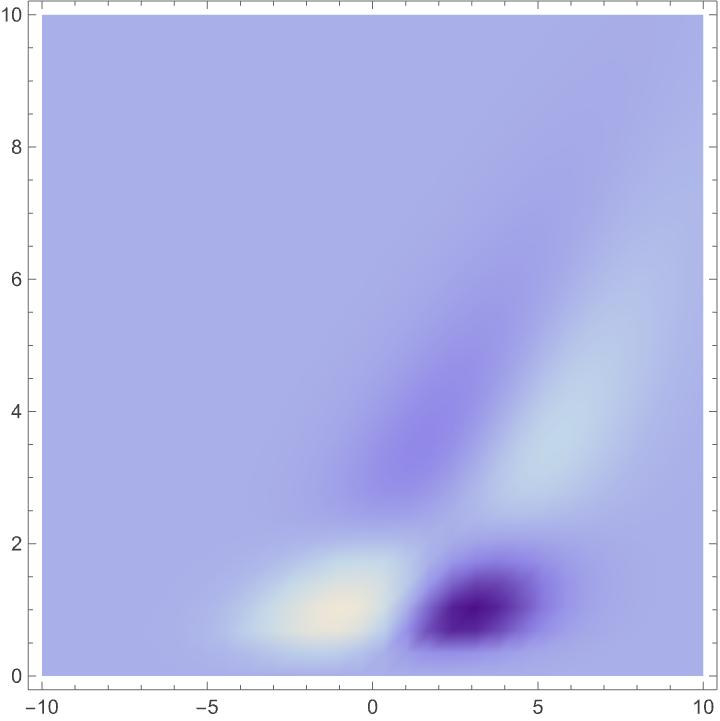}                                          
      \\
      $\scriptsize{\sigma_x = 1, \sigma_t = 2}$
      & \includegraphics[width=0.12\textwidth]{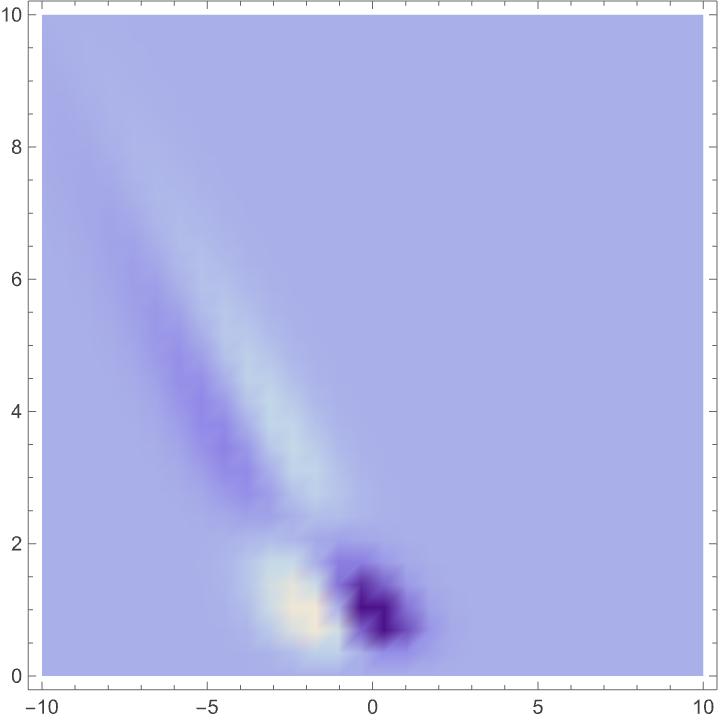}
      & \includegraphics[width=0.12\textwidth]{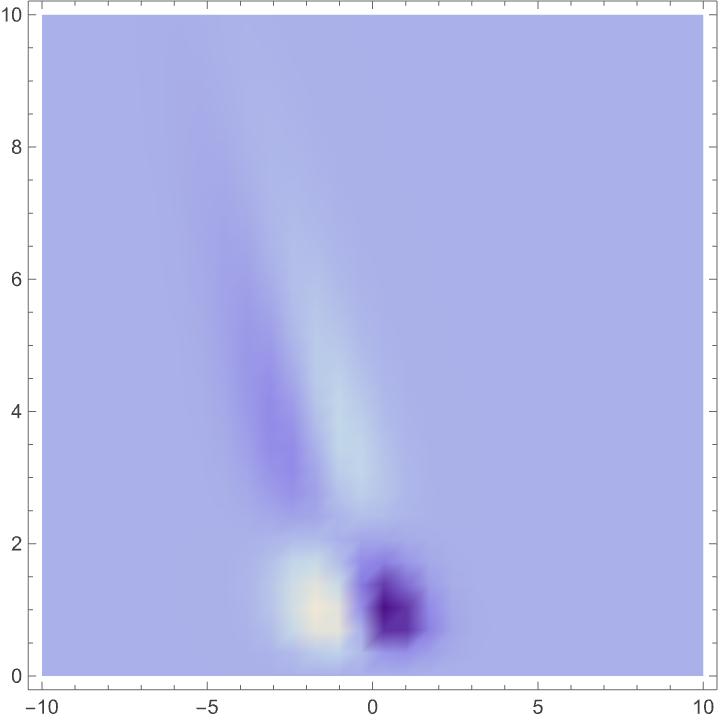}
      & \includegraphics[width=0.12\textwidth]{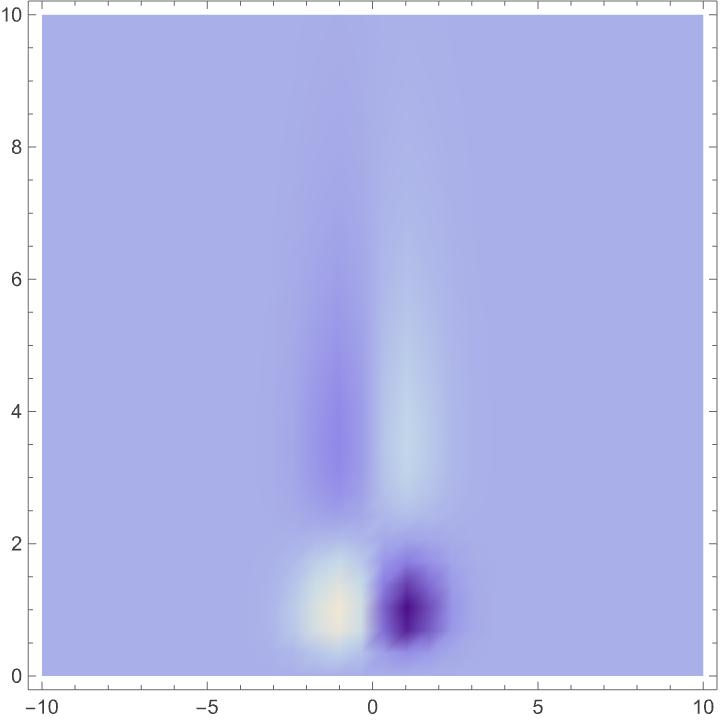}
      & \includegraphics[width=0.12\textwidth]{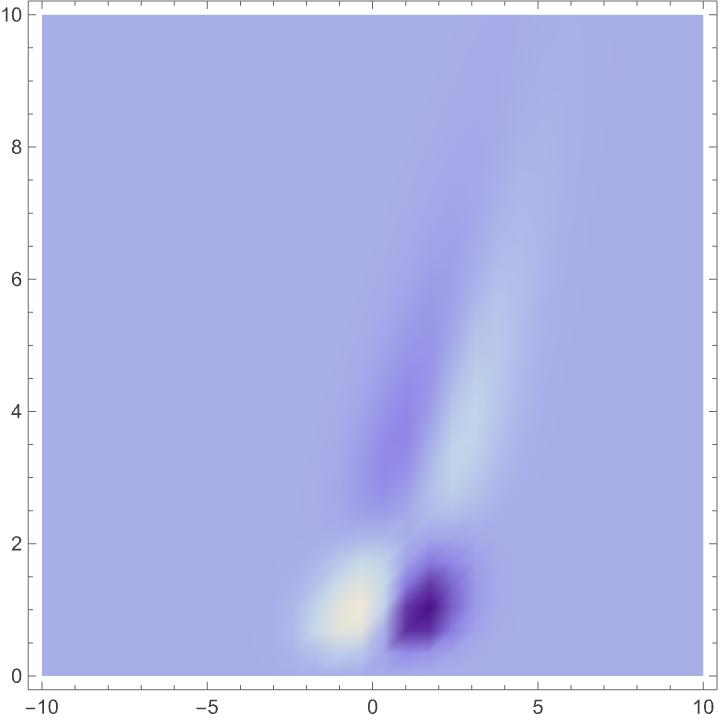}
      & \includegraphics[width=0.12\textwidth]{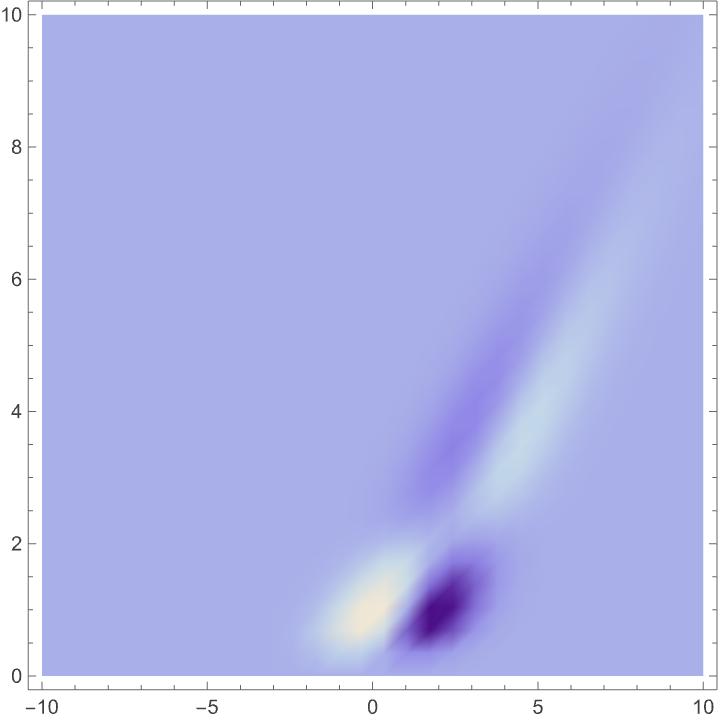}                               
      \\
      $\scriptsize{\sigma_x = 2, \sigma_t = 1}$
      & \includegraphics[width=0.12\textwidth]{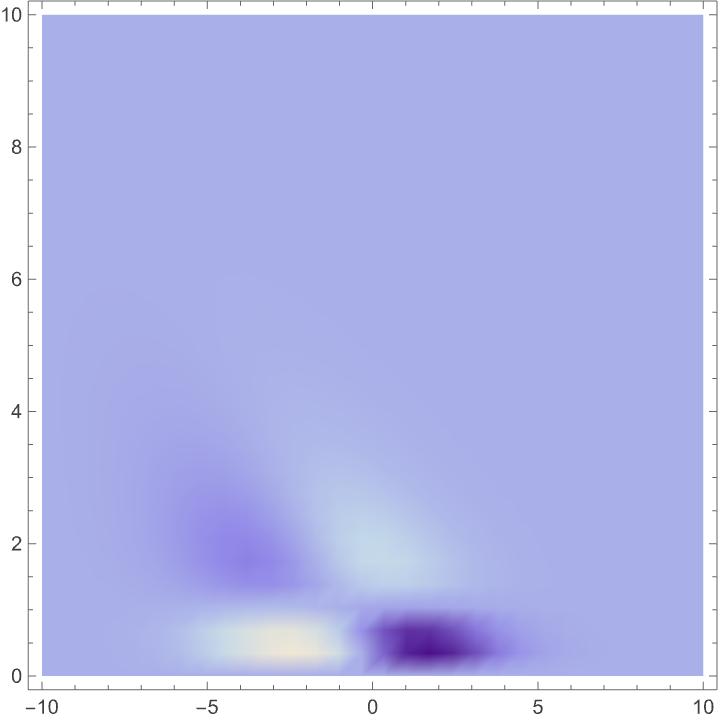}
      & \includegraphics[width=0.12\textwidth]{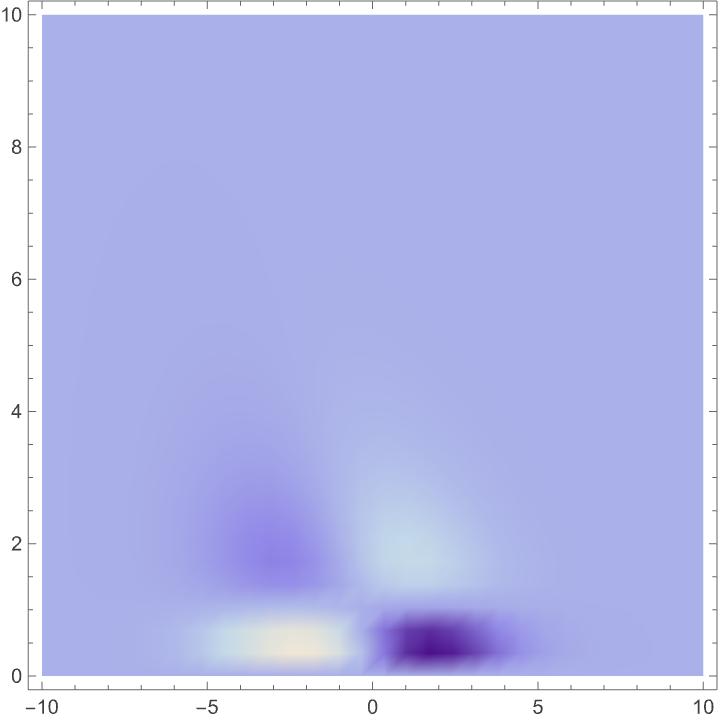}
      & \includegraphics[width=0.12\textwidth]{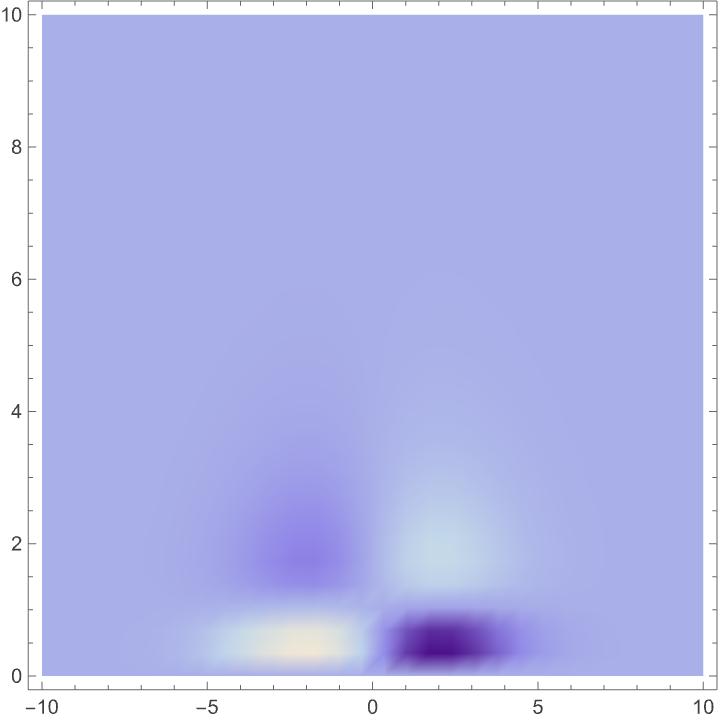}
      & \includegraphics[width=0.12\textwidth]{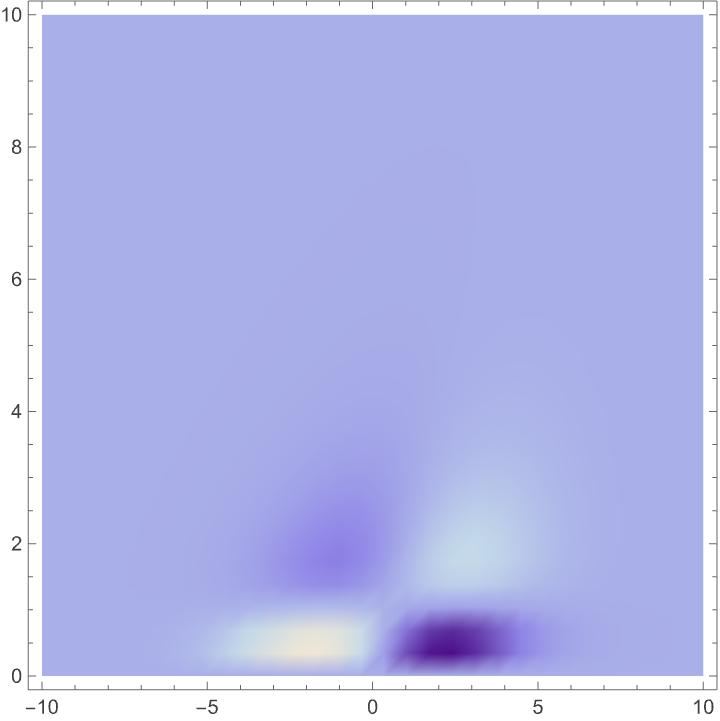}
      & \includegraphics[width=0.12\textwidth]{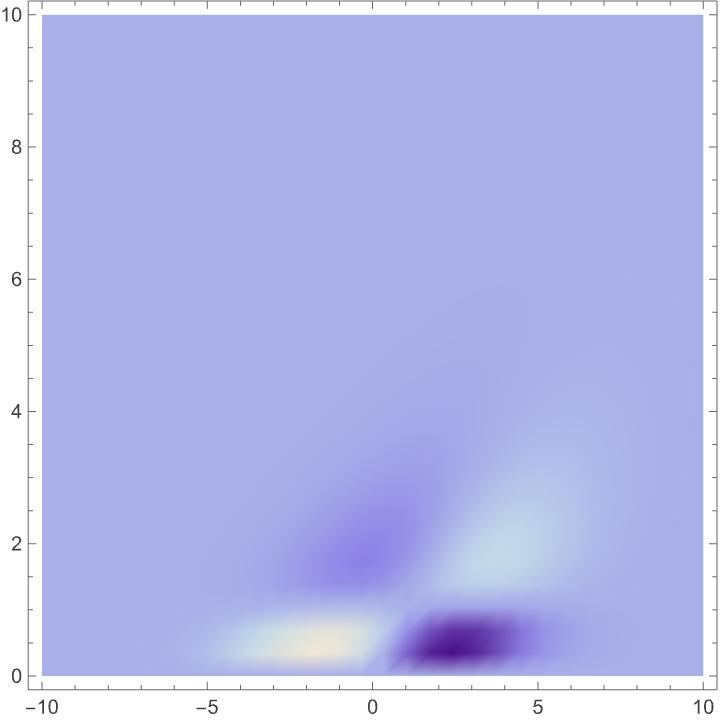}                                          
      \\
      $\scriptsize{\sigma_x = 1, \sigma_t = 1}$
      & \includegraphics[width=0.12\textwidth]{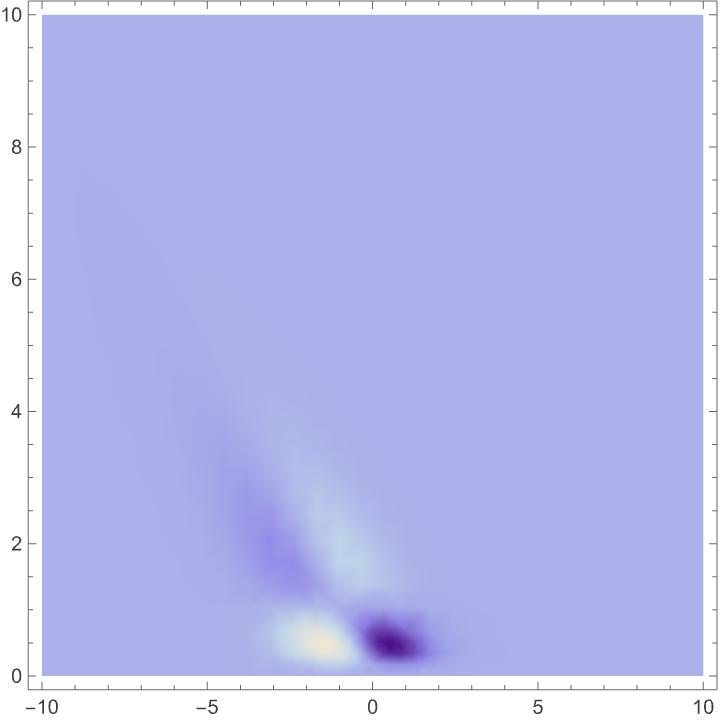}
      & \includegraphics[width=0.12\textwidth]{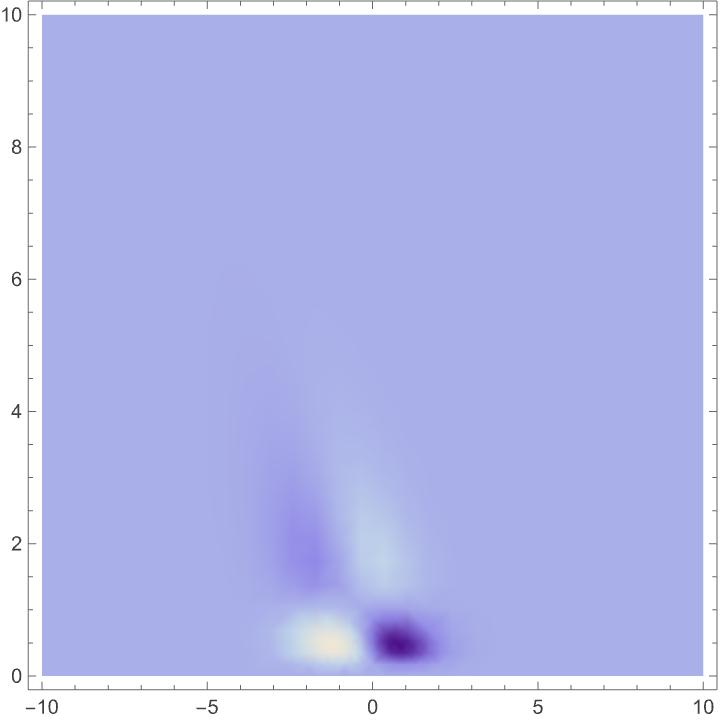}
      & \includegraphics[width=0.12\textwidth]{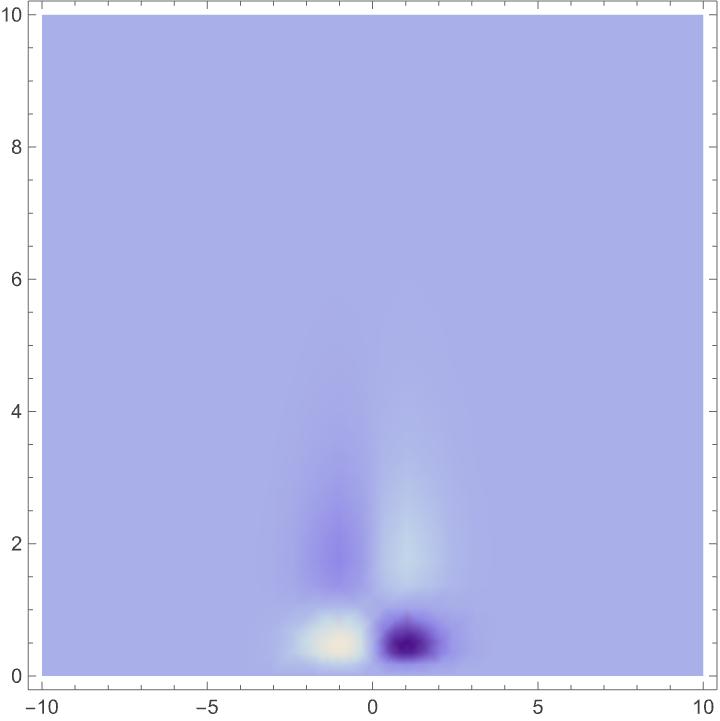}
      & \includegraphics[width=0.12\textwidth]{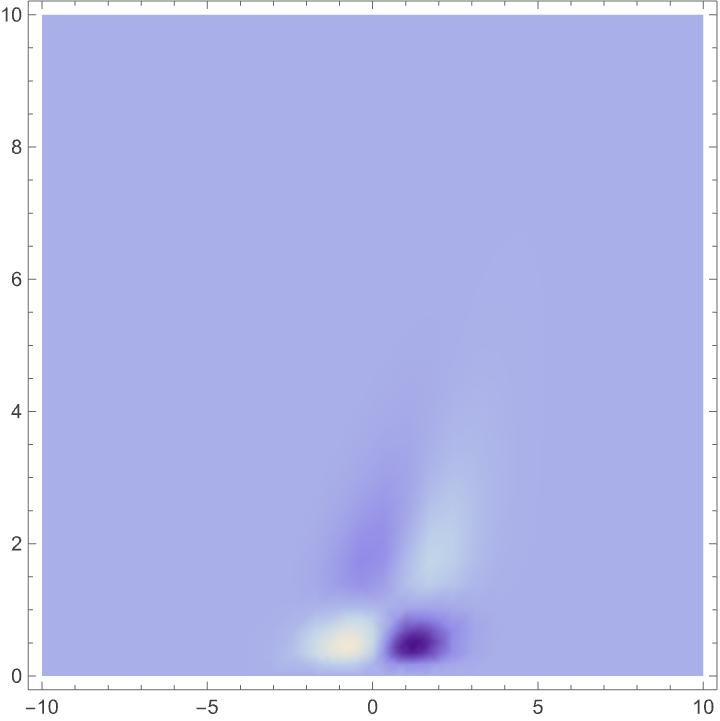}
      & \includegraphics[width=0.12\textwidth]{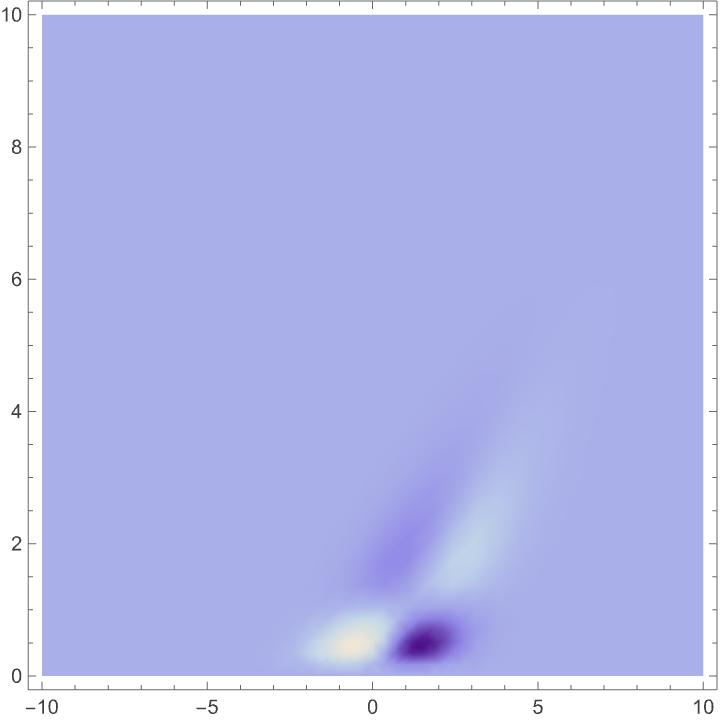}                                          
      \\
    \end{tabular}
  \end{center}
  \caption{Time-causal joint spatio-temporal receptive fields over
    a 1+1D spatio-temporal domain in terms of the mixed first-order
    spatial derivative and the first-order velocity-adapted temporal
    derivative of the form $T_{x\bar{t}}(x, t;\; s, \tau, v)$
    according to (\ref{eq-mixed-strf-1-1}) as the product of a
    velocity-adapted 1-D Gaussian kernel $g_{1D}(x - v \, t;\; s)$
    over the spatial domain and the time-causal limit kernel
    $h(t;\; \tau) = \psi(t;\; \tau, c)$ over the
    temporal domain according to (\ref{eq-time-caus-lim-kern})
    with the distribution parameter set to $c = 2$.
    The spatio-temporal receptive fields are shown for different
    values of the spatial scale parameter $\sigma_x = \sqrt{s}$ and
    the temporal scale parameter $\sigma_t = \sqrt{\tau}$ in
    dimensions of $[\mbox{length}]$ and $[\mbox{time}]$.
    (Horizontal axes: Spatial image coordinate $x \in [-10, 10]$.
  Vertical axes: Temporal variable $t \in [0, 8]$.)}
  \label{fig-timecaus-strfs}
\end{figure*}

\subsection{Receptive field families in terms of spatial and
  spatio-temporal derivatives}

For defining either purely spatial or joint spatio-temporal receptive
fields, to be used for processing visual image data,
the either purely spatial or joint spatio-temporal smoothing
operations are, in turn, combined with either purely spatial
differential operators ${\cal D}_x$ or joint spatio-temporal
differentiation operators ${\cal D}_{x,t}$, to give rise to spatial
receptive fields applied to spatial image data
$f \colon \bbbr^2 \rightarrow \bbbr$ of the form
\begin{equation}
  \label{eq-spat-rf-response}
  {\cal R} \, f(\cdot)  = {\cal D}_x (T(\cdot;\; s, \Sigma) * f(\cdot)),
\end{equation}
or joint spatio-temporal receptive fields applied to video sequences
or video streams $f \colon \bbbr^2 \times \bbbr \rightarrow \bbbr$
of the form
\begin{equation}
  \label{eq-spattemp-rf-response}
  {\cal R} \, f(\cdot, \cdot)
  = {\cal D}_{x,t} (T(\cdot, \cdot;\; s, \Sigma, \tau, v) * f(\cdot, \cdot)).
\end{equation}
Regarding the choice of what spatial differentiation operators
${\cal D}_x$ to use,
common choices include directional derivative operators in the orientation
$\varphi \in [-\pi, \pi]$ of the form
\begin{equation}
    \label{eq-def-spat-dir-der}
  {\cal D}_x
  = \partial_{\varphi}^m
  = (\cos \varphi \, \partial_{x_1} + \sin \varphi \, \partial_{x_2})^m,
\end{equation}
the spatial gradient operator
\begin{equation}
  \label{eq-def-spat-grad}
  {\cal D}_x
  = \nabla_x
  = (\partial_{x_1},  \partial_{x_2})^T,
\end{equation}
or the Hessian matrix operator
\begin{equation}
  \label{eq-def-spat-hess}
  {\cal D}_x
  = {\cal H}_x = \nabla_x \nabla_x^T
  = \left(
        \begin{array}{cc}
           \partial_{x_1 x_1} & \partial_{x_1 x_2} \\
           \partial_{x_1 x_2} & \partial_{x_2 x_2} \\
        \end{array}
      \right).
\end{equation}
Concerning joint spatio-temporal derivative operators ${\cal D}_{x,t}$,
a natural
choice is to combine the above purely spatial derivative operators
${\cal D}_x$ with velocity-adapted temporal derivative operators
$\partial_{\bar t}^n$ according to
\begin{equation}
  \label{eq-vel-adapt-temp-der}
  \partial_{\bar t}^n = (\partial_t + v_1 \, \partial_{x_1} + v_2 \, \partial_{x_2}),
\end{equation}
leading to joint spatio-temporal derivative operators ${\cal D}_{x,t}$
of the form
\begin{equation}
  {\cal D}_{x,t} = {\cal D}_x \, \partial_{\bar t}^n,
\end{equation}
or alternatively using space-time separable (not velocity-adapted)
spatio-temporal derivative
operators of the form
\begin{equation}
  {\cal D}_{x,t} = {\cal D}_x \, \partial_t^n.
\end{equation}

\subsection{Visualizations of spatial and spatio-temporal receptive fields}

Figures~\ref{fig-affgaussders}--\ref{fig-timecaus-strfs}
show sample illustrations of receptive fields according to the generalized Gaussian
derivative model for visual receptive fields.
Figure~\ref{fig-affgaussders} shows purely spatial receptive fields
of the form
\begin{equation}
  T_{\varphi^m}(x;\; s, \Sigma)
  = \partial_{\varphi^m} g(x;\; s, \Sigma),
\end{equation}
obtained by applying directional derivative operators of the form
(\ref{eq-def-spat-dir-der}) of orders $m = 1$ and $m = 2$ to
affine Gaussian kernels according to (\ref{eq-gauss-fcn-2D}),
based on the specific parameterization of the elements in the
spatial covariance matrix $\Sigma$ according to 
 \begin{align}
    \begin{split}
       \label{eq-expl-par-Cxx}
       \Sigma_{11} & = \lambda_1 \cos^2 \varphi + \lambda_2 \sin^2 \varphi,
    \end{split}\\
    \begin{split}
          \label{eq-expl-par-Cxy}
        \Sigma_{12} & = (\lambda_1 - \lambda_2)  \cos \varphi  \, \sin \varphi,
    \end{split}\\
    \begin{split}
       \label{eq-expl-par-Cyy}
        \Sigma_{22} & = \lambda_1  \sin^2 \varphi + \lambda_2  \cos^2 \varphi,
   \end{split}
\end{align}
with the eigenvalues $\lambda_1$ and $\lambda_2$ of
the spatial covariance matrix $\Sigma$
parameterized in terms of the corresponding
spatial standard deviations $\sigma_1$ and $\sigma_2$
according to
\begin{align}
  \begin{split}
       \label{eq-expl-par-lambda1}    
       \lambda_1 & = \sigma_1^2,
    \end{split}\\
    \begin{split}
       \label{eq-expl-par-lambda2}        
       \lambda_2 & = \sigma_2^2.
   \end{split}
\end{align}
This then leads to the following explicit expression for the affine
Gaussian derivative kernel
(Lindeberg \citeyear{Lin24-JMIV} Equations~(163) and~(164))
\begin{equation}
  \label{eq-def-aff-gauss-cont}
  g_{\text{aff}}(x, y;\; \sigma_1, \sigma_2, \varphi) =
  \frac{1}{2 \pi \sigma_1 \sigma_2} \,
  e^{-A/2 \, \sigma_1^2 \, \sigma_2^2},
\end{equation}
where
\begin{multline}
  \label{eq-def-aff-gauss-cont-arg}  
  A = (\sigma_2^2 \, x_1^2 + \sigma_1^2 \, x_2^2)  \cos^2 \varphi  
        + (\sigma_1^2 \, x_1^2 + \sigma_2^2 \, x_2^2) \sin^2 \varphi  \\
        - 2 \, (\sigma_1^2 - \sigma_2^2) \, x_1 \, x_2 \, \cos \varphi \, \sin \varphi.
\end{multline}
The eccentricity of such a kernel, which represents its degree of
elongation, is defined as the ratio
\begin{equation}
  \epsilon = \frac{\sigma_2}{\sigma_1}.
\end{equation}
Figures~\ref{fig-noncaus-strfs}--\ref{fig-timecaus-strfs}
show spatio-temporal receptive fields
corresponding to the mixed spatio-temporal derivative
\begin{equation}
  \label{eq-mixed-strf-1-1}
  T_{x\bar{t}}(x, t;\; s, \tau, v)
  = \partial_x \partial_{\bar{t}} \left ( g(x - v \, t;\; s) \, h(t;\; \tau) \right)
\end{equation}
over a 1+1-D spatio-temporal domain, where the 2-D affine Gaussian
kernel (\ref{eq-gauss-fcn-2D}) reduces to a 1-D Gaussian kernel
\begin{equation}
  \label{eq-1D-spat-gauss}
  g(x;\, s) = \frac{1}{\sqrt{2 \pi \, s}} \, e^{-x^2/2s}.
\end{equation}
Figure~\ref{fig-noncaus-strfs} shows such receptive fields in the case of a non-causal
temporal domain, where the temporal smoothing is performed with
the non-causal temporal Gaussian kernel according to (\ref{eq-non-caus-temp-gauss}),
whereas Figure~\ref{fig-timecaus-strfs} shows such receptive fields in the case of a
time-causal temporal domain, where the temporal smoothing is performed with
the time-causal limit kernel according to (\ref{eq-time-caus-lim-kern}),

\subsection{Relations to scale-normalized derivatives}

In Lindeberg
(\citeyear{Lin21-Heliyon,Lin25-JMIV,Lin25-arXiv-cov-props-review}), 
so-called unnormalized spatio-temporal derivative expressions of the
above forms are additionally combined with different forms of scale
normalization, as depending on the
actual values of the spatial scale parameter $s$, the spatial
covariance matrix $\Sigma$ and the temporal scale parameter $\tau$,
to define scale-normalized spatial and spatio-temporal derivative
operators%
\footnote{For in-depth treatments about how to extend the regular
  unnormalized spatial and spatio-temporal derivative operators
  ${\cal D}_x$ and ${\cal D}_{x,t}$ to corresponding scale-normalized
  spatial and spatio-temporal derivative operators, that make the full
  derivative-based receptive field responses in (\ref{eq-spat-rf-response}) and
  (\ref{eq-spattemp-rf-response}) provably covariant under the
  composed geometric image transformations in
  (\ref{eq-spat-geom-img-transf}), (\ref{eq-spattemp-geom-img-transf}),
  (\ref{eq-x-transf}) and (\ref{eq-t-transf}),
  see Lindeberg (\citeyear{Lin25-JMIV,Lin25-arXiv-cov-props-review}).
  Basically, to achieve covariance under with respect to uniform spatial
  scaling transformations of the form $x' = S_x \, x$, where $S_x \in \bbbr_+$ is
  a spatial scaling factor, the partial derivative operators
  $\partial_{x_i}$ in the spatial gradient operator $\nabla_x$
  according to
  (\ref{eq-def-spat-grad}) and the spatial Hessian operator ${\cal H}_x$
  according to (\ref{eq-def-spat-hess}) should be replaced by regular
  scale-normalized spatial derivatives according to
  $\partial_{x_i,\text{norm}} = s^{1/2} \, \partial_{x_i}$ according to
  Lindeberg (\citeyear{Lin97-IJCV}) Equation~(6) for the scale
  normalization power $\gamma = 1$, while the regular directional
  derivative operator $\partial_{\varphi}^m$ according to
  (\ref{eq-def-spat-dir-der}) should be replaced by the
  scale-normalized directional derivative operator $\partial_{\varphi,\text{norm}}^m$
  according to Lindeberg (\citeyear{Lin25-JMIV}) Equation~(33).
  To achieve covariance with respect to non-isotropic spatial affine
  transformations of the form $x' = A \, x$, where $A$ is a positive
  definite $2 \times 2$ affine transformation matrix, the regular gradient
  operator $\nabla_x$ according to (\ref{eq-def-spat-grad}) should be replaced
  by the scale-normalized affine gradient operator $\nabla_{x,\text{affnorm}}$ 
  according to Lindeberg (\citeyear{Lin25-JMIV}) Equation~(111), while
  the regular Hessian operator ${\cal H}_x$ according to
  (\ref{eq-def-spat-hess}) should be replaced by the
  scale-normalized affine Hessian operator ${\cal H}_{x,\text{affnorm}}$ 
  according to Lindeberg (\citeyear{Lin25-JMIV}) Equation~(140).
  To achieve both covariance under Galilean transformations of the
  form $x' = x + u \, t$, where $u = (u_1, u_2)^T \in \bbbr^2$ is a
  2-D motion vector, and covariance under temporal scaling
  transformations of the form $t' = S_t \, t$, where $S_t \in \bbbr_+$
  is a temporal scaling factor, the temporal differentiation of the
  receptive fields should be performed in terms of scale-normalized
  velocity-adapted temporal derivative operators obtained by
  replacing the regular velocity-adapted temporal derivative operator
  $\partial_{\bar{t}}^n$ in (\ref{eq-vel-adapt-temp-der}) by its
  scale-normalized correspondence $\partial_{\bar{t},\text{norm}}^n$
  according to Lindeberg (\citeyear{Lin25-JMIV}) Equation~(168).}%
\footnote{If using scale-normalized derivative operators in the models
  for the visual receptive fields, it should, however, be observed that the
  introduction of scale-normalized derivatives will substantially
  influence the evolution properties over scales of the receptive field
  responses. This follows, since the scale-dependent scale
  normalization factors used in the formulations of scale-normalized
  derivative operators, as extensively described in
  Lindeberg (\citeyear{Lin25-JMIV}), would then also need to be
  differentiated with respect to the spatial and the temporal scale
  parameters. For this reason, we do in this paper isolate the
  evolution properties over the filter parameters to only receptive
  field models expressed in terms of regular, not scale-normalized,
  spatial and/or spatio-temporal derivative expressions.
  If the use of scale-normalized derivatives is warranted, as it can
  be because of the requirement of making it possible to match the
  output of derivative-based receptive field operators on the image
  data, this can be easily accomplished by performing the scale
  normalization in a second stage, after the computation of regular,
  not scale-normalized, derivative operators in a first processing stage.}
that are provably scale covariant, in the sense that the
magnitude of the filter responses can be perfectly matched under specific
families of geometric image transformations.
In this treatment, we will, however, not consider the effects of such
scale normalization, and will instead focus solely on the effects of the
either purely spatial or joint spatio-temporal smoothing operations,
obtained by convolution with kernels of the forms
(\ref{eq-gauss-fcn-2D}) or (\ref{eq-form-spat-temp-kernel}) in
combination with either purely spatial or joint spatio-temporal
derivative computations, in turn, leading to receptive field responses of the
forms (\ref{eq-spat-rf-response}) or (\ref{eq-spattemp-rf-response}).
For the task of relating the responses of linear receptive fields
of the forms (\ref{eq-spat-rf-response}) or
(\ref{eq-spattemp-rf-response}), such an analysis is sufficient,
provided that the spatial differentiation operators ${\cal D}_x$
and the spatio-temporal differential operators ${\cal D}_{x,t}$
in these receptive field models are purely linear operators.

\subsection{Covariance properties of the spatial and the
  spatio-temporal smoothing kernels}
\label{eq-cov-props}

With regard to variabilities of the shapes of the spatial or the
spatio-temporal receptive fields in relation to the variabilities in
spatial or spatio-temporal image transformations involved in the image
formation process, that determines the structures of the either
purely spatial or joint spatio-temporal image data $f$, an important
consequence of the formulation of the spatial smoothing kernel
$T(x;\; s, \Sigma)$ according to (\ref{eq-gauss-fcn-2D})
and the formulation of the joint spatio-temporal smoothing kernel
$T(x, t;\; s, \Sigma, \tau, v)$ according to
(\ref{eq-form-spat-temp-kernel})
is that these forms of convolution kernels are covariant under
important basic classes of geometric image transformations.

\begin{figure}[btp]
  \vspace{-40mm}
  \begin{center}
     \includegraphics[width=0.50\textwidth]{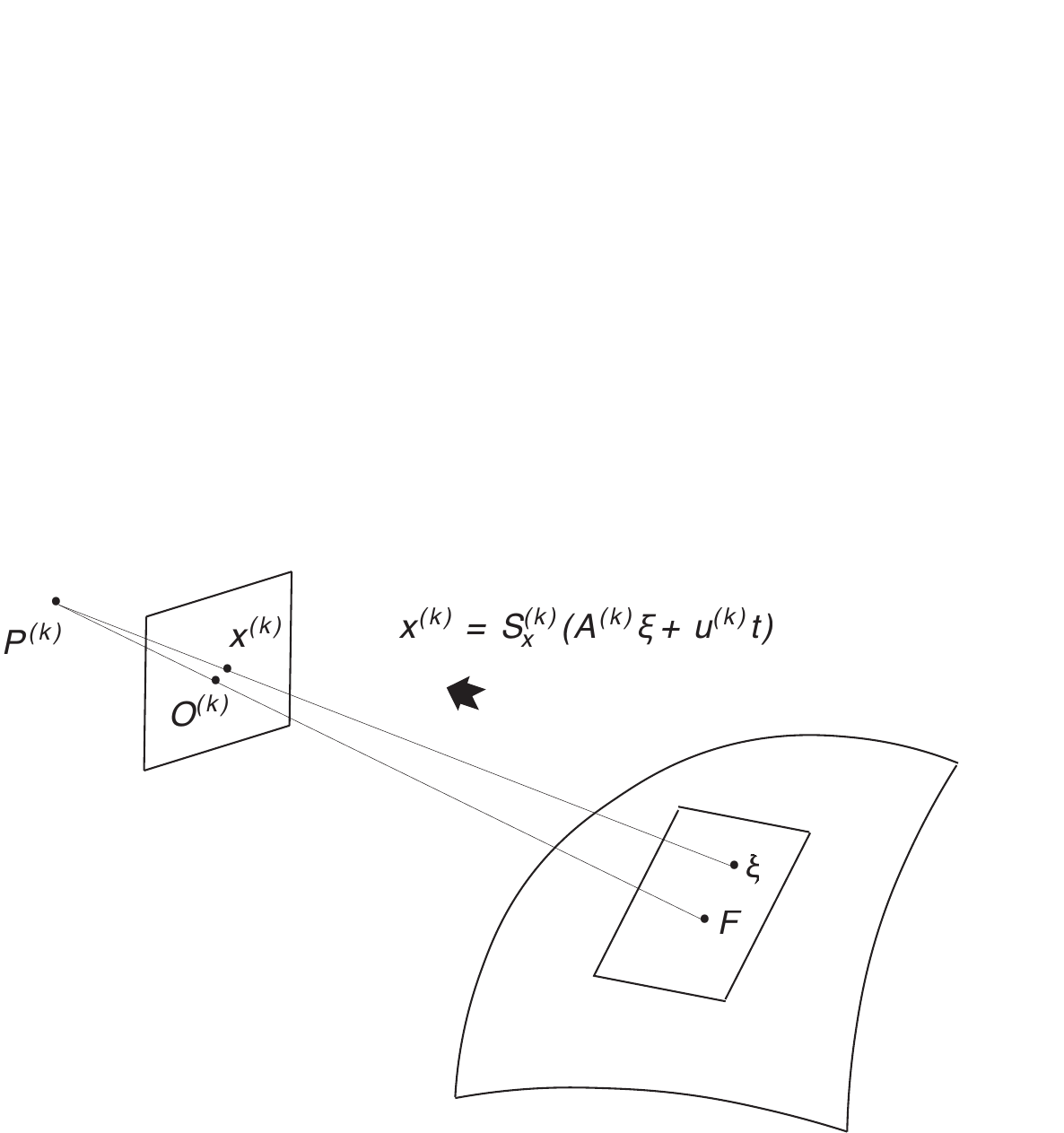} 
   \end{center}
   \caption{Illustration of the geometry underlying the composed
     locally linearized projection models in
     Equation~(\ref{eq-spat-geom-img-transf}) and
     Equations~(\ref{eq-spattemp-geom-img-transf})--(\ref{eq-t-transf}),
     when extended to a multi-view imaging situation, with each view
     indexed by $k$.
     We consider a local, in the spatio-temporal case
     possibly moving, surface patch, which is projected to an
     arbitrary view in a multi-view locally linearized
     projection model, with
     the fixation point $F$ on the surface mapped to the origin
     $O^{(k)} = 0$ in the image plane for the observer with the
     optic center $P^{(k)}$. Then, any point in the
     tangent plane to the surface at the fixation point, as
     parameterized by the local coordinates $\xi$ in a coordinate
     frame attached to the tangent plane of the surface with $\xi = 0$
     at the fixation point $F$, is by the
     local linearization mapped to the image point $x^{(k)}$.
     (Figure reproduced from Lindeberg (\citeyear{Lin25-JMIV}) with
     permission (OpenAccess).)}
   \label{fig-singlegeom}
 \end{figure}
 
\subsubsection{Purely spatial covariance properties}
\label{sec-spat-cov-prop}

For the case of a purely spatial domain, let us consider an image
transformation of the form
\begin{equation}
  \label{eq-spat-geom-img-transf}
  f'(x') = f(x) \quad\quad\mbox{for}\quad\quad x' = S_x \, A,
\end{equation}
where
\begin{itemize}
\item
  $S_x \in \bbbr_+$ is a spatial scaling factor,
\item
  $A$ is a non-singular $2 \times 2$ matrix, that we could assume to
  be normalized in a way as corresponding to orthographic
  projection of a smooth local surface patch.
\end{itemize}
The geometric motivation for using this model is that
(see Figure~\ref{fig-singlegeom} for an illustration):
\begin{itemize}
\item
  the spatial scaling
  factor $S_x$ does to first order of approximation reflect the size
  variations caused by varying the distance between the object and the
  observer, whereas
\item
  the spatial affine transformation matrix $A$
does to first order of approximation reflect the image deformations
caused by varying the viewing direction for a visual observer that
views a smooth local surface patch.
\end{itemize}
Then, if we define purely spatial scale-space representations
$L \colon \bbbr^2 \times \bbbr_+ \times \bbbs_+^2 \rightarrow \bbbr$
and
$L' \colon \bbbr^2 \times \bbbr_+ \times \bbbs_+^2 \rightarrow \bbbr$
of the purely spatial images $f$ and $f'$, respectively,  according to
\begin{align}
  \begin{split}
    L(\cdot;\; s, \Sigma) = T(\cdot;\; s, \Sigma) * f(\cdot),
  \end{split}\\
  \begin{split}
    L'(\cdot;\; s', \Sigma') = T(\cdot;\; s', \Sigma') * f'(\cdot),
  \end{split}
\end{align}
it follows that the scale-space representations can be perfectly matched under
the geometric image transformation (\ref{eq-spat-geom-img-transf})
\begin{equation}
  L'(x';\; s', \Sigma') = L(x;\; s, \Sigma),
\end{equation}
provided that the parameters of the receptive fields are matched
according to
Lindeberg (\citeyear{Lin25-JMIV}) Equations~(252)--(253):
\begin{align}
  \begin{split}
    \label{eq-s-transf-result}
    s' & = S_x^2 \, s,
  \end{split}\\
  \begin{split}
    \label{eq-Sigma-transf-result}
    \Sigma' & = A \, \Sigma \, A^{T}.
  \end{split}
\end{align}
Due to the overparameterization of the degrees of freedom spanned by
the spatial scale parameter $s$ and the spatial covariance matrix
$\Sigma$, however, a minimal necessity requirement can also be stated
in terms of the following combined matching criterion
(Lindeberg \citeyear{Lin25-JMIV} Equation~(118)):
\begin{equation}
  \label{eq-transf-prop-sc-par-spat-cov-mat-pure-aff-scsp-full}
  s' \, \Sigma' = s \, (S_x \, A) \, \Sigma \, (S_x A)^T = s \, S_x^2 \, A \, \Sigma \, A^T.
\end{equation}

\subsubsection{Joint spatio-temporal covariance properties}
\label{sec-spattemp-cov-prop}

For the case of a joint spatio-temporal domain, let us consider a
joint spatio-temporal image transformation of the form
\begin{equation}
  \label{eq-spattemp-geom-img-transf}
  f'(x', t') = f(x, t) 
\end{equation}
for
\begin{align}
  \begin{split}
     \label{eq-x-transf}
     x' = S_x \, (A \,  x + u \, t),
   \end{split}\\
  \begin{split}
     \label{eq-t-transf}
     t' = S_t \, t,
   \end{split}
\end{align}
where for the entities not already defined in connection with
Equation~(\ref{eq-spat-geom-img-transf}), we have that
\begin{itemize}
\item
  $u = (u_1, u_2)^T \in \bbbr^2$ is a 2-D velocity vector, that may
  correspond to the orthographic projection of the 3-D motion field
  $U$ of the local surface patch observed by the vision system, and
\item
  $S_t \in \bbbr_+$ is a temporal scaling factor.
\end{itemize}
The geometric motivation for using this model is, beyond the reasons
for using a combination of uniform spatial scaling transformations and
spatial affine transformations according to the previously stated
motivation in Section~\ref{sec-spat-cov-prop}, that
(see Figure~\ref{fig-singlegeom} for an illustration):
\begin{itemize}
\item
  the Galilean
  transformation of the form $x' = x + u \, t$ models the effect of
  relative motions between the observed objects or spatial-temporal
  events and the viewing direction, whereas
\item
  the temporal scaling factor
  $S_t$ models the effect of observing similar spatio-temporal events,
  that may occur either faster or slower relative to a previously
  observed reference view.
\end{itemize}
Then, if we define the joint spatio-temporal scale-space representations
$L \colon \bbbr^2 \times \bbbr \times \bbbr_+ \times \bbbs_+^2 \times \bbbr_+
\times \bbbr^2 \rightarrow \bbbr$
and
$L' \colon \bbbr^2 \times \bbbr \times \bbbr_+ \times \bbbs_+^2 \times \bbbr_+
\times \bbbr^2 \rightarrow \bbbr$
of the video sequences or video streams $f$ and $f'$, respectively,  according to
\begin{align}
  \begin{split}
    L(\cdot, \cdot;\; s, \Sigma, \tau, v)
    = T(\cdot, \cdot;\; s, \Sigma, \tau, v) * f(\cdot, \cdot),
  \end{split}\\
  \begin{split}
    L'(\cdot, \cdot;\; s', \Sigma', \tau', v')
    = T(\cdot, \cdot;\; s', \Sigma', \tau', v') * f'(\cdot, \cdot),
  \end{split}
\end{align}
it follows that the scale-space representations can be perfectly matched under
the composed geometric image transformation given by
(\ref{eq-spattemp-geom-img-transf}), (\ref{eq-x-transf}) and (\ref{eq-t-transf})
\begin{equation}
  L'(x', t';\; s', \Sigma', \tau', v') = L(x, t;\; s, \Sigma, \tau, v),
\end{equation}
provided that the purely spatial parameters
$s$, $s'$, $A$ and $A'$ are matched
according to (\ref{eq-s-transf-result})
and (\ref{eq-Sigma-transf-result}) and additionally the temporal scale
parameters $\tau$ and $\tau'$ as well as the velocity parameters
$v$ and $v'$ are matched according to
Lindeberg (\citeyear{Lin25-JMIV}) Equations~(254)--(255):
\begin{align}
  \begin{split}
    \label{eq-tau-transf-result}
    \tau' & = S_t^2 \, \tau,
  \end{split}\\
  \begin{split}
    \label{eq-v-transf-result}    
    v' & = \frac{S_x}{S_t} (A \, v + u).
  \end{split}
\end{align}

\subsection{Motivation for filter banks of receptive field responses}

A main consequence of the above covariance properties is therefore that
from a requirement of the vision system to be covariant under the
basic classes of geometric image transformations, if we want to have
the ability to match the effects of the either purely spatial or joint
spatio-temporal smoothing operations, that occur in the models for the
receptive fields, then we have to have the ability to make use of
receptive field responses computed for matching sets of filter parameters,
either $(s, \Sigma)$ or $(s', \Sigma')$ in the purely spatial case
or $(s, \Sigma, \tau, v)$ and $(s', \Sigma', \tau', v')$ in the joint
spatio-temporal case.

Thus, the requirement of establishing an
identity between receptive field responses under the influence
of {\em a priori\/} unknown geometric image transformations, caused by
different viewing conditions between different observations of the
same scene or a similar type of spatio-temporal event, leads to the
strategy of basing the filtering operations in an idealized vision
system on filter banks with receptive field responses computed for a
large variability of the filter parameters.
This theoretical background forms the conceptual foundation for the main
topic in this paper of relating receptive field responses computed for
different values of the filter parameters of the receptive fields.

\section{Infinitesimal hybrid Lie semi-group structures for the
  generalized Gaussian derivative model}
\label{sec-inf-rels-non-caus-rfs}

\begin{figure}[hbt]
  \begin{center}
    \includegraphics[width=0.45\textwidth]{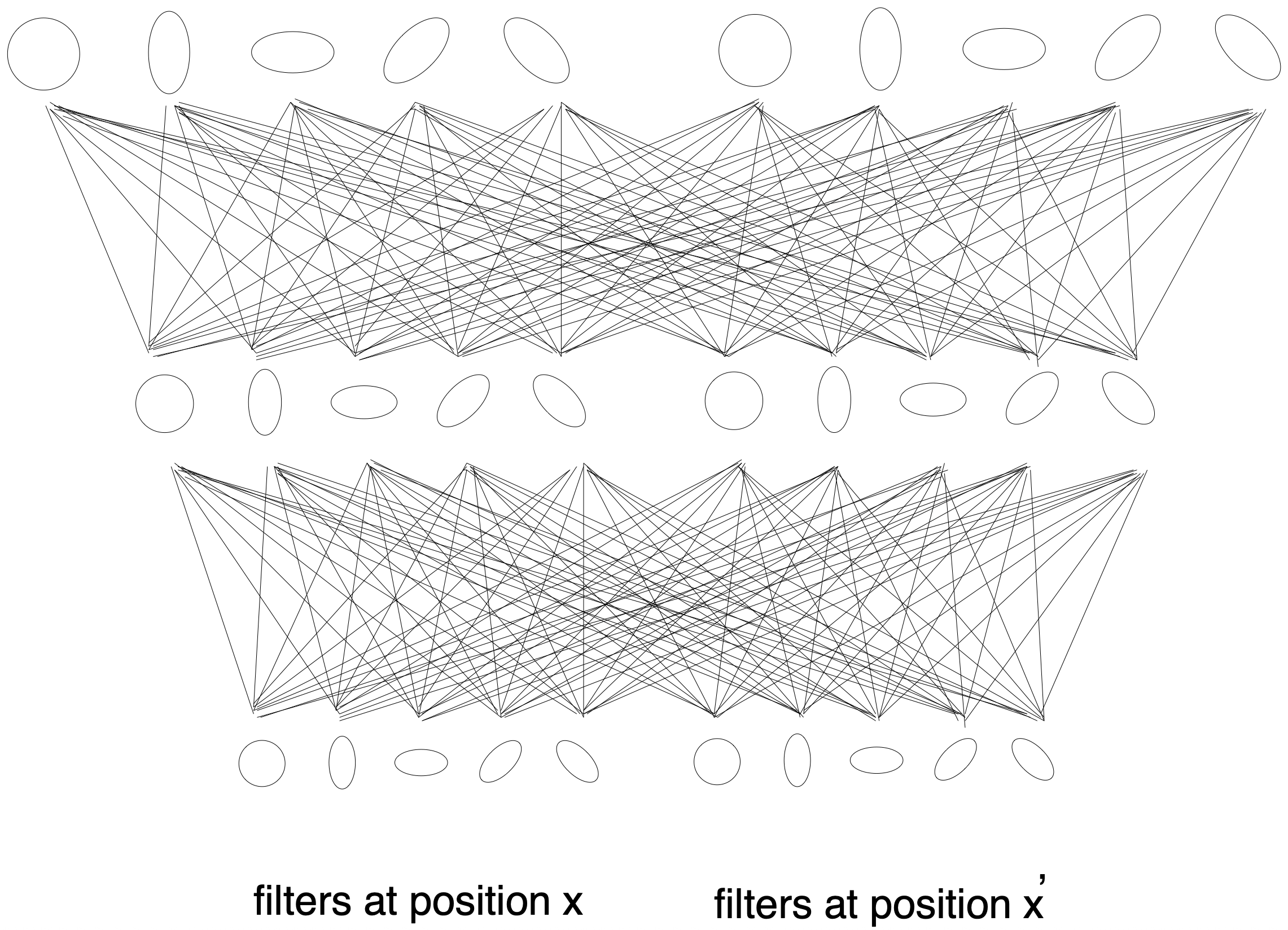}
  \end{center}
  \caption{Schematic illustration of how receptive fields for
    different parameter values and at different image positions can be
    interrelated to each other, as constituting the main subject of study
    in this paper. Here, we derive such relationships between
    receptive field responses obtained for different values of the
    shape parameters of the receptive fields, either in terms of
    (i)~infinitesimal relationships closely related to the notions of Lie
    groups and infinitesimal generators for semi-groups, and
    (ii)~macroscopic relationships in terms of cascade smoothing
    properties, with close relations to the notion of Lie algebras, as
    can be related to Lie groups {\em via\/} the exponential
    map. Compared to regular Lie groups and Lie algebras, some of the
    evolution relations do, however, have a directional preference,
    implying that the evolution can only be performed in a single
    direction, and not in the reverse direction. (In this figure, each
    line represents a connection between two receptive fields for
    different values of their parameters and/or their spatial positions.)}
  \label{fig-conn-rels}
\end{figure}

Given the above formulations of models for either purely spatial or
joint spatio-temporal receptive fields in terms of either purely
spatial or joint spatio-temporal derivative expressions applied to
either purely spatial or joint spatio-temp\-oral smoothing kernels,
as obtained from multi-parameter scale-space representations, one may ask
how such receptive field responses are related between different
values of the filter parameters. Specifically, one may ask if and then
how a receptive field response at a coarser level of spatial and
temporal scales may be computed from receptive field responses at
finer spatial and temporal scales, as schematically
illustrated in Figure~\ref{fig-conn-rels}.

In this section, we will derive a set of differential 
relationships that show how either spatial receptive field responses
\begin{equation}
  {\cal R} f(\cdot)
  = {\cal D}_x T(\cdot;\; s, \Sigma) * f(\cdot)
\end{equation}
or spatio-temporal receptive field responses
\begin{equation}
  {\cal R} f(\cdot, \cdot)
  = {\cal D}_{x,t} T(\cdot, \cdot;\; s, \Sigma, \tau, v) * f(\cdot, \cdot)
\end{equation}
computed for different values of the filter parameters
$(s, \Sigma)$ or $(s, \Sigma, \tau, v)$ are related under
infinitesimal perturbations of the filter parameters, in the
respective cases of either purely spatial receptive fields or joint
spatio-temporal receptive fields, initially restricted to the
situation when both the spatial and the temporal smoothing kernels are
based on Gaussian kernels, and therefore differentiable with respect
to all of their respective parameters.%
\footnote{When using the time-causal limit kernel
  $h(t;\; \tau) = \psi(t;\; \tau, c)$ according to
  (\ref{eq-time-caus-lim-kern}) as the temporal smoothing kernel in
  the spatio-temporal smoothing kernel $T(x, t;\; s, \Sigma, \tau,
  v)$, an important restriction is that the temporal scale levels are
  genuinely discrete. Therefore, the corresponding temporal or
  spatio-temporal smoothing kernels cannot be differentiated with
  respect to the discrete scale parameter $\tau_k = \tau_0 \, c^{2k}$
  according to (\ref{eq-disc-temp-sc-limit-kern}).}

To a major extent, this will be structurally similar  to the
notion of a Lie group structure over
the filter parameters, as expressed in terms of relations between
derivatives with respect to the filter parameters
$(s, \Sigma)$ or $(s, \Sigma, \tau, v)$ in relation to derivatives
with respect to the either spatial or joint spatio-temporal image
coordinates $x$ or $(x, t)$. Since the evolution properties with
respect to the spatial scale parameter $s$ and the temporal scale
parameter $\tau$ can, however, only performed in the direction of increasing
values of the scale parameters, the corresponding transformations will therefore
not form a full group, but only a semi-group with regard to these two
dimensions of variabilities in the filter parameters.
For the derived infinitesimal evolution equations with respect to
these scale parameters, the resulting notions will therefore be closer
to the notion of infinitesimal generators for semi-groups over
continuous evolution parameters.
For two of the
dimensions of the variability, concerning the mixed element
$\Sigma_{12}$ in the spatial covariance matrix $\Sigma$
as well as the variabilities with respect to the elements $v_1$ and
$v_2$ in the velocity vector $v$, the
variabilities are, on the other hand, not as constrained.
For these reasons, we will refer to the resulting structures,
to be derived, as hybrid Lie semi-group structures. 

Before starting deriving the resulting infinitesimal relationships
over the
different filter parameters, let us first describe the methodology
that we will use in this section, by exploiting the fact that for the purely spatial
receptive field model as well as for the joint spatio-temporal
receptive field model, for the case of a non-causal temporal domain, the convolution
kernels are all expressed in terms of higher-dimensional Gaussian
kernels. This implies that we can make use of similar algebraic properties
for performing the proofs as arise in the definition of Hermite
functions and Hermite polynomials, although here generalized to a
higher-dimensional setting for the specific forms of spatial or
spatio-temporal smoothing kernels, as occur in our theoretically
principled models for visual receptive fields.

\subsection{Methodology to be used}

For the one-dimensional Gaussian kernel, it is
well-known that the computation of derivatives of this kernel corresponds to
multiplying the Gaussian function with Hermite polynomials.
For our way of parameterizing the Gaussian kernel, the
probabilistic Hermite polynomials
$\operatorname{He}_m(x)$, defined by
\begin{equation}
\operatorname{He}_m(x) = (-1)^m e^{x^2/2} \, \partial_{x^m} \left( e^{-x^2/2} \right),
\end{equation}
imply that
\begin{equation}
\partial_{x^m} \left( e^{-x^2/2} \right) =  (-1)^m \operatorname{He}_m(x) \, e^{-x^2/2} 
\end{equation}
and
\begin{equation}
  \partial_{x^m} \left( e^{-x^2/2s} \right)
  =  (-1)^m \operatorname{He}_m(\frac{x}{\sqrt{s}}) \,
       e^{-x^2/2s} \frac{1}{\sqrt{s}^m}.
\end{equation}
This means that the $n$:th-order Gaussian derivative kernel in 1-D can be
written as
\begin{align}
  \begin{split}
    \partial_{x^m} \left( g(x;\; s) \right) 
    = \frac{1}{\sqrt{2 \pi} \sqrt{s}} \, \partial_{x^m} \left( e^{-x^2/2s} \right) 
  \end{split}\nonumber\\
  \begin{split}
    = \frac{1}{\sqrt{2 \pi} \sqrt{s}}  \frac{(-1)^m}{\sqrt{s}^m}
    \operatorname{He}_m(\frac{x}{\sqrt{s}}) \, e^{-x^2/2s}
  \end{split}\nonumber\\
  \begin{split}
    \label{eq-gauss-der-herm-pol}
    = \frac{(-1)^m}{\sqrt{s}^m} \operatorname{He}_m(\frac{x}{\sqrt{s}}) \, g(x;\; s),
  \end{split}
\end{align}
in other words as a polynomial multiplied with the original Gaussian
function.

In this section, we will derive corresponding generalizations of Hermite
polynomials for the 2-D and 2+1-D Gaussian kernels, that correspond to
the either purely spatial or joint spatio-temporal smoothing
operations in the spatial and spatio-temporal receptive field models
according to (\ref{eq-spat-rf-response}) and (\ref{eq-spattemp-rf-response}).

Specifically, we will frequently make use of the property that if
two different differential operators give rise to the same generalized
Hermite polynomials, when applied to the either purely spatial or
joint spatio-temporal smoothing kernels, then also the corresponding
spatial or spatio-temporal scale-space representations have to be equal,
as well as any spatial or temporal derivative operator, or any
first-order differentiation operator with respect to a parameter of
the spatial or temporal smoothing kernel.

Stated more formally in a compact manner, let
\begin{itemize}
\item
  $p$ denote either 2-D spatial or 2+1-D spatio-temporal image coordinates,
  $p = (x_1, x_2)^T$ or $p = (x_1, x_2, t)^T$,
\item
  $g(p;\; P)$ denote a corresponding
  either 2-D spatial or joint 2+1-D spatio-temporal smoothing kernel
  depending on a set of filter parameters $P$,
\item
  $\partial_{P_i}$ denote a partial derivative
  operator with respect to the $i$:th filter parameter, and
\item
  ${\cal D}_{i,p}$ denote a possibly composed differentiation operator
  with respect to the image coordinates $p$.
\end{itemize}
Then, if we for the partial derivative operator $\partial_{P_i}$ with
respect to the filter parameter $P_i$, are able to show that this
partial derivative operator does for the either purely spatial or
joint spatio-temporal smoothing function
$g(p;\; P)$ for the corresponding either spatial or joint
spatio-temporal scale-space representation
\begin{equation}
  L(\cdot;\; P) = g(\cdot;\; P) * f(\cdot)
\end{equation}
of the either purely spatial or joint spatio-temporal input image
$f \colon \bbbr^2 \rightarrow \bbbr$ or
$f \colon \bbbr^2 \times \bbbr \rightarrow \bbbr$
satisfy
\begin{equation}
  \partial_{P_i} g(p;\; P) = {\cal D}_{i,p} g(p;\; P),
\end{equation}
then it follows from the linearity of the scale-space representation
$L$, that
\begin{multline}
  \partial_{P_i} L(\cdot;\; P) 
  = \partial_{P_i} g(\cdot;\; P) * f(\cdot) = \\
  = {\cal D}_{i,p} g(\cdot;\; P) * f(\cdot)
  = {\cal D}_{i,p} L(\cdot;\; P).
\end{multline}
Similarly, for any linear either purely spatial receptive field operator
${\cal R}$ defined
from applying a spatial differential operator ${\cal D}_x$ to a
purely spatial scale-space representation
\begin{equation}
  {\cal R} f(\cdot) = {\cal D}_{x} L(\cdot;\; P),
\end{equation}
or for any linear joint spatio-temporal receptive field operator
${\cal R}$ defined
from applying a spatio-temporal differential operator ${\cal D}_{x,t}$ to a
the joint spatio-temporal scale-space representation
\begin{equation}
  {\cal R} f(\cdot) = {\cal D}_{x,t} L(\cdot;\; P),
\end{equation}
it holds that
\begin{equation}
  \partial_{P_i} ({\cal R} f) =  {\cal D}_{i,p} ({\cal R} f).
\end{equation}
Specifically, since for all these generalized Gaussian derivative
models for visual receptive fields, the result of applying the
differential operators to the either 2-D purely spatial or 2+1-D
joint spatio-temporal Gaussian kernels will be generalized Hermite
functions multiplied by the underlying Gaussian kernels
\begin{align}
  \begin{split}
    \partial_{P_i} g(p;\; P)  = H_{P_i}(p;\; P) \,  g(p;\; P),
  \end{split}\\
  \begin{split}
    {\cal D}_{i,p} g(p;\; P)  = H_{i,p}(p;\; P) \,  g(p;\; P),
  \end{split}
\end{align}
implying that the respective generalized Hermite polynomials are given
by
\begin{align}
  \begin{split}
    H_{P_i}(p;\; P) = \frac{\partial_{P_i} g(p;\; P)}{ g(p;\; P)},
  \end{split}\\
  \begin{split}
    H_{i,p}(p;\; P) = \frac{{\cal D}_{i,p} g(p;\; P)}{g(p;\; P)},
  \end{split}
\end{align}
it will for the purposes of the following proofs to be performed be
sufficient to show that the corresponding generalized Hermite
polynomials are equal
\begin{equation}
   H_{P_i}(p;\; P) = H_{i,p}(p;\; P),
\end{equation}
in order to conclude that a differentiation operator ${\cal D}_{i,p}$
with respect to the image coordinates $p$
for the first-order partial derivative $\partial_{P_i}$ with respect to
the filter parameter $P_i$ of the receptive field is given by
\begin{equation}
  \label{eq-general-inf-rel-ders-pars-spattemp-ders}
  \partial_{P_i} = {\cal D}_{i,p}.
\end{equation}
Specifically, if we let the operator ${\cal D}_p$ denote the either
purely spatial differentiation operator ${\cal D}_x$ used for
computing a spatial-derivative-based purely spatial receptive field response according to
(\ref{eq-spat-rf-response}), or alternatively using the same notation
${\cal D}_p$ for denoting a spatio-temporal-derivative-based joint
spatio-temporal receptive field response ${\cal D}_{x,t}$ according to
(\ref{eq-spattemp-rf-response}),
such that those receptive field responses can be combined on the
compact form
\begin{equation}
  \label{eq-unified-spat-or-spattemp-rf-response}
  {\cal R} \, f(\cdot)  = {\cal D}_p (g(;\; P) * f(\cdot)),
\end{equation}
then it follows from the linearity of the involved differentiation
operators, that the infinitesimal relationships between differentiation
with respect to a filter parameter $\partial_{P_i}$ and the
corresponding either purely spatial or joint spatio-temporal
differentiation operator ${\cal D}_{i,p}$
according to (\ref{eq-general-inf-rel-ders-pars-spattemp-ders})
generalize to corresponding receptive field responses according to
\begin{equation}
  \partial_{P_i} \, {\cal R} \, f(\cdot) = {\cal D}_{i,p} \, {\cal R} \, f(\cdot).
\end{equation}
Based in this general property, we will in the following
Sections~\ref{sec-lie-group-spat}--\ref{sec-lie-group-spattemp-aff}
focus solely on the properties of the generalized Hermite polynomials,
that arise from differentiating the either purely spatial or joint
spatio-temporal smoothing kernels in the idealized receptive field models
with respect to the image coordinates and the filter parameters
for the respective cases of: (i)~purely spatial receptive fields,
(ii)~joint spatio-temporal receptive fields based on smoothing with
spatially isotropic Gaussian kernels, for which the spatial covariance
matrix $\Sigma$ reduces to a unit matrix $\Sigma = I$, and (iii)~joint spatio-temporal
receptive fields based on smoothing with possibly anisotropic affine
Gaussian kernels, for which we assume that the spatial covariance
matrix $\Sigma$ can generally assume values not equal to a unit matrix
$I$.

By dividing the analysis into these three types of receptive field
models, we specifically begin with the simpler purely spatial
model, in order to then increase the complexity via the isotropic
spatio-temporal model towards the fully
affine spatio-temporal model, which obeys covariance properties over
the largest set of geometric image transformations considered in this study.

\begin{figure*}[hbtp]
  \begin{align}
    \begin{split}
      & \frac{\partial_{x_1} g(x;\; s, \Sigma)}{g(x;\; s, \Sigma)}
      = \frac{\Sigma_{22} x_1-\Sigma_{12} x_2}{\Sigma_{12}^2 s-\Sigma_{11} \Sigma_{22} s},
    \end{split}\\
    \begin{split}
      & \frac{\partial_{x_2} g(x;\; s, \Sigma)}{g(x;\; s, \Sigma)}
      = \frac{\Sigma_{11} x_2-\Sigma_{12} x_1}{s \left(\Sigma_{12}^2-\Sigma_{11}
   \Sigma_{22}\right)},
    \end{split}\\
    \begin{split}
      & \frac{\partial_{x_1 x_1} g(x;\; s, \Sigma)}{g(x;\; s, \Sigma)} 
        = \frac{\Sigma_{22}^2 \left(x_1^2-\Sigma_{11} s\right)+\Sigma_{12}^2 \left(\Sigma_{22}
   s+x_2^2\right)-2 \Sigma_{12} \Sigma_{22} x_1 x_2}{s^2
   \left(\Sigma_{12}^2-\Sigma_{11} \Sigma_{22}\right)^2},
    \end{split}\\
    \begin{split}
      & \frac{\partial_{x_1 x_2} g(x;\; s, \Sigma)}{g(x;\; s, \Sigma)} 
        = \frac{\Sigma_{12} \left(\Sigma_{11} \Sigma_{22} s-\Sigma_{11} x_2^2-\Sigma_{22}
   x_1^2\right)+\Sigma_{11} \Sigma_{22} x_1 x_2+\Sigma_{12}^3
   (-s)+\Sigma_{12}^2 x_1 x_2}{s^2 \left(\Sigma_{12}^2-\Sigma_{11}
   \Sigma_{22}\right)^2},
    \end{split}\\
    \begin{split}
      & \frac{\partial_{x_2 x_2} g(x;\; s, \Sigma)}{g(x;\; s, \Sigma)} 
        = \frac{\Sigma_{11}^2 \left(x_2^2-\Sigma_{22} s\right)+\Sigma_{11} \Sigma_{12}
   (\Sigma_{12} s-2 x_1 x_2)+\Sigma_{12}^2 x_1^2}{s^2
   \left(\Sigma_{12}^2-\Sigma_{11} \Sigma_{22}\right)^2},
    \end{split}\\
    \begin{split}
      & \frac{\partial_{s} g(x;\; s, \Sigma)}{g(x;\; s, \Sigma)}
      = -\frac{\Sigma_{11} \left(x_2^2-2 \Sigma_{22} s\right)+2 \Sigma_{12}^2 s-2 \Sigma_{12}
   x_1 x_2+\Sigma_{22} x_1^2}{2 s^2 \left(\Sigma_{12}^2-\Sigma_{11}
   \Sigma_{22}\right)},
    \end{split}\\
    \begin{split}
      & \frac{\partial_{\Sigma_{11}} g(x;\; s, \Sigma)}{g(x;\; s, \Sigma)}
      = \frac{\Sigma_{22}^2 \left(x_1^2-\Sigma_{11} s\right)+\Sigma_{12}^2 \left(\Sigma_{22}
   s+x_2^2\right)-2 \Sigma_{12} \Sigma_{22} x_1 x_2}{2 s
   \left(\Sigma_{12}^2-\Sigma_{11} \Sigma_{22}\right)^2},
    \end{split}\\
    \begin{split}
      & \frac{\partial_{\Sigma_{12}} g(x;\; s, \Sigma)}{g(x;\; s, \Sigma)}
      = \frac{\Sigma_{12} \left(\Sigma_{11} \Sigma_{22} s-\Sigma_{11} x_2^2-\Sigma_{22}
   x_1^2\right)+\Sigma_{11} \Sigma_{22} x_1 x_2+\Sigma_{12}^3
   (-s)+\Sigma_{12}^2 x_1 x_2}{s \left(\Sigma_{12}^2-\Sigma_{11}
   \Sigma_{22}\right)^2},
    \end{split}\\
    \begin{split}
      & \frac{\partial_{\Sigma_{22}} g(x;\; s, \Sigma)}{g(x;\; s, \Sigma)}
      = \frac{\Sigma_{11}^2 \left(x_2^2-\Sigma_{22} s\right)+\Sigma_{11} \Sigma_{12}
   (\Sigma_{12} s-2 x_1 x_2)+\Sigma_{12}^2 x_1^2}{2 s
   \left(\Sigma_{12}^2-\Sigma_{11} \Sigma_{22}\right)^2}.
    \end{split}
  \end{align}
  \caption{Generalized Hermite polynomials as arising from
    derivatives of the purely spatial affine Gaussian kernel
    $T(x;\; s, \Sigma) = g(x;\; s, \Sigma)$ according to
    (\ref{eq-gauss-fcn-2D}) with respect the image coordinates
    $x = (x_1, x_2)^T$ up to order 2, as well as with respect to the spatial
    scale parameter $s$ and the elements $\Sigma_{11}$,  $\Sigma_{12}$
    and  $\Sigma_{22}$ of the spatial covariance matrix $\Sigma$.}
  \label{fig-ders-spat-aff}
\end{figure*}

\subsection{Purely spatial receptive fields based on affine Gaussian smoothing}
\label{sec-lie-group-spat}

For the case of a purely spatial domain, let us consider the following
parameterization of the affine Gaussian kernel
\begin{equation}
  \label{eq-gauss-fcn-2D-again}
  g(x;\; s, \Sigma)
  = \frac{1}{2 \pi \, s \sqrt{\det \Sigma}} \, e^{-x^T  \Sigma^{-1} x/2 s},
\end{equation}
which leads to the corresponding affine Gaussian scale space of
any 2-D spatial image $f \colon \bbbr^2 \rightarrow \bbbr$:
\begin{equation}
  L(\cdot;\; s, \Sigma) = g(\cdot;\; s, \Sigma) * f(\cdot).
\end{equation}
The motivation why we overparameterize the effective variability in the
shapes of the receptive fields in this way is for reasons of geometric
interpretation with regard to viewing variations for a visual observer,
that may observe the same local surface patch from different distances
and viewing directions. If the affine matrix $A$ in a locally linearized
perspective viewing model
\begin{equation}
  f'(x') = f(x) \quad\quad\mbox{for}\quad\quad x' = S_x \, A \, x
\end{equation}
is normalized such that it corresponds to an a local orthonormal
projection, then by the requirement of covariance under the group of geometric
image transformations, a variability in depth $Z$ directly maps to a
variability in the spatial scaling factor $S_x$, which in terms of the
receptive fields will correspond to a variability in the spatial scale
parameter $s$. Similarly, a variability in the affine transformation
matrix $A$ will map to a variability in the spatial covariance matrix
$\Sigma$.

By differentiating the affine Gaussian kernel (\ref{eq-gauss-fcn-2D})
with respect to the image coordinates $x = (x_1, x_2)^T$ and the
parameters $s$ and the elements $\Sigma_{i,j}$ in the 2-D positive symmetric definite
covariance matrix 
\begin{equation}
  \Sigma =
  \left(
    \begin{array}{cc}
      \Sigma_{11} & \Sigma_{12} \\
      \Sigma_{12} & \Sigma_{22}
    \end{array}
  \right),
\end{equation}
we get the explicit expressions for these derivatives according to
Figure~\ref{fig-ders-spat-aff}.%
\footnote{The computations of these expressions, as well as all the other
  generalized Hermite polynomials in this paper, have been performed in
  Wolfram Mathematica.}
By comparing these expressions,
we can then verify the
following infinitesimal relationships with respect to variations in the
scale parameter $s$ and the elements $\Sigma_{11}$,  $\Sigma_{12}$
and  $\Sigma_{22}$ of the spatial covariance matrix $\Sigma$:
\begin{align}
  \begin{split}
    \partial_s
    = \frac{1}{2}
      \nabla_x^T\Sigma \, \nabla_x =
  \end{split}\nonumber\\
  \begin{split}
    = \frac{1}{2}
    \left(
      \Sigma_{11} \, \partial_{x_1 x_1}
      + 2 \Sigma_{12} \, \partial_{x_1 x_2}
      + \Sigma_{22} \, \partial_{x_2 x_2}
    \right),
  \end{split}\\
  \begin{split}
    \partial_{\Sigma_{11}} = \frac{1}{2} \, s \, \partial_{x_1 x_1},
  \end{split}\\
  \begin{split}
    \partial_{\Sigma_{12}} = \frac{1}{2} \, s \, \partial_{x_1 x_2},
  \end{split}\\
  \begin{split}
    \partial_{\Sigma_{22}} = \frac{1}{2} \, s \, \partial_{x_2 x_2}.
  \end{split}    
\end{align}
With regard to the differential relationships over the parameters
$s$, $\Sigma_{11}$ and $\Sigma_{22}$, the corresponding partial
differential equations are parabolic, and can thereby only be evolved in the
corresponding positive directions. Hence, the expressions for
$\partial_s$, $\partial_{\Sigma_{11}}$ and $\partial_{\Sigma_{22}}$
correspond to the notions of infinitesimal generators for
multi-parameter semi-groups.
The differential relationship over the parameter $\Sigma_{12}$ does,
on the other hand, correspond to a hyperbolic partial differential
equation, and could therefore be evolved in both directions.

\subsection{Spatio-temporal receptive fields based on isotropic
  spatial smoothing}
\label{sec-lie-group-spattemp-iso}

Over a joint spatio-temporal image domain, let us first combine an
isotropic Gaussian kernel over the spatial domain, for which the spatial covariance matrix
$\Sigma$ is a unit matrix $\Sigma = I$, with a non-causal temporal Gaussian
kernel of the form
\begin{equation}
  \label{eq-temp-gauss}
  h(t;\; \tau) = \frac{1}{\sqrt{2 \pi \tau}} \, e^{-t^2/2 \tau},
\end{equation}
which for any value of the velocity parameter
$v = (v_1, v_2)^T \in \bbbr^2$
gives the following joint spatio-temporal Gaussian kernel
\begin{align}
  \begin{split}
    T(x, t;\; s, \tau, v) = g(x - v \, t;\; s, I) \, h(t;\; \tau) 
  \end{split}\nonumber\\
  \begin{split}
    \label{eq-isotrop-spattemp-gauss}
    = \frac{1}{2\sqrt{2} \pi^{3/2} \, s \sqrt{\tau}}
    \, e^{-(x-v \, t)^T (x-v \, t)/2 s} \, e^{-t^2/2 \tau}.
  \end{split}
\end{align}
The corresponding spatio-temporal scale-space representation
$L \colon \bbbr^2 \times \bbbr \times \bbbr_+ \times \bbbr_+ \times
\bbbr^2 \rightarrow \bbbr$
of any 2+1-D video sequence $f \colon \bbbr^2 \times \bbbr \rightarrow \bbbr$
is then given by
\begin{equation}
  L(\cdot, \cdot;\; s, \tau, v)
  = T(\cdot, \cdot;\; s, \tau, v) * f(\cdot, \cdot),
\end{equation}
for which it is natural to introduce a velocity-adapted temporal
derivative operator according to
\begin{equation}
  \partial_{\bar t} = \partial_t + v_1 \, \partial_{x_1} + v_2 \, \partial_{x_2}.
\end{equation}
Differentiating the spatio-temporal kernel $T(x, t;\; s, \tau, v)$
according to (\ref{eq-isotrop-spattemp-gauss})
with respect to the spatial image coordinates $x = (x_1, x_2)^T$ and the
time variable $t$
up to order 2, as well as with respect to the spatial
scale parameter $s$, the temporal scale parameter $\tau$ and the
elements $v_1$ and $v_2$ of the velocity vector $v$, leads to the
partial derivatives shown in Figure~\ref{fig-ders-spattemp-iso}
in Appendix~\ref{app-ders-spattemp-iso}.

From these derivative expressions, we can in turn verify the following
infinitesimal relationships over variations of the set of filter parameters
$(s, \tau, v)$ of the receptive fields:
\begin{align}
  \begin{split}
    \partial_s
    = \frac{1}{2} \left( \partial_{xx} + \partial_{yy} \right),
  \end{split}\\
  \begin{split}
    \partial_{\tau}
    = \frac{1}{2} \partial_{\bar{t}\bar{t}}
  \end{split}\nonumber\\
  \begin{split}
    \quad\,\,\,
    = \frac{1}{2}
    \left( \partial_{tt}
      + 2 v_1 \, \partial_{xt} + 2 v_2 \, \partial_{yt}
    \right.
  \end{split}\nonumber\\
  \begin{split}
    \left. \quad\quad\quad\quad
    + v_1^2 \, \partial_{xx} + 2 v_1 \, v_2 \, \partial_{xy} + v_2^2 \, \partial_{yy}
    \right),
  \end{split}\\
  \begin{split}
    \partial_{v_1}  = \tau \, \partial_{x_1 \bar{t}},
  \end{split}\\
  \begin{split}
    \partial_{v_2}  = \tau \, \partial_{x_2 \bar{t}}.
  \end{split}
\end{align}
Here, concerning the differential relationships over the parameters $s$ and
$\tau$, the corresponding partial differential equations are
parabolic, and can therefore only be evolved in the corresponding positive
directions. Hence, the expressions for $\partial_s$ and $\partial_{\tau}$
correspond to the notions of infinitesimal generators for
multi-parameter semi-groups.
The differential relationship over the parameters $v_1$ and $v_2$ do,
on the other hand, correspond to a hyperbolic partial differential
equations, and could therefore be evolved in both directions.

\subsection{Spatio-temporal receptive fields based on affine Gaussian
  smoothing}
\label{sec-lie-group-spattemp-aff}

To combine the variabilities under spatial affine transformations and
Galilean transformations, let us finally consider a joint spatio-temporal
kernel obtained by combining the general affine Gaussian kernel
$g(x;\; s, \Sigma)$ according to (\ref{eq-gauss-fcn-2D})
with the non-causal temporal Gaussian kernel
$h(t;\; \tau) = g_{1D}(t;\; \tau)$ according to
(\ref{eq-temp-gauss}), leading to the joint spatio-temporal kernel
\begin{equation}
  \label{eq-aff-spattemp-gauss}
  T(x, t;\; s, \Sigma, \tau, v) = g(x - v \, t;\; s, \Sigma) \, g_{1D}(t;\; \tau)
\end{equation}
with the corresponding spatio-temporal scale-space representation
$L \colon \bbbr^2 \times \bbbr \times \bbbr_+ \times \bbbs_+^2 \times \bbbr_+ \times
\bbbr^2 \rightarrow \bbbr$ of
any 2+1-D video sequence $f \colon \bbbr^2 \times \bbbr \rightarrow \bbbr$:
according to
\begin{equation}
  L(\cdot, \cdot;\; s, \Sigma, \tau, v)
  = T(\cdot, \cdot;\; s, \Sigma, \tau, v) * f(\cdot, \cdot).
\end{equation}
Differentiating the above spatio-temporal kernel $T(x, t;\; s, \Sigma, \tau, v)$
with respect the spatial image coordinates $x = (x_1, x_2)^T$ and the
time variable $t$
up to order 2, as well as with respect to the spatial
scale parameter $s$, the elements $\Sigma_{11}$, $\Sigma_{12}$ and
$\Sigma_{22}$
of the spatial covariance matrix $\Sigma$, 
the temporal scale parameter $\tau$ and the
elements $v_1$ and $v_2$ of the velocity vector $v$, then gives the
partial derivatives shown in Figure~\ref{fig-ders-spattemp-aff}
in Appendix~\ref{app-ders-spattemp-aff}.

From these derivative expressions, we can in turn verify the following
infinitesimal relationships over the set of the filter parameters
$(s, \Sigma, \tau, v)$ of the receptive fields:
\begin{align}
  \begin{split}
    \partial_s
    & = \frac{1}{2} \nabla_x^T\Sigma \, \nabla_x =
  \end{split}\nonumber\\
  \begin{split}
    & = \frac{1}{2}
    \left(
      \Sigma_{11} \, \partial_{x_1 x_1}
      + 2 \Sigma_{12} \, \partial_{x_1 x_2}
      + \Sigma_{22} \, \partial_{x_2 x_2}
    \right),
  \end{split}\\
  \begin{split}
    \partial_{\Sigma_{11}}
    & = \frac{1}{2} \, s \, \partial_{x_1 x_1},
  \end{split}\\
  \begin{split}
    \partial_{\Sigma_{12}}
    & = \frac{1}{2} \, s \, \partial_{x_1 x_2},
  \end{split}\\
  \begin{split}
    \partial_{\Sigma_{22}}
    & = \frac{1}{2} \, s \, \partial_{x_2 x_2},
  \end{split}
\end{align}
\begin{align}
  \begin{split}
    \partial_{\tau}
    & = \frac{1}{2} \partial_{\bar{t}\bar{t}}
  \end{split}\nonumber\\
  \begin{split}
    & = \frac{1}{2}
    \left( \partial_{tt}
      + 2 v_1 \, \partial_{xt} + 2 v_2 \, \partial_{yt}
    \right.
  \end{split}\nonumber\\
  \begin{split}
    & \phantom{= \frac{1}{2} \left( \right.}
    \left. 
    + v_1^2 \, \partial_{xx} + 2 v_1 \, v_2 \, \partial_{xy} + v_2^2 \, \partial_{yy}
    \right),
  \end{split}\\
  \begin{split}
    \partial_{v_1}
    & = \tau \, \partial_{x_1 \bar{t}},
  \end{split}\\
  \begin{split}
    \partial_{v_2}
    & = \tau \, \partial_{x_2 \bar{t}}.
  \end{split}
\end{align}
For this joint combination of the computational structures in the two
partially simplified receptive field models in
Sections~\ref{sec-lie-group-spat} and~\ref{sec-lie-group-spattemp-iso},
for the differential relationships over the parameters
$s$, $\Sigma_{11}$, $\Sigma_{22}$ and $\tau$, it holds that the corresponding partial
differential equations are parabolic, and can therefore only be evolved in the
corresponding positive directions. Hence, the expressions for 
$\partial_s$, $\partial_{\Sigma_{11}}$, $\partial_{\Sigma_{22}}$ and $\partial_{\tau}$
correspond to the notions of infinitesimal generators for multi-parameter semi-groups.
The differential relationships over the parameters $\Sigma_{12}$,
$v_1$ and $v_2$ do, on the other hand, correspond to hyperbolic partial differential
equations, and could therefore be evolved in both directions.

\section{Macroscopic cascade structures for the generalized
  Gaussian derivative model}
\label{sec-macroscop-rels}

To avoid explicit integration of the previous infinitesimal
relationships between the receptive field responses for different values
of the filter parameters, an alternative way to model such
relationships is in terms of macroscopic relations.
For simplicity, we again start by deriving such relations for the receptive
field models that are purely based on smoothing with Gaussian kernels,
thus initially excluding the non-causal spatio-temporal receptive
field model based on convolution with the time-causal limit kernel
(\ref{eq-time-caus-lim-kern}) over the temporal domain.

\subsection{Cascade smoothing properties based on spatial or
  spatio-temporal covariance matrices}
\label{sec-casc-prop-non-caus-spat-strf-models}

Due to the special properties of the $N$-dimensional Gaussian
kernel, it holds that the convolution of two Gaussian kernels
$g \colon \bbbr^N \times \bbbr^N \times \bbbs_+^N \rightarrow \bbbr$
\begin{equation}
  g(\cdot;\; \mu_1, \Sigma_1) * g(\cdot;\; \mu_2, \Sigma_2)
  = g(\cdot;\; \mu_1 + \mu_2, \Sigma_1 + \Sigma_2)
\end{equation}
with mean vectors $m_1, m_2 \in \bbbr^N$  and spatial covariance
matrices $\Sigma_1, \Sigma_2 \in \bbbs_+^N$ is a Gaussian kernel
with added mean vectors and added covariance matrices.

\subsubsection{Purely spatial receptive fields based on affine Gaussian smoothing}
\label{sec-macro-spat}

Applied to the purely spatial smoothing operation, corresponding to
convolution with the affine Gaussian kernel, it therefore follows that the
scale-space representation $L(\cdot;\; s_2, \Sigma_2)$
at a coarser spatial scale $(s_2, \Sigma_2)$ can be computed from
the scale-space representation $L(\cdot;\; s_1, \Sigma_1)$ at any
finer spatial scale $(s_1, \Sigma_1)$ according to
\begin{equation}
  \label{eq-casc-prop-spat}
  L(\cdot;\; s_2, \Sigma_2)
  = g(\cdot;\; \Delta s, \Delta \Sigma) * L(\cdot;\; s_1, \Sigma_1).
\end{equation}
This result holds provided the filter parameters $(\Delta s, \Delta \Sigma)$ of
the incremental smoothing kernel $g(\cdot;\; \Delta s, \Delta \Sigma)$
satisfy
\begin{equation}
  \label{eq-casc-rel-pure-spat-rf}
  s_2 \, \Sigma_2 = \Delta s \, \Delta \Sigma + s_1 \, \Sigma_1,
\end{equation}
that is, provided that the incremental effective spatial covariance matrix
\begin{equation}
  \Delta s \, \Delta \Sigma = s_2 \, \Sigma_2 - s_1 \, \Sigma_1
\end{equation}
is a positive symmetric definite matrix. In other words,
the resulting cascade smoothing property according to
(\ref{eq-casc-prop-spat}) can be computed in the forward direction of
a cone in the parameter space of the purely spatial kernels.

Due to the linearity of the corresponding operators, this cascade
smoothing property extends to the receptive field responses defined by
applying spatial derivative operators ${\cal D}_x$ to the
spatial scale-space representation of the form
\begin{equation}
  {\cal R} \, f(\cdot)
  = {\cal D}_x L(\cdot;\; s, \Sigma),
\end{equation}
thus implying an explicit cascade smoothing property
for spatial-derivative-based receptive fields of the form
\begin{equation}
  {\cal D}_x L(\cdot;\; s_2, \Sigma_2) 
  = T(\cdot;\; \Delta s, \Delta \Sigma) * {\cal D}_x L(\cdot;\; s_1, \Sigma_1).
\end{equation}

\subsubsection{Joint spatio-temporal receptive fields based on 
  affine Gaussian smoothing}
\label{sec-macro-spattemp-aff}

To express a corresponding cascade smoothing property over the joint
spatio-temporal domain, let us make use of the fact that the composed effect
of the smoothing operation, with a 2-D either isotropic Gaussian kernel or an
anisotropic affine Gaussian kernel over the spatial domain in
combination with smoothing with a 1-D non-causal temporal Gaussian kernel
over the temporal domain, can equivalently be obtained by smoothing
with a joint spatio-temporal Gaussian kernel over the 2+1-D joint
spatio-temporal domain. Thus, when using an anisotropic affine
spatial Gaussian kernel, leading to a joint spatio-temporal kernel
of the form (\ref{eq-aff-spattemp-gauss})
\begin{align}
  \begin{split}
    T(x, t;\; s, \Sigma, \tau, v) = g(x - v \, t;\; s, \Sigma) \, h(t;\; \tau) =
  \end{split}\nonumber\\
  \begin{split}
    \label{eq-aff-spattemp-gauss-again}
    = \frac{1}{2\sqrt{2} \pi^{3/2} \, s \sqrt{\tau}}
    \, e^{-(x-v \, t)^T \Sigma^{-1} (x-v \, t)/2 s} \, e^{-t^2/2 \tau},
  \end{split}
\end{align}
this operation can equivalently be described as a 2+1-D convolution with a joint
spatio-temporal kernel
$g_{2+1\text{-D}} \colon \bbbr^3 \times \bbbr^3 \times \bbbs_+^3 \rightarrow \bbbr$
of the form
\begin{multline}
  \label{eq-joint-spattemp-gauss-2p1d}
  g_{2+1\text{-D}}(p;\, m_{2+1\text{-D}}, \Sigma_{2+1\text{-D}}) = \\
  = \frac{1}{(2 \pi)^{3/2} \sqrt{\det{\Sigma_{2+1\text{-D}}}}} \,
    e^{-(p - m_{2+1\text{-D}})^T \Sigma_{2+1\text{-D}}^{-1} (p - m_{2+1\text{-D}})/2}
\end{multline}
over the spatio-temporal image coordinates
$p = (x_1, x_2, t)^T$.

To determine the filter parameters $m_{2+1\text{-D}}$ and $\Sigma_{2+1\text{-D}}$
in the joint spatio-temporal smoothing kernel
(\ref{eq-joint-spattemp-gauss-2p1d}), let us with the reformulation
of the spatio-temporal smoothing kernel
(\ref{eq-isotrop-spattemp-gauss}) as
\begin{equation}
  T(p;\; s, \Sigma, \tau, v)  = T(x, t;\; s, \Sigma, \tau, v) 
\end{equation}
calculate the joint mean vector $m_{2+1\text{-D}}$ and the
joint covariance matrix $\Sigma_{2+1\text{-D}}$ according to
\begin{align}
  \begin{split}
    m_{2+1\text{-D}}
    = \frac{\int_{p \in \bbbr^{2+1}} p \, T(p;\; s, \Sigma, \tau, v) \, dp}
    {\int_{p \in \bbbr^{2+1}} T(p;\; s, \Sigma, \tau, v) \, dp},
  \end{split}\\
  \begin{split}
    \Sigma_{2+1\text{-D}} 
    = \frac{\int_{p \in \bbbr^{2+1}} p \, p^T \, T(p;\; s, \Sigma, \tau, v) \, \, dp}
    {\int_{p \in \bbbr^{2+1}} T(p;\; s, \Sigma, \tau, v) \, dp}
    - m_{2+1\text{-D}} \, m_{2+1\text{-D}}^T,
  \end{split}
\end{align}
which gives $m_{2+1\text{-D}} = (0, 0, 0)^T$ and
\begin{equation}
  \label{eq-spattemp-cov-mat-noncaus-spattemp-scsp}
  \Sigma_{2+1\text{-D}} =
  \left(
    \begin{array}{ccc}
      \Sigma_{11} \, s + \tau \, v_1^2 & \Sigma_{12} \, s + \tau \, v_1 \, v_2 & \, \tau \, v_1 \\
      \Sigma_{12} \, s + \tau \, v_1 \, v_2 & \Sigma_{22} \, s + \tau \, v_2^2 & \, \tau \, v_2 \\
      \tau \, v_1 & \tau \, v_2 & \, \tau  \\
    \end{array}
  \right).
\end{equation}
Notably, concerning the structure of the expression for the joint
spatio-temporal covariance matrix, a general $3 \times 3$
spatio-temporal covariance matrix has 6 degrees of freedom,
while this expression also has 6 effective degrees of freedom in terms
of the filter parameters $s$, $\Sigma_{11}$, $\Sigma_{12}$,
$\Sigma_{22}$, $\tau$, $v_1$ and $v_2$, if we disregard the
overparameterization induced by the combination of the spatial scale
parameter $s$ with the elements $\Sigma_{11}$, $\Sigma_{12}$,
$\Sigma_{22}$ of the spatial covariance matrix $\Sigma$.

Thus, let us assume that we have two configurations of parameter settings,
that would correspond to a cascade smoothing property over the filter
parameters for two joint spatio-temporal smoothing kernels of the
form (\ref{eq-aff-spattemp-gauss-again}), with
\begin{itemize}
\item
  the joint
  convolution kernel $T(\cdot, \cdot;\; s_i, \Sigma_i, \tau_i, v_i)$ in the lower
  layer with the velocity vector $v_i = (v_{i,1}, v_{i,2})^T$ and the elements of the
  spatial covariance matrix $\Sigma_i$ being $\Sigma_{i,11}$,
  $\Sigma_{i,12}$ and $\Sigma_{i,22}$,
  and with
\item
  the joint convolution kernel $T(\cdot, \cdot;\; s_j, \Sigma_j, \tau_j, v_j)$ in the
  higher layer with the velocity vector $v_j = (v_{j,1}, v_{j,2})^T$ and the elements of the
  spatial covariance matrix $\Sigma_j$ being $\Sigma_{j,11}$,
  $\Sigma_{j,12}$ and $\Sigma_{j,22}$.
\end{itemize}
Then, from the additive properties of the joint
spatio-temporal covariance matrices over 2+1-D space-time,
a cascade smoothing property would be possible, if there would exist

\begin{itemize}
\item
  an incremental joint spatio-temporal convolution kernel
  $T(\cdot, \cdot;\; \Delta s, \Delta \Sigma, \Delta \tau, \Delta v)$
   with the velocity vector $\Delta v = (\Delta v_1, \Delta v_2)^T$ and the elements of the
   spatial covariance matrix $\Delta \Sigma$ being
   $\Delta \Sigma_{11}$, $\Delta \Sigma_{12}$ and $\Delta \Sigma_{22}$,
\end{itemize}
such that
\begin{multline}
  \label{eq-casc-smooth-spat-temp-non-caus}
  T(\cdot, \cdot;\; s_j, \Sigma_j, \tau_j, v_j) = \\
  = T(\cdot, \cdot;\; \Delta s, \Delta \Sigma, \Delta \tau, \Delta v)
  * T(\cdot, \cdot;\; s_i, \Sigma_i, \tau_i, v_i),
\end{multline}
thus implying an explicit cascade smoothing property of the form
\begin{multline}
  \label{eq-casc-smooth-spat-temp-non-caus-scsp}
  L(\cdot, \cdot;\; s_j, \Sigma_j, \tau_j, v_j) = \\
  = T(\cdot, \cdot;\; \Delta s, \Delta \Sigma, \Delta \tau, \Delta v)
  * L(\cdot, \cdot;\; s_i, \Sigma_i, \tau_i, v_i)
\end{multline}
between the non-causal spatio-temporal scale-space representations
$L(\cdot, \cdot;\; s_i, \Sigma_i, \tau_i, v_i)$ and
$L(\cdot, \cdot;\; s_j, \Sigma_j, \tau_j, v_j)$.

Thereby, the following relations between the filter parameters
would have to be satisfied:
\begin{align}
  \begin{split}
    \Sigma_{j,11} \, s_j+\tau_j \, v_{j,1}^2 =
  \end{split}\nonumber\\
  \begin{split}
    \label{eq-noncaus-casc-crit-sigma11}
    \quad\quad
    = \Delta \Sigma_{11} \, \Delta s + \Delta \tau  \, \Delta v_1^2
        + \Sigma_{i,11} \, s_i+ \tau_i  \, v_{i,1}^2,
  \end{split}\\
  \begin{split}
    \Sigma_{j,12} \, s_j +\tau_j  \, v_{j,1} \, v_{j,2} =
  \end{split}\nonumber\\
  \begin{split}
    \label{eq-noncaus-casc-crit-sigma12}
    \quad\quad     
    = \Delta \Sigma_{12} \, \Delta s+ \Delta \tau  \, \Delta v_1 \, \Delta v_2
    + \Sigma_{i,12} \, s_i + \tau_i  \, v_{i,1} \, v_{i,2},
  \end{split}\\
  \begin{split}
    \Sigma_{j,22} \, s_j +\tau_j \, v_{j,2}^2 =
  \end{split}\nonumber\\
  \begin{split}
    \label{eq-noncaus-casc-crit-sigma22}
    \quad\quad    
    = \Delta \Sigma_{22} \, \Delta s + \Delta \tau  \, \Delta v_2^2
    + \Sigma_{i,22} \, s_i + \tau_i \,  v_{i,2}^2,
  \end{split}\\
  \begin{split}
        \label{eq-noncaus-casc-crit-v1}
    \tau_j \,  v_{j,1} = \Delta \tau \, \Delta v_1 + \tau_i \,  v_{i,1},
  \end{split}\\
  \begin{split}
        \label{eq-noncaus-casc-crit-v2}    
    \tau_j \, v_{j,2} = \Delta \, \tau \, \Delta \, v_2  + \tau_i \, v_{i,2},
  \end{split}\\
  \begin{split}
        \label{eq-noncaus-casc-crit-tau}    
    \tau_j  = \Delta \tau  + \tau_i.
  \end{split}
\end{align}
Solving for the parameters $\Delta s$, $\Delta \Sigma_{11}$, $\Delta \Sigma_{12}$,
$\Delta \Sigma_{22}$, $\Delta \tau$, $\Delta v_1$ and $\Delta v_2$ of the
incremental spatio-temporal convolution kernel
$T(\cdot, \cdot;\; \Delta s, \Delta \Sigma, \Delta \tau, \Delta v)$,
then gives
\begin{align}
  \begin{split}
    \Delta \tau = \tau_j - \tau_i,
  \end{split}\\
  \begin{split}
    \Delta v_1 = \frac{\tau_j \, v_{j,1} - \tau_i \, v_{i,1}}{\tau_j - \tau_i},
  \end{split}\\
  \begin{split}
    \Delta v_2 = \frac{\tau_j \, v_{j,2} - \tau_i \, v_{i,2}}{\tau_j - \tau_i},
  \end{split}\\
  \begin{split}
    \Delta s \, \Delta \Sigma_{11}
    = \Sigma_{j,11} \, s_j - \Sigma_{i,11} \, s_i 
    + \frac{\tau_i \, \tau_j \, (v_{i,1}-v_{j,1})^2}{\tau_i-\tau_j},
  \end{split}\\
  \begin{split}
    \Delta s \, \Delta \Sigma_{12}
    = \Sigma_{j,12} \, s_j - \Sigma_{i,12} \, s_i +
  \end{split}\\
  \begin{split}
    \quad\quad \quad\quad\quad\,
    + \frac{\tau_i \, \tau_j  \, (v_{i,1}-v_{j,1}) (v_{i,2}-v_{j,2})}{\tau_i-\tau_j},
  \end{split}\\
  \begin{split}
    \Delta s \, \Delta \Sigma_{22} 
    = \Sigma_{j,22} \, s_j - \Sigma_{i,22} \, s_i
    + \frac{\tau_i \, \tau_j \, (v_{i,2}-v_{j,2})^2}{\tau_i-\tau_j}.
  \end{split}
\end{align}
Given that the resulting parameters $\Delta s$, $\Delta \Sigma_{11}$, $\Delta \Sigma_{12}$,
$\Delta \Sigma_{22}$, $\Delta \tau$, $\Delta v_1$ and $\Delta v_2$ for the
incremental spatio-temporal convolution kernel should obey the consistency
requirements
\begin{align}
  \begin{split}
    \Delta \Sigma_{11} \, \Delta s \geq 0,
  \end{split}\\
  \begin{split}
    \Delta \Sigma_{22} \, \Delta s \geq 0,
  \end{split}\\
  \begin{split}
    (\Delta \Sigma_{11} \, \Delta s) (\Delta \Sigma_{22} \, \Delta s)
    - (\Delta \Sigma_{12} \, \Delta s)^2 \geq 0,
  \end{split}\\
  \begin{split}
    \Delta \tau \geq 0,
  \end{split}
\end{align}
it thereby follows that it is possible to
implement a cascade smoothing property
\begin{multline}
  L(\cdot, \cdot;\; s_j, \Sigma_j, \tau_j, v_j) = \\
  = T(\cdot, \cdot;\; \Delta s, \Delta \Sigma, \Delta \tau, \Delta v)
  * L(\cdot, \cdot;\; s_i, \Sigma_i, \tau_i, v_i)
\end{multline}
for the joint spatio-temporal scale-space representations of video
sequences $f$ according to
\begin{equation}
  L(\cdot, \cdot;\; s, \Sigma, \tau, v) 
  = T(\cdot, \cdot;\; s, \Sigma, \tau, v) * f(\cdot, \cdot).
\end{equation}
Due to the linearity of the corresponding operators, this cascade
smoothing property extends to receptive field responses defined by
applying spatio-temporal derivative operators ${\cal D}_{x,t}$ to the
spatio-temporal scale-space representation of the form
\begin{equation}
  {\cal R} \, f(\cdot, \cdot)
  = {\cal D}_{x,t} L(\cdot, \cdot;\; s, \Sigma, \tau, v),
\end{equation}
thus implying an explicit cascade smoothing property of the
spatio-temporal derivative-based spatio-temporal receptive fields of
the form
\begin{multline}
  {\cal D}_{x,t} L(\cdot, \cdot;\; s_j, \Sigma_j, \tau_j, v_j) = \\
  = T(\cdot, \cdot;\; \Delta s, \Delta \Sigma, \Delta \tau, \Delta v)
  * {\cal D}_{x,t} L(\cdot, \cdot;\; s_i, \Sigma_i, \tau_i, v_i).
\end{multline}

\subsubsection{Joint spatio-temporal receptive fields based on 
  isotropic Gaussian smoothing}
\label{sec-macro-spattemp-iso}

In relation to the results from the previous section,
it is worth observing that if the
spatial smoothing operation would instead only be performed based on an isotropic
Gaussian kernel, with the spatial covariance matrix $\Sigma$ equal to
the unit matrix $\Sigma = I$, then the parameters $(s, \tau, v)$ in
the spatio-temporal receptive fields would only span 4 out of the
general 6 degrees of freedom in the joint spatio-temporal covariance
matrix $\Sigma_{2+1\text{-D}}$ according to
(\ref{eq-spattemp-cov-mat-noncaus-spattemp-scsp}).
Thereby, the configurations for which it would be possible to
implement a corresponding cascade smoothing property
\begin{multline}
  L(\cdot, \cdot;\; s_j, \tau_j, v_j) = \\
  = T(\cdot, \cdot;\; \Delta s, \Delta \tau, \Delta v)
  * L(\cdot, \cdot;\; s_i, \tau_i, v_i)
\end{multline}
for the joint spatio-temporal scale-space representations of video
sequences $f$ according to
\begin{equation}
  L(\cdot, \cdot;\; s, \tau, v) 
  = T(\cdot, \cdot;\; s, \tau, v) * f(\cdot, \cdot)
\end{equation}
would on the other hand be highly degenerate.
Thus, beyond the previously mentioned closedness property under
anisotropic spatial affine transformations, the use of an affine
Gaussian kernel for spatial smoothing also offers clear advantages in
terms of the ability for an idealized vision system, to implement
spatio-temporal cascade smoothing properties for the receptive fields,
in order to decrease the amount of computational work in filter bank
implementations of joint spatio-temporal receptive fields.

\subsubsection{The special case with equal image velocities}
\label{sec-eq-velocities-non-caus-strfs}

Concerning the relationships derived in
Section~\ref{sec-macro-spattemp-aff},
it is of particular interest to consider
the special case when all the image velocities are equal,
{\em i.e.\/},
\begin{equation}
   v_j = \Delta v = v_i.
\end{equation}
Then, Equations~(\ref{eq-noncaus-casc-crit-v1}) and
(\ref{eq-noncaus-casc-crit-v2}) both reduce to the
criterion (\ref{eq-noncaus-casc-crit-tau})
\begin{equation}
  \label{eq-noncaus-casc-crit-tau-again}
  \tau_j = \Delta \tau + \tau_i.
\end{equation}
Furthermore, given the assumption that this condition
is satisfied, the relationships
(\ref{eq-noncaus-casc-crit-sigma11})--(\ref{eq-noncaus-casc-crit-sigma22})
then reduce to the following relationships
\begin{align}
  \begin{split}
    \label{eq-noncaus-casc-crit-sigma11-coupled-v}
    \Sigma_{j,11} \, s_j 
    = \Delta \Sigma_{11} \, \Delta s 
        + \Sigma_{i,11} \, s_i,
  \end{split}\\
  \begin{split}
    \label{eq-noncaus-casc-crit-sigma12-coupled-v}
    \Sigma_{j,12} \, s_j 
    = \Delta \Sigma_{12} \, \Delta s
    + \Sigma_{i,12} \, s_i,
  \end{split}\\
  \begin{split}
    \label{eq-noncaus-casc-crit-sigma22-coupled-v}
    \Sigma_{j,22} \, s_j 
    = \Delta \Sigma_{22} \, \Delta s 
    + \Sigma_{i,22} \, s_i,
  \end{split}
\end{align}
corresponding to similar relationships as for the elements of
the purely spatial receptive field model in
Equation~(\ref{eq-casc-rel-pure-spat-rf}).

An explanation for this simplicity of the resulting relationships is because
of the Galilean covariance property of the spatio-temporal receptive field
model, which then up to a Galilean transformation is isomorphic to a
purely space-time separable model for the spatio-temporal receptive
fields based on spatio-temporal smoothing operations of the form
\begin{align}
  \begin{split}
    T(x, t;\; s, \Sigma, \tau) = g(x;\; s, \Sigma) \, h(t;\; \tau) =
  \end{split}\nonumber\\
  \begin{split}
    \label{eq-aff-spattemp-gauss-again2}
    = \frac{1}{2\sqrt{2} \pi^{3/2} \, s \sqrt{\tau}}
    \, e^{-x^T \Sigma^{-1} x/2 s} \, e^{-t^2/2 \tau}.
  \end{split}
\end{align}
With regard to the special case with joint
spatio-temporal receptive fields based on isotropic
Gaussian smoothing with $\Sigma = I$
considered in Section~\ref{sec-macro-spattemp-iso},
it is, however, interesting to note that those degeneracies do not in
any negative way affect the possibility for the receptive fields to
span that subgroup of receptive fields covariant to spatial and
temporal scaling transformations and Galilean transformations.

\section{Relationships between receptive field responses for different
  parameter settings for spatio-temporal receptive fields based on
  temporal smoothing with the time-causal limit kernel}
\label{sec-inf-macro-rels-time-caus-spat-temp}

In the treatment so far, we have made use of special algebraic
properties, as arising from using spatial and spatio-temporal smoothing
kernels solely based on Gaussian kernels, for deriving both
infinitesimal and macroscopic relationships between the receptive field
responses computed for different values of the filter parameters.

When replacing the non-causal temporal Gaussian kernel
$h(t;\; \tau) = g(t;\; \tau)$ according to
(\ref{eq-non-caus-temp-gauss}) by the time-causal limit kernel
$h(t;\; \tau) = \psi(t;\; \tau, c)$  according to
(\ref{eq-time-caus-lim-kern}),
where $c >1$ is a distribution parameter that specifies the ratio
between the temporal scale levels in units of the temporal standard
deviation of the kernel,
we obtain a joint spatio-temporal smoothing kernel of the form
\begin{align}
  \begin{split}
    T(x, t;\; s, \Sigma, \tau, v, c) = g(x - v \, t;\; s, \Sigma) \, \psi(t;\; \tau, c)
  \end{split}\nonumber\\
  \begin{split}
    \label{eq-aff-spattemp-kern-timecaus}
    = \frac{1}{2 \pi \, s}
    \, e^{-(x-v \, t)^T \Sigma^{-1} (x-v \, t)/2 s} \, \psi(t;\; \tau, c).
  \end{split}
\end{align}
This kernel cannot, however, be differentiated with respect to the here
fully discrete temporal scale parameter
$\tau_i = \tau_0 \, c^{2i}$. Additionally, expressing the derivative of the
time-causal limit kernel with respect to the
temporal variable $t$ is neither straightforward, because of the relatively
complex explicit expression for the time-causal limit kernel over the
temporal domain:
\begin{equation}
  \Psi(\cdot;\; \tau, c) = *_{k = 1}^{\infty} h_{\exp}(\cdot;\; \mu_k),
\end{equation}
where
\begin{equation}
  \label{eq-trunc-exp-kernel}
    h_{\exp}(t;\; \mu_k) 
    = \left\{
        \begin{array}{ll}
          \frac{1}{\mu_k} \, e^{-t/\mu_k} & t \geq 0, \\
          0         & t < 0,
        \end{array}
      \right.
\end{equation}
and the time constants $\mu_k$ of the individual truncated
exponential kernels are given by
\begin{equation}
  \mu_k = c^{-k} \sqrt{c^2-1} \sqrt{\tau}.
\end{equation}
Specifically, we do not have any straightforward
correspondence to generalized Hermite polynomials based on temporal
derivatives of the time-causal limit kernel.

Furthermore, in terms of cascade smoothing properties, the time-causal
limit kernel does not form a semi-group as the 1-D temporal
Gaussian kernel $h(t;\; \tau) = g(t;\; \tau)$ does.
Instead, the time-causal limit obeys the following cascade
smoothing property between adjacent levels of temporal scales
(see Lindeberg (\citeyear{Lin23-BICY}) Equation~(28)):
\begin{equation}
  \label{eq-recur-rel-limit-kernel}
  \Psi(\cdot;\; \tau, c)
  = h_{\exp}(\cdot;\; \tfrac{\sqrt{c^2-1}}{c} \sqrt{\tau}) * \Psi(\cdot;\; \tfrac{\tau}{c^2}, c).
\end{equation}
From these structural differences, arising from using the time-causal
limit kernel as the temporal smoothing kernel instead of the
non-causal temporal Gaussian kernel, we cannot expect to be able to directly apply similar
methodologies for deriving infinitesimal or macroscopic evolution
properties over the filter parameters for the time-causal
spatio-temporal kernel (\ref{eq-aff-spattemp-kern-timecaus}),
as we used for deriving such relationships for the fully
Gaussian-smoothing based receptive field models
in Sections~\ref{sec-inf-rels-non-caus-rfs}
and~\ref{sec-macroscop-rels}.

As will be described further in
Section~\ref{sec-inf-rels-time-caus-spat-temp}
below, the infinitesimal relationships do, however, carry over
regarding the purely spatial image coordinates $x$, the spatial scale
parameter $s$, as well as for the elements $\Sigma_{11}$,
$\Sigma_{12}$ and $\Sigma_{22}$ of the
spatial covariance matrix $\Sigma$.
Furthermore, as will be described in the following
Section~\ref{sec-macro-rels-time-caus-spat-temp},
in the special case when the image velocities are all equal in the
cascade model for the receptive field responses, the macroscopic
relationships for the spatial components in the spatio-temporal
receptive field model do also carry over from the purely spatial case.

\subsection{Infinitesimal relationships between receptive field
  responses computed for different values of the filter parameters}
\label{sec-inf-rels-time-caus-spat-temp}

Regarding infinitesimal relationships, the differential relationships,
that we derived in Section~\ref{sec-lie-group-spattemp-aff} regarding
derivative operators solely expressed over the spatial domain, will
also hold if we replace the non-causal Gaussian kernel
$h(t;\; \tau) = g_{1D}(t;\; \tau)$ in (\ref{eq-aff-spattemp-gauss})
by the time-causal limit kernel $h(t;\; \tau) = \psi(t;\; \tau, c)$.
Thus, if we for any spatio-temporal video stream
$f \colon \bbbr^2 \times \bbbr \rightarrow \bbbr$
define a time-causal spatio-temporal scale-space
representation according to 
\begin{equation}
  L(\cdot, \cdot;\; s, \Sigma, \tau, v, c)
  = T(\cdot, \cdot;\; s, \Sigma, \tau, v, c) * f(\cdot, \cdot)
\end{equation}
with the joint time-causal spatio-temporal smoothing kernel
$T(\cdot;\; s, \Sigma, \tau, v, c)$ according
to (\ref{eq-aff-spattemp-kern-timecaus}),
then it follows that the following infinitesimal relationships will
hold over the purely spatial filter parameters $(s, \Sigma)$:
\begin{align}
  \begin{split}
    \partial_s
    & = \frac{1}{2} \nabla_x^T\Sigma \, \nabla_x =
  \end{split}\nonumber\\
  \begin{split}
    \label{eq-inf-gen-s-time-caus-spat-temp}
    & = \frac{1}{2}
    \left(
      \Sigma_{11} \, \partial_{x_1 x_1}
      + 2 \Sigma_{12} \, \partial_{x_1 x_2}
      + \Sigma_{22} \, \partial_{x_2 x_2}
    \right),
  \end{split}\\
  \begin{split}
    \label{eq-inf-gen-Sigma11-time-caus-spat-temp}        
    \partial_{\Sigma_{11}}
    & = \frac{1}{2} \, s \, \partial_{x_1 x_1},
  \end{split}\\
  \begin{split}
    \label{eq-inf-gen-Sigma12-time-caus-spat-temp}        
    \partial_{\Sigma_{12}}
    & = \frac{1}{2} \, s \, \partial_{x_1 x_2},
  \end{split}\\
  \begin{split}
    \label{eq-inf-gen-Sigma22-time-caus-spat-temp}    
    \partial_{\Sigma_{22}}
    & = \frac{1}{2} \, s \, \partial_{x_2 x_2}.
  \end{split}
\end{align}
This can be easily understood by observing that when calculating the
ratios between derivatives with respect to either the spatial
coordinates $x$ or the spatially based filter parameters $(s, \Sigma)$,
the occurence of the time-causal limit kernel will cancel in the
resulting expressions. Thus, we obtain similar results for these
ratios computed from the time-causal spatio-temporal smoothing kernel
as shown in Figure~\ref{fig-ders-spattemp-aff-timecaus}
in Appendix~\ref{app-ders-spattemp-aff-timecaus}, as we
previously obtained for the corresponding non-causal spatio-temporal
smoothing kernel shown in Figure~\ref{fig-ders-spattemp-aff}
in Appendix~\ref{app-ders-spattemp-aff}.

By comparing those generalized Hermite polynomials, we can then
establish that the infinitesimal relationships
(\ref{eq-inf-gen-s-time-caus-spat-temp})--(\ref{eq-inf-gen-Sigma22-time-caus-spat-temp})
are valid.

Notably, without differentiating the time-causal limit kernel, we can also
compute the generalized Hermite polynomials corresponding to
differentiation with respect to the elements $v_1$ and $v_2$ of the
velocity vector $v$. So far, we have, however, not yet been able to
match these partial derivatives to corresponding expressions in terms
of derivatives with respect to the spatio-temporal image coordinates
$(x_1, x_2, t)$.

\subsection{Macroscopic cascade structures between receptive field
  responses computed for different values of the filter parameters}
\label{sec-macro-rels-time-caus-spat-temp}

Based on the cascade smoothing property (\ref{eq-recur-rel-limit-kernel})
of the time-causal limit kernel (\ref{eq-time-caus-lim-kern}),
let us for spatio-temporal
smoothing kernels $T(x, t;\; s, \Sigma, \tau, v, c)$ of the form
(\ref{eq-aff-spattemp-kern-timecaus}) consider formulating a cascade
smoothing property for the time-causal spatio-temporal kernels kernels of the form
\begin{multline}
    \label{eq-casc-smooth-spat-temp-time-caus}
  L(\cdot, \cdot;\; s_j, \Sigma_j, \tau_j, v_j, c) = \\
  = \Delta T(\cdot, \cdot;\; \Delta s, \Delta \Sigma, \Delta \tau, \Delta v, c)
  * L(\cdot, \cdot;\; s_i, \Sigma_i, \tau_i, v_i, c),
\end{multline}
where the incremental kernel
$\Delta T(\cdot, \cdot;\; \Delta s, \Delta \Sigma, \Delta \tau, \Delta v, c)$
is given by
\begin{align}
  \begin{split}
    \Delta T(x, t;\; \Delta s, \Sigma, \Delta \tau, \Delta v, c) =
  \end{split}\nonumber\\
  \begin{split}
   = g(x - \Delta v \, t;\; \Delta s, \Delta \Sigma) \,
    h_{\exp}(t;\; \sqrt{\Delta \tau})
  \end{split}\nonumber\\
  \begin{split}
    = \frac{1}{2 \pi \, \Delta s}
    \, e^{-(x-\Delta v \, t)^T \Delta \Sigma^{-1} (x-\Delta v \, t)/2 \Delta s} \times
  \end{split}\nonumber\\
  \begin{split}
    \label{eq-aff-spattemp-kern-timecaus-incr}
    \hphantom{=} \,\,
    h_{\exp}(t;\; \sqrt{\Delta \tau})
  \end{split}
\end{align}
here, for simplicity, restricted to adjacent temporal scales
\begin{equation}
  \label{eq-timecaus-casc-crit-tau}
  \tau_i = \frac{\tau_j}{c^2}
\end{equation}
with
\begin{equation}
  \Delta \tau = \tau_j - \tau_i = \frac{c^2-1}{c^2} \, \tau_j.
\end{equation}
The motivation for the last relationships is that the temporal scale
levels in these types of temporal and spatio-temporal scale-space
representations are inherently discrete according to a geometric
distribution of the form (\ref{eq-disc-temp-sc-limit-kern}).

\subsubsection{The special case with equal image velocities}
\label{sec-eq-velocities-time-caus-strfs}

Unfortunately, it appears as a rather complex problem to derive
closed-form expressions for the relationships between the
parameters  $(\Delta s, \Delta \Sigma, \Delta \tau, \Delta v)$
of the incremental kernel from the parameters 
$(s_i, \Sigma_i, \tau_i, v_i)$ and $(s_j, \Sigma_j, \tau_j, v_j)$
of two involved layers of receptive field responses in the general
case, when the image velocities $\Delta v$, $v_i$ and $v_j$ may be
different. A main reason to technical complications is that the
time-causal limit kernel does not have any compact explicit
expression, since it is given as the convolution of an infinite number
of truncated exponential kernels.

In the special case when the image velocities are all equal
\begin{equation}
   v_j = \Delta v = v_i,
\end{equation}
we can, however, in analogy with the previous treatment in
Section~\ref{sec-eq-velocities-non-caus-strfs}, make use of the
Galilean covariance property to infer that, beyond the relationship
between adjacent scale levels according to
(\ref{eq-timecaus-casc-crit-tau})
\begin{equation}
  \label{eq-timecaus-casc-crit-tau-again}
  \tau_i = \frac{\tau_j}{c^2},
\end{equation}
the purely spatial parameters of the receptive fields should obey
similar relationships as in
Equations~(\ref{eq-noncaus-casc-crit-sigma11-coupled-v})--(\ref{eq-noncaus-casc-crit-sigma22-coupled-v}):
\begin{align}
  \begin{split}
    \label{eq-timecaus-casc-crit-sigma11-coupled-v}
    \Sigma_{j,11} \, s_j 
    = \Delta \Sigma_{11} \, \Delta s 
        + \Sigma_{i,11} \, s_i,
  \end{split}\\
  \begin{split}
    \label{eq-timecaus-casc-crit-sigma12-coupled-v}
    \Sigma_{j,12} \, s_j 
    = \Delta \Sigma_{12} \, \Delta s
    + \Sigma_{i,12} \, s_i,
  \end{split}\\
  \begin{split}
    \label{eq-timecaus-casc-crit-sigma22-coupled-v}
    \Sigma_{j,22} \, s_j 
    = \Delta \Sigma_{22} \, \Delta s 
    + \Sigma_{i,22} \, s_i.
  \end{split}
\end{align}
In this way, we can hence determine the parameters of the incremental
spatio-temporal kernel according to
(\ref{eq-aff-spattemp-kern-timecaus-incr}) in the special case of
equal velocity parameters $v_i$, $\Delta v$ and $v_j$.

We leave it as an open problem for future work to analyze the
potential applicability of the incremental kernel of the form
(\ref{eq-aff-spattemp-kern-timecaus-incr}) to formulate
time-causal cascade model between the scale-space representations
$L(\cdot, \cdot;\; s_i, \Sigma_i, \tau_i, v_i, c)$ and
$L(\cdot, \cdot;\; s_j, \Sigma_j, \tau_j, v_j, c)$ at adjacent
scales $\tau_i$ and $\tau_j$ of the form
(\ref{eq-casc-smooth-spat-temp-time-caus})
in the case of non-equal
velocity parameters $v_i$, $\Delta v$ and $v_j$.

\section{Summary and discussion}
\label{sec-summ-disc}

We have presented an in-depth theoretical analysis about how receptive
field responses computed using linear receptive fields corresponding
to simple cells in the primary visual cortex, as expressed in terms of
the generalized Gaussian derivative model for visual receptive fields,
can be related between different values of the filter parameters.

After an overview of the generalized Gaussian derivative model for visual
receptive fields in Section~\ref{sec-gen-gauss-der-model},
with emphasis on its relation to handle variabilities
in image structures, as generated by natural geometric image
transformations, and as modelled by local linearizations in terms of
(i)~uniform spatial scaling transformations,
(ii)~spatial affine transformations,
(iii)~Galilean transformations and
(iv)~temporal scaling transformations, we have specifically, from the
requirement of covariance properties of the receptive fields under
such geometric image transformations, motivated the importance
of multi-parameter families of visual receptive fields.
This motivation originates from the desirable requirement of
making it possible to, up to first order of approximation, match the receptive
field responses that have been computed from image data acquired under
different viewing conditions. The requirement of closedness of the
representation of receptive field responses under such geometric image
transformations then means that the shapes of the receptive fields
ought to be expanded over the degrees of freedom of the corresponding
geometric image transformations.

For our considered axiomatically
determined normative theory of visual receptive fields, this leads
to spatial and spatio-temporal receptive
fields, that are based on applying spatial and/or spatio-temporal
derivative operators to a multi-parameter spatial and/or
spatio-temporal scale-space representation of the either purely
spatial or joint spatio-temporal image data.
Specifically, we have, by the covariance properties of the receptive
fields in Equations~(\ref{eq-s-transf-result})--(\ref{eq-Sigma-transf-result})
and~(\ref{eq-tau-transf-result})--(\ref{eq-v-transf-result})
related the variabilities in the shapes of the receptive
fields to the parameters of the general class of composed geometric
image transformations of the form (\ref{eq-x-transf}) and
(\ref{eq-t-transf}), which, as illustrated in
Figure~\ref{fig-singlegeom}, represents the first-order variability in
image structures, when observing the surfaces of smooth objects in
dynamic scenes.

Then, we have in Sections~\ref{sec-inf-rels-non-caus-rfs}
and~\ref{sec-macroscop-rels} developed two main ways of interrelating the receptive
field responses between different parameter settings for this
generalized Gaussian derivative model for visual receptive fields,
either in terms of (i)~infinitesimal differential relations closely
related to the notions of Lie groups and the infinitesimal generators
of continuous semi-groups, or
(ii)~macroscopic cascade smoothing relations, closely related to the
notion of Lie algebras, although with a directional preference, in the
respect that the evolution can only performed in positive directions
for those dimensions of the parameter space that correspond to
uni-directional semi-group structures, as opposed to bi-directional
Lie-group structures.

Specifically, in Section~\ref{sec-inf-rels-non-caus-rfs}, we studied
three variants of models for spatial or spatio-temporal smoothing
operations in the receptive fields based on pure Gaussian smoothing
over the spatial or spatial-temporal domains, in terms of either
(i)~a purely spatial model with 3 effective degrees of freedom in the
parameter space,
(ii)~a joint spatio-temporal model with a restriction to isotropic
Gaussian smoothing over the spatial domain, resulting in 4 effective
degrees of freedom in the parameter space, and
(ii)~a joint spatio-temporal model with general affine Gaussian
smoothing over the spatial domain, leading to 6 effective degrees of
freedom in the parameter space. By the special algebraic structures of
the resulting receptive field representations, when using solely
Gaussian smoothing over both the spatial and the temporal domains,
we could introduce the notion of generalized Hermite polynomials of
the resulting spatial or spatio-temporal smoothing operations, to
substantially simplify the process of deriving and proving the
corresponding differential evolution equations over infinitesimal
perturbations of the filter parameters, largely corresponding to the
notion of infinitesimal generators for more traditional semi-groups.

Then, in Section~\ref{sec-macroscop-rels} we derived macroscopic
evolution properties of the either purely spatial or joint
spatio-temporal receptive field representations based on the sole use
of Gaussian smoothing operators over both the spatial and the temporal
domains, by combining
the separate Gaussian smoothing operations over the either spatial or
temporal domains into a joint spatio-temporal Gaussian smoothing
kernel over the joint spatio-temporal domain.
Specifically, by calculating the joint spatio-temporal
covariance matrix of that 2+1-D joint spatio-temporal smoothing
kernel, we derived a joint spatio-temporal cascade property, which
specifies how a receptive field response at a coarser spatial and
temporal scale can be computed by applying an incremental
spatio-temporal smoothing kernel to the receptive field responses at
finer spatial and temporal scales.

The way that the macroscopic cascade smoothing properties
in Section~\ref{sec-macroscop-rels} are related to the infinitesimal
evolution properties in Section~\ref{sec-inf-rels-non-caus-rfs}
is structurally similar to the way a Lie algebra is related to a Lie
group according to the exponential map,
see {\em e.g.\/}\ Hall (\citeyear{Hal15-book}) Section~3.7.
A structural difference compared to Lie groups and Lie algebras,
however, is that here we have directional preferences in the evolution
equations with regard to some of the parameters, thus with closer
relationships to the notions of infinitesimal generators and
semi-groups, as described by Hille and Phillips (\citeyear{HilPhi57}),
Pazy (\citeyear{Paz83-Book}) and Goldstein (\citeyear{Gol85-book}).
The evolution equations that we have derived are, however, neither
strict infinitesimal generators or semi-groups with respect to all the
parameters.
In this respect, the derived relationships can be seen as hybrid
relationships between Lie groups and Lie algebras on one side and
infinitesimal generators and semi-groups on the other side.

In Section~\ref{sec-inf-macro-rels-time-caus-spat-temp}, we then
extended the relationships derived for the non-causal fully
Gaussian-based spatio-temporal model to the time-causal
spatio-temporal model, where the temporal smoothing operation is
performed by convolution with the time-causal limit kernel (\ref{eq-time-caus-lim-kern})
instead of the non-causal temporal Gaussian kernel
(\ref{eq-non-caus-temp-gauss}). Due to the discrete nature of the
temporal scale levels $\tau_k$ in the corresponding spatio-temporal scale
representation, differential evolution properties over the temporal
scale parameter cannot be expressed for that time-causal
spatio-temporal scale-space representation. Additionally, due to
the current lack of sufficiently good tools to derive compact
closed-form expressions for the temporal derivatives of the
time-causal limit kernel, we restricted the differential evolution
properties to derivatives with respect the the spatial scale parameter $s$
and the spatial covariance matrix $\Sigma$ of the spatial affine Gaussian
kernel, which are similar as for the non-causal spatio-temporal 
scale-space representation studied in
Section~\ref{sec-lie-group-spattemp-iso}.
Similarly, regarding the notion of a cascade property for the case of
time-causal spatio-temporal receptive fields, we restricted the
treatment to the special case when the image velocity of the
incremental kernel is equal to the image velocities used for
computing the receptive field responses before and after the cascade
smoothing step, which are also assumed to be equal. 

From the viewpoint of the underlying normative theory for visual
receptive fields, the receptive field shapes are parameterized by
over the 4-D (although effectively 3-D) parameters $(s, \Sigma)$
in the purely spatial case and by the 7-D (although effectively 6-D)
parameters $(s, \Sigma, \tau, v)$ in the joint spatio-temporal case.
If we regard this theory as an idealized model of the visual
processing in the primary visual cortex, then
flattening out these higher-dimensional spaces onto the 2-D cortical
surface, implies that points $p$ and parameters $P$ that may be
relatively nearby in the higher-dimensional spaces
$(p;\; P) = (x;\; s, \Sigma)$ or $(p;\; P)  = (x;\; s, \Sigma, \tau, v)$
may be further away from each other when flattened out on the 2-D
cortical surface.

In this way, if we transfer the cascade smoothing
properties derived in this paper
onto such a flattened cortical manifold, then the corresponding
connections between receptive fields at different positions on
the cortical surface will lead to a set of
``horizontal connections'' between different neurons.
Specifically, due to the flattening of the higher-dimensional manifold
of parameterized receptive responses onto the 2-D cortical surface,
some of these connections will by necessity be ``long range'' on the
cortical surface, although they would have been of shorter range in
the original higher-dimensional manifold of receptive field responses.
In this respect, the presented theory offers a possible explanation
of a certain kind of ``horizontal connections'' constructed from a different purpose than the
more commonly used explanation of integrating receptive field
responses at different positions in the visual field into
higher-order ``gestalt'' entities, such as contours
(see {\em e.g.\/}\ Bolz and Gilbert (\citeyear{BolGIl89-EJNeuroSci}),
Bosking {\em et al.\/} (\citeyear{BosZhaSchFit97-JNeuroSci}),
Ben-Shahar and Zucker (\citeyear{BenZuc04-NeurComp}) and
Sarti {\em et al.\/} (\citeyear{SarCitPet09-JPhysPar})).%
\footnote{Let us, however, remark that with this statement we do not
  in any way intend to argue against the use of ``horizontal
  connections'' between different visual neurons for computing
  higher-order ``gestalt'' entities, such as contours. Instead, our
  purpose is to propose a complementary explanation regarding
  a possible subset of such ``horizontal connections''.}

For modelling visual receptive fields, there is, as previously
described in Section~\ref{sec-rel-work}, also an alternative school of
using idealized models based on Gabor functions, which consist of the
multiplication of Gaussian functions with complex sine waves.
For the purely Gaussian-smoothing-based component in the Gabor model,
it seems likely that structurally related evolution properties over
the receptive field parameters could be obtained for the corresponding
Gabor models of visual receptive fields. Regarding variations in the
wavelength or the orientation of the complex sine
wave component in the Gabor model, the corresponding relationships could,
however, be expected to be of a different nature.

In this treatment,
we have therefore focused solely on the topic of deriving evolution
properties over the parameters of the generalized Gaussian derivative
model,
which obeys very special closedness properties, that could not be
expected to hold for more generic classes of kernels. Thus,
we leave the topic of deriving corresponding infinitesimal, or, if
possible, macroscopic relationships for the Gabor model to future
work. For a complementary in-depth analysis of similarities
{\em vs.\/}\ differences between the generalized Gaussian derivative
and the Gabor models regarding orientation selectivity properties,
see Lindeberg (\citeyear{Lin25-JCompNeurSci-orisel}).

By the derived evolution properties over variations of the filter
parameters of the receptive fields according to the generalized
Gaussian derivative model for visual receptive fields, we do foremost obtain a better
theoretical understanding about how receptive field responses are
related between different values of the filter parameters.
These either infinitesimal or macroscopic evolution properties can, in
turn, be used in computational schemes to compute banks of receptive
field responses suitable for matching the responses of receptive
fields applied to image data that have been acquired under substantial
variations in the viewing conditions, which according to the theory
described in Section~\ref{eq-cov-props} implies that the shapes of the
receptive fields would have to be expanded over the degrees of freedom
of the corresponding geometric image transformations. Thereby, it
becomes possible to, to first order of approximation, match the receptive
field responses between different views of the same object or a
similar spatio-temporal event, and thus establish an identity
operation between the receptive field responses computed
for different observations of the same object or a similar
spatio-temporal event.

These theoretical results do also have possible implications for
theories about biological vision, where indeed such a model concerning
an expansion over receptive field shapes over the degrees of freedom,
corresponding to the degrees of freedom in the here studied
generalized Gaussian derivative model for visual receptive fields,
has in Lindeberg (\citeyear{Lin25-arXiv-cov-props-review}) been recently
proposed for predicting and explaining variabilities of simple cells
in the primary visual cortex of higher mammals.
From statistics of the populations of neurons in the early visual
pathway, with about 1~M output channels from the retina are mapped to 1~M
output channels from the lateral geniculate
nucleus (LGN) to the primary visual cortex (V1),
to about 190~M neurons in V1 with 37~M output channels
(see DiCarlo {\em et al.\/} (\citeyear{DiCZocRus12-Neuron}) Figure~3),
such a substantial expansion of the number of receptive fields
from the LGN to V1 would indeed be
consistent with an expansion of the shapes of the receptive fields
over shape parameters of the receptive fields.

Specifically, the evolution properties between receptive field
responses for different sets of filter parameters derived in this paper
provide a
theoretical explanation for the alternative design strategy of a
representation of the receptive field responses of simple cells
proposed in Lindeberg (\citeyear{Lin25-arXiv-cov-props-review}),
where an idealized vision system could by the use of cascade smoothing
properties over the spatial and temporal scale parameters choose to
only implement receptive field responses corresponding to the finest
spatial and temporal scale levels in the sensorium, corresponding to
the simple cells in the primary visual cortex of higher mammals.
Then, at higher levels in the visual hierarchy, equivalent receptive
field responses, corresponding to coarser levels of spatial and
temporal scales, could be computed in an indirect manner, either based on
integration of the infinitesimal evolution equations derived 
in Section~\ref{sec-inf-rels-non-caus-rfs}, or based on the macroscopic
evolution properties derived in Section~\ref{sec-macroscop-rels}.

See also Lindeberg (\citeyear{Lin25-BICY}) for a closely related theoretical
study of this problem in relation to the degrees of freedom of spatial
affine transformations, the degrees of freedom of the spatial
component of the generalized Gaussian derivative model, and
existing neurophysiological support for such variabilities of the
shapes of simple cells in the primary visual cortex.

\appendix

\section*{Appendix}

\section{Generalized Hermite polynomials for the isotropic spatio-temporal
  Gaussian kernel $T(x, t;\; s, \tau, v)$}
\label{app-ders-spattemp-iso}

\begin{figure*}[hbt]
  \begin{align}
    \begin{split}
      \frac{\partial_{x_1} T(x, t;\; s, \tau, v)}{T(x, t;\; s, \tau, v)}
      = \frac{t v_1-x_1}{s},
    \end{split}\\
    \begin{split}
      \frac{\partial_{x_2} T(x, t;\; s, \tau, v)}{T(x, t;\; s, \tau, v)}
      = \frac{t v_2-x_2}{s},
    \end{split}\\
    \begin{split}
      \frac{\partial_{t} T(x, t;\; s, \tau, v)}{T(x, t;\; s, \tau, v)}
      = \frac{-t \left(v_1^2+v_2^2\right)+v_1 x_1+v_2
   x_2}{s}-\frac{t}{\tau },
    \end{split}\\
    \begin{split}
      \frac{\partial_{\bar{t}} T(x, t;\; s, \tau, v)}{T(x, t;\; s, \tau, v)}
      = -\frac{t}{\tau },
    \end{split}\\
    \begin{split}
      \frac{\partial_{x_1 x_1} T(x, t;\; s, \tau, v)}{T(x, t;\; s, \tau, v)}
      = \frac{(x_1-t v_1)^2-s}{s^2},
    \end{split}\\
    \begin{split}
      \frac{\partial_{x_1 x_2} T(x, t;\; s, \tau, v)}{T(x, t;\; s, \tau, v)}
      = \frac{(t v_1-x_1) (t v_2-x_2)}{s^2},
    \end{split}\\
    \begin{split}
      \frac{\partial_{x_2 x_2} T(x, t;\; s, \tau, v)}{T(x, t;\; s, \tau, v)}
      = \frac{(x_2-t v_2)^2-s}{s^2},
    \end{split}\\
    \begin{split}
      \frac{\partial_{x_1 t} T(x, t;\; s, \tau, v)}{T(x, t;\; s, \tau, v)}
      = \frac{s \left(t^2 (-v_1)+t x_1+\tau  v_1\right)-\tau  (t
   v_1-x_1) \left(t \left(v_1^2+v_2^2\right)-v_1
   x_1-v_2 x_2\right)}{s^2 \tau },
    \end{split}\\
    \begin{split}
      \frac{\partial_{x_2 t} T(x, t;\; s, \tau, v)}{T(x, t;\; s, \tau, v)}
      = \frac{s \left(t^2 (-v_2)+t x_2+\tau  v_2\right)-\tau  (t
   v_2-x_2) \left(t \left(v_1^2+v_2^2\right)-v_1
   x_1-v_2 x_2\right)}{s^2 \tau },
    \end{split}\\
    \begin{split}
      \frac{\partial_{x_1 \bar{t}} T(x, t;\; s, \tau, v)}{T(x, t;\; s, \tau, v)}
      = \frac{t (x_1-t v_1)}{s \tau },
    \end{split}\\
    \begin{split}
      \frac{\partial_{x_2 \bar{t}} T(x, t;\; s, \tau, v)}{T(x, t;\; s, \tau, v)}
      = \frac{t (x_2-t v_2)}{s \tau },
    \end{split}\\
    \begin{split}
      \frac{\partial_{tt} T(x, t;\; s, \tau, v)}{T(x, t;\; s, \tau, v)}
      = \frac{\left(-t \left(v_1^2+v_2^2\right)+v_1 x_1+v_2
   x_2\right)^2}{s^2}-\frac{-2 t^2 \left(v_1^2+v_2^2\right)+2 t
   (v_1 x_1+v_2 x_2)+\tau 
   \left(v_1^2+v_2^2\right)}{s \tau }+\frac{t^2-\tau }{\tau ^2},
     \end{split}\\
    \begin{split}
      \frac{\partial_{\bar{t}\bar{t}} T(x, t;\; s, \tau, v)}{T(x, t;\; s, \tau, v)}
      = \frac{t^2-\tau }{\tau ^2},
    \end{split}\\
    \begin{split}
      \frac{\partial_{s} T(x, t;\; s, \tau, v)}{T(x, t;\; s, \tau, v)}
      = \frac{-2 s+t^2 \left(v_1^2+v_2^2\right)-2 t (v_1 x_1+v_2
   x_2)+x_1^2+x_2^2}{2 s^2},
    \end{split}\\
    \begin{split}
      \frac{\partial_{\tau} T(x, t;\; s, \tau, v)}{T(x, t;\; s, \tau, v)}
      = \frac{t^2-\tau }{2 \tau ^2},
    \end{split}\\
    \begin{split}
      \frac{\partial_{v_1} T(x, t;\; s, \tau, v)}{T(x, t;\; s, \tau, v)}
      = \frac{t (x_1-t v_1)}{s},
    \end{split}\\
    \begin{split}
      \frac{\partial_{v_2} T(x, t;\; s, \tau, v)}{T(x, t;\; s, \tau, v)}
      = \frac{t (x_2-t v_2)}{s}.
    \end{split}
  \end{align}
  
  \caption{Generalized Hermite polynomials as arising from
    derivatives of the isotropic spatio-temporal Gaussian kernel
    $T(x, t;\; s, \tau, v)$ according to
    (\ref{eq-isotrop-spattemp-gauss}), based on smoothing with a
    spatially isotropic Gaussian kernel, with respect to the image coordinates
    $x = (x_1, x_2)^T$ and the time variable $t$ up to order 2,
    as well as with respect to the spatial
    scale parameter $s$, the temporal scale parameter $\tau$ and the
    elements $v_1$ and $v_2$ of the velocity vector $v$.}
  \label{fig-ders-spattemp-iso}
\end{figure*}

Figure~\ref{fig-ders-spattemp-iso} lists generalized Hermite
polynomials for the isotropic spatio-temporal Gaussian kernel
$T(x, t;\; s, \tau, v)$ according to
(\ref{eq-isotrop-spattemp-gauss}).

\begin{figure*}[hbt]
  \begin{align}
    \begin{split}
      \frac{\partial_{x_1} T(x, t;\; s, \Sigma, \tau, v)}{T(x, t;\; s, \Sigma, \tau, v)}
      = \frac{\Sigma_{12} t v_2-\Sigma_{12} x_2-\Sigma_{22} t v_1+\Sigma_{22}
   x_1}{s \left(\Sigma_{12}^2-\Sigma_{11} \Sigma_{22}\right)},
    \end{split}\\
    \begin{split}
      \frac{\partial_{x_2} T(x, t;\; s, \Sigma, \tau, v)}{T(x, t;\; s, \Sigma, \tau, v)}
      = \frac{-\Sigma_{11} t v_2+\Sigma_{11} x_2+\Sigma_{12} t v_1-\Sigma_{12}
   x_1}{\Sigma_{12}^2 s-\Sigma_{11} \Sigma_{22} s},
    \end{split}\\
    \begin{split}
      \frac{\partial_{t} T(x, t;\; s, \Sigma, \tau, v)}{T(x, t;\; s, \Sigma, \tau, v)}
      = \frac{\Sigma_{11} \Sigma_{22} s t+\Sigma_{11} \tau  v_2 (t
   v_2-x_2)+\Sigma_{12}^2 (-s) t+\Sigma_{12} \tau  (-2 t v_1
   v_2+v_1 x_2+v_2 x_1)+\Sigma_{22} \tau  v_1 (t
   v_1-x_1)}{s \tau  \left(\Sigma_{12}^2-\Sigma_{11} \Sigma_{22}\right)},
    \end{split}\\
    \begin{split}
      \frac{\partial_{\bar{t}} T(x, t;\; s, \Sigma, \tau, v)}{T(x, t;\; s, \Sigma, \tau, v)}
      = -\frac{t}{\tau },
    \end{split}\\
    \begin{split}
      \frac{\partial_{x_1 x_1} T(x, t;\; s, \Sigma, \tau, v)}{T(x, t;\; s, \Sigma, \tau, v)}
      = \frac{\Sigma_{22}^2 \left((x_1-t v_1)^2-\Sigma_{11} s\right)+\Sigma_{12}^2
   \left(\Sigma_{22} s+(x_2-t v_2)^2\right)-2 \Sigma_{12} \Sigma_{22} (t
   v_1-x_1) (t v_2-x_2)}{s^2 \left(\Sigma_{12}^2-\Sigma_{11}
   \Sigma_{22}\right)^2},
    \end{split}\\
    \begin{split}
      \frac{\partial_{x_1 x_2} T(x, t;\; s, \Sigma, \tau, v)}{T(x, t;\; s, \Sigma, \tau, v)}
      = \frac{1}{s^2 \left(\Sigma_{12}^2-\Sigma_{11} \Sigma_{22}\right)^2}
      \left(
      \Sigma_{11} \Sigma_{12} \Sigma_{22} s-\Sigma_{11} \Sigma_{12} (x_2-t
   v_2)^2+\Sigma_{11} \Sigma_{22} (t v_1-x_1) (t
   v_2-x_2) + \right.
   \end{split}\nonumber\\
    \begin{split}
      \left.
        \quad\quad \quad\quad \quad\quad \quad\quad \quad\quad \quad\quad
        \quad\quad \quad\quad \quad\quad \quad\quad \quad\quad
   +\Sigma_{12}^3 (-s)+\Sigma_{12}^2 (t v_1-x_1) (t
   v_2-x_2)-\Sigma_{12} \Sigma_{22} (x_1-t v_1)^2
   \right)
    \end{split}\\
    \begin{split}
      \frac{\partial_{x_2 x_2} T(x, t;\; s, \Sigma, \tau, v)}{T(x, t;\; s, \Sigma, \tau, v)}
      = \frac{\Sigma_{11}^2 \left((x_2-t v_2)^2-\Sigma_{22} s\right)+\Sigma_{11}
   \Sigma_{12} (\Sigma_{12} s-2 (t v_1-x_1) (t
   v_2-x_2))+\Sigma_{12}^2 (x_1-t v_1)^2}{s^2
   \left(\Sigma_{12}^2-\Sigma_{11} \Sigma_{22}\right)^2},
    \end{split}\\
     \begin{split}
      \frac{\partial_{x_1 \bar{t}} T(x, t;\; s, \Sigma, \tau, v)}{T(x, t;\; s, \Sigma, \tau, v)}
      = \frac{t (-\Sigma_{12} t v_2+\Sigma_{12} x_2+\Sigma_{22} t v_1-\Sigma_{22}
   x_1)}{s \tau  \left(\Sigma_{12}^2-\Sigma_{11} \Sigma_{22}\right)},
    \end{split}\\
    \begin{split}
      \frac{\partial_{x_2 \bar{t}} T(x, t;\; s, \Sigma, \tau, v)}{T(x, t;\; s, \Sigma, \tau, v)}
      = \frac{t (\Sigma_{11} t v_2-\Sigma_{11} x_2-\Sigma_{12} t v_1+\Sigma_{12}
   x_1)}{s \tau  \left(\Sigma_{12}^2-\Sigma_{11} \Sigma_{22}\right)},
    \end{split}\\
    \begin{split}
      \frac{\partial_{\bar{t}\bar{t}} T(x, t;\; s, \Sigma, \tau, v)}{T(x, t;\; s, \Sigma, \tau, v)}
      = \frac{t^2-\tau }{\tau ^2},
    \end{split}\\
    \begin{split}
      \frac{\partial_{s} T(x, t;\; s, \Sigma, \tau, v)}{T(x, t;\; s, \Sigma, \tau, v)}
      = -\frac{\Sigma_{11} \left((x_2-t v_2)^2-2 \Sigma_{22} s\right)+2 \Sigma_{12}^2
   s-2 \Sigma_{12} (t v_1-x_1) (t v_2-x_2)+\Sigma_{22} (x_1-t
   v_1)^2}{2 s^2 \left(\Sigma_{12}^2-\Sigma_{11} \Sigma_{22}\right)},
    \end{split}\\
    \begin{split}
      \frac{\partial_{\Sigma_{11}} T(x, t;\; s, \Sigma, \tau, v)}{T(x, t;\; s, \Sigma, \tau, v)}
      = \frac{\Sigma_{22}^2 \left((x_1-t v_1)^2-\Sigma_{11} s\right)+\Sigma_{12}^2
   \left(\Sigma_{22} s+(x_2-t v_2)^2\right)-2 \Sigma_{12} \Sigma_{22} (t
   v_1-x_1) (t v_2-x_2)}{2 s \left(\Sigma_{12}^2-\Sigma_{11}
   \Sigma_{22}\right)^2},
    \end{split}\\
    \begin{split}
      \frac{\partial_{\Sigma_{12}} T(x, t;\; s, \Sigma, \tau, v)}{T(x, t;\; s, \Sigma, \tau, v)}
      = \frac{1}{s \left(\Sigma_{12}^2-\Sigma_{11} \Sigma_{22}\right)^2}
      \left(
      \Sigma_{11} \Sigma_{12} \Sigma_{22} s-\Sigma_{11} \Sigma_{12} (x_2-t
   v_2)^2+\Sigma_{11} \Sigma_{22} (t v_1-x_1) (t
   v_2-x_2)+
   \right.
    \end{split}\nonumber\\
    \begin{split}
      \left.
        \quad\quad  \quad\quad  \quad\quad  \quad\quad  \quad\quad
        \quad\quad  \quad\quad  \quad\quad  \quad\quad  \quad\quad
   +\Sigma_{12}^3 (-s)+\Sigma_{12}^2 (t v_1-x_1) (t
   v_2-x_2)-\Sigma_{12} \Sigma_{22} (x_1-t v_1)^2
   \right),
    \end{split}\\
    \begin{split}
      \frac{\partial_{\Sigma_{22}} T(x, t;\; s, \Sigma, \tau, v)}{T(x, t;\; s, \Sigma, \tau, v)}
      = \frac{\Sigma_{11}^2 \left((x_2-t v_2)^2-\Sigma_{22} s\right)+\Sigma_{11}
   \Sigma_{12} (\Sigma_{12} s-2 (t v_1-x_1) (t
   v_2-x_2))+\Sigma_{12}^2 (x_1-t v_1)^2}{2 s
   \left(\Sigma_{12}^2-\Sigma_{11} \Sigma_{22}\right)^2},
    \end{split}\\
    \begin{split}
      \frac{\partial_{\tau} T(x, t;\; s, \Sigma, \tau, v)}{T(x, t;\; s, \Sigma, \tau, v)}
      = \frac{t^2-\tau }{2 \tau ^2},
    \end{split}\\
    \begin{split}
      \frac{\partial_{v_1} T(x, t;\; s, \Sigma, \tau, v)}{T(x, t;\; s, \Sigma, \tau, v)}
      = \frac{t (-\Sigma_{12} t v_2+\Sigma_{12} x_2+\Sigma_{22} t v_1-\Sigma_{22}
   x_1)}{s \left(\Sigma_{12}^2-\Sigma_{11} \Sigma_{22}\right)},
    \end{split}\\
    \begin{split}
      \frac{\partial_{v_2} T(x, t;\; s, \Sigma, \tau, v)}{T(x, t;\; s, \Sigma, \tau, v)}
      = -\frac{t (-\Sigma_{11} t v_2+\Sigma_{11} x_2+\Sigma_{12} t v_1 -\Sigma_{12}
   x_1)}{s \left(\Sigma_{12}^2-\Sigma_{11} \Sigma_{22}\right)}.
    \end{split}
  \end{align}
  
  \caption{Generalized Hermite polynomials as arising from
    derivatives of the affine spatio-temporal Gaussian kernel
    $T(x, t;\; s, \Sigma, \tau, v)$ according to
    (\ref{eq-aff-spattemp-gauss}), based on smoothing with a
    spatially anisotropic affine Gaussian kernel, with respect to the image coordinates
    $x = (x_1, x_2)^T$ and the time variable $t$ up to order 2, as
    well as with respect to the spatial
    scale parameter $s$, the elements $\Sigma_{11}$, $\Sigma_{12}$ and $\Sigma_{22}$
of the spatial covariance matrix $\Sigma$,  the temporal scale parameter $\tau$ and the
elements $v_1$ and $v_2$ of the velocity vector $v$.
(The explicit expressions for $\partial_{x_1 t} T(x, t;\; s, \Sigma, \tau, v)$,
$\partial_{x_2 t} T(x, t;\; s, \Sigma, \tau, v)$ and $\partial_{tt} T(x, t;\; s, \Sigma, \tau, v)$
have been omitted here, because they do not fit within two lines each.)}
\label{fig-ders-spattemp-aff}
\end{figure*}

\section{Generalized Hermite polynomials for the affine spatio-temporal
  Gaussian kernel $T(x, t;\; s, \Sigma, \tau, v)$}
\label{app-ders-spattemp-aff}

Figure~\ref{fig-ders-spattemp-aff} lists generalized Hermite
polynomials for the affine spatio-temporal Gaussian kernel
$T(x, t;\; s, \Sigma, \tau, v)$ according to
(\ref{eq-aff-spattemp-gauss}).

\begin{figure*}[hbt]
  \begin{align}
    \begin{split}
      \frac{\partial_{x_1} T(x, t;\; s, \Sigma, \tau, v, c)}{T(x, t;\;
        s, \Sigma, \tau, v, c)}
      = \frac{\Sigma_{12} t v_2-\Sigma_{12} x_2-\Sigma_{22} t v_1+\Sigma_{22}
   x_1}{s \left(\Sigma_{12}^2-\Sigma_{11} \Sigma_{22}\right)},
    \end{split}\\
    \begin{split}
      \frac{\partial_{x_2} T(x, t;\; s, \Sigma, \tau, v, c)}{T(x, t;\;
        s, \Sigma, \tau, v, c)}
      = \frac{-\Sigma_{11} t v_2+\Sigma_{11} x_2+\Sigma_{12} t v_1-\Sigma_{12}
   x_1}{\Sigma_{12}^2 s-\Sigma_{11} \Sigma_{22} s},
    \end{split}\\
    \begin{split}
      \frac{\partial_{x_1 x_2} T(x, t;\; s, \Sigma, \tau, v, c)}{T(x,
        t;\; s, \Sigma, \tau, v, c)}
      = \frac{1}{s^2 \left(\Sigma_{12}^2-\Sigma_{11} \Sigma_{22}\right)^2}
      \left(
      \Sigma_{11} \Sigma_{12} \Sigma_{22} s-\Sigma_{11} \Sigma_{12} (x_2-t
   v_2)^2+\Sigma_{11} \Sigma_{22} (t v_1-x_1) (t
   v_2-x_2) + \right.
   \end{split}\nonumber\\
    \begin{split}
      \left.
        \quad\quad \quad\quad \quad\quad \quad\quad \quad\quad \quad\quad
        \quad\quad \quad\quad \quad\quad \quad\quad \quad\quad
   +\Sigma_{12}^3 (-s)+\Sigma_{12}^2 (t v_1-x_1) (t
   v_2-x_2)-\Sigma_{12} \Sigma_{22} (x_1-t v_1)^2
   \right),
    \end{split}\\
    \begin{split}
      \frac{\partial_{x_2 x_2} T(x, t;\; s, \Sigma, \tau, v, c)}{T(x,
        t;\; s, \Sigma, \tau, v, c)}
      = \frac{\Sigma_{11}^2 \left((x_2-t v_2)^2-\Sigma_{22} s\right)+\Sigma_{11}
   \Sigma_{12} (\Sigma_{12} s-2 (t v_1-x_1) (t
   v_2-x_2))+\Sigma_{12}^2 (x_1-t v_1)^2}{s^2
   \left(\Sigma_{12}^2-\Sigma_{11} \Sigma_{22}\right)^2},
    \end{split}\\
    \begin{split}
      \frac{\partial_{s} T(x, t;\; s, \Sigma, \tau, v, c)}{T(x, t;\;
        s, \Sigma, \tau, v, c)}
      = -\frac{\Sigma_{11} \left((x_2-t v_2)^2-2 \Sigma_{22} s\right)+2 \Sigma_{12}^2
   s-2 \Sigma_{12} (t v_1-x_1) (t v_2-x_2)+\Sigma_{22} (x_1-t
   v_1)^2}{2 s^2 \left(\Sigma_{12}^2-\Sigma_{11} \Sigma_{22}\right)},
    \end{split}\\
    \begin{split}
      \frac{\partial_{\Sigma_{11}} T(x, t;\; s, \Sigma, \tau, v,
        c)}{T(x, t;\; s, \Sigma, \tau, v, c)}
      = \frac{\Sigma_{22}^2 \left((x_1-t v_1)^2-\Sigma_{11} s\right)+\Sigma_{12}^2
   \left(\Sigma_{22} s+(x_2-t v_2)^2\right)-2 \Sigma_{12} \Sigma_{22} (t
   v_1-x_1) (t v_2-x_2)}{2 s \left(\Sigma_{12}^2-\Sigma_{11}
   \Sigma_{22}\right)^2},
    \end{split}\\
    \begin{split}
      \frac{\partial_{\Sigma_{12}} T(x, t;\; s, \Sigma, \tau, v,
        c)}{T(x, t;\; s, \Sigma, \tau, v, c)}
      = \frac{1}{s \left(\Sigma_{12}^2-\Sigma_{11} \Sigma_{22}\right)^2}
      \left(
      \Sigma_{11} \Sigma_{12} \Sigma_{22} s-\Sigma_{11} \Sigma_{12} (x_2-t
   v_2)^2+\Sigma_{11} \Sigma_{22} (t v_1-x_1) (t
   v_2-x_2)+
   \right.
    \end{split}\nonumber\\
    \begin{split}
      \left.
        \quad\quad  \quad\quad  \quad\quad  \quad\quad  \quad\quad\quad
        \quad\quad  \quad\quad  \quad\quad  \quad\quad  \quad\quad\,\,
   +\Sigma_{12}^3 (-s)+\Sigma_{12}^2 (t v_1-x_1) (t
   v_2-x_2)-\Sigma_{12} \Sigma_{22} (x_1-t v_1)^2
   \right),
    \end{split}\\
    \begin{split}
      \frac{\partial_{\Sigma_{22}} T(x, t;\; s, \Sigma, \tau, v,
        c)}{T(x, t;\; s, \Sigma, \tau, v, c)}
      = \frac{\Sigma_{11}^2 \left((x_2-t v_2)^2-\Sigma_{22} s\right)+\Sigma_{11}
   \Sigma_{12} (\Sigma_{12} s-2 (t v_1-x_1) (t
   v_2-x_2))+\Sigma_{12}^2 (x_1-t v_1)^2}{2 s
   \left(\Sigma_{12}^2-\Sigma_{11} \Sigma_{22}\right)^2},
    \end{split}\\
    \begin{split}
      \frac{\partial_{v_1} T(x, t;\; s, \Sigma, \tau, v, c)}{T(x, t;\;
        s, \Sigma, \tau, v, c)}
      = \frac{t (-\Sigma_{12} t v_2+\Sigma_{12} x_2+\Sigma_{22} t v_1-\Sigma_{22}
   x_1)}{s \left(\Sigma_{12}^2-\Sigma_{11} \Sigma_{22}\right)},
    \end{split}\\
    \begin{split}
      \frac{\partial_{v_2} T(x, t;\; s, \Sigma, \tau, v, c)}{T(x, t;\;
        s, \Sigma, \tau, v, c)}
      = -\frac{t (-\Sigma_{11} t v_2+\Sigma_{11} x_2+\Sigma_{12} t v_1 -\Sigma_{12}
   x_1)}{s \left(\Sigma_{12}^2-\Sigma_{11} \Sigma_{22}\right)}.
    \end{split}
  \end{align}
  
  \caption{Generalized Hermite polynomials as arising from
    derivatives of the time-causal affine spatio-temporal kernel
    $T(x, t;\; s, \Sigma, \tau, v, c)$ according to
    (\ref{eq-aff-spattemp-kern-timecaus}), based on smoothing with a
    spatially anisotropic affine Gaussian kernel, with respect to only
    the image coordinates
    $x = (x_1, x_2)^T$ up to order 2, as
    well as with respect to only the spatial
    scale parameter $s$, the elements $\Sigma_{11}$, $\Sigma_{12}$ and $\Sigma_{22}$
of the spatial covariance matrix $\Sigma$ and the elements $v_1$ and
$v_2$ of the velocity vector $v$.}
\label{fig-ders-spattemp-aff-timecaus}
\end{figure*}

\section{Generalized Hermite polynomials for the time-causal 
  affine spatio-temporal kernel $T(x, t;\; s, \Sigma, \tau, v, c)$}
\label{app-ders-spattemp-aff-timecaus}

Figure~\ref{fig-ders-spattemp-aff-timecaus} lists generalized Hermite
polynomials for the time-causal affine spatio-temporal Gaussian kernel
$T(x, t;\; s, \Sigma, \tau, v, c)$ according to
(\ref{eq-aff-spattemp-kern-timecaus}).

{\footnotesize
\bibliographystyle{abbrvnat}
\bibliography{defs,tlmac}}

\end{document}